\documentclass{article}

\usepackage{arxiv}

\usepackage[utf8]{inputenc}
\usepackage[T1]{fontenc}
\usepackage{acronym}
\usepackage{comment}
\usepackage{graphicx,color}
\usepackage{graphics}
\usepackage{caption}
\usepackage{subcaption}
\usepackage{booktabs}
\usepackage{amsmath}
\usepackage{hyperref}       
\usepackage{url}            
\usepackage{nicefrac}       
\usepackage{placeins}
\usepackage{ kantlipsum}

\usepackage{siunitx}
\usepackage{array}
\usepackage{makecell}
\usepackage{enumitem}
    \newlist{tabenum}{enumerate}{1}
    \setlist[tabenum]{nosep,
                      label*=\textbf{\alph*)},
                      wide,
                      before=\begin{minipage}[t]{\hsize},%
                      after=\end{minipage}}
\usepackage{newtxtext}
\usepackage{microtype}
\usepackage{xcolor}
\usepackage{tabularray}
\UseTblrLibrary{varwidth}  

\usepackage{array,ragged2e,longtable}
\usepackage{ulem}
\usepackage{comment}
\usepackage{multirow}
\usepackage[most]{tcolorbox}

\acrodef{ai}[AI]{Artificial Intelligence}
\acrodef{sgx}[SGX]{Software Guard Extensions}
\acrodef{ico}[ICO]{Information Commissioner's Office}
\acrodef{ml}[ML]{Machine Learning}
\acrodef{fido}[FIDO]{Framework for Inhibiting Data Overcollection}
\acrodef{ehr}[EHR]{Electronic Health Record}
\acrodef{sail}[SAIL]{Secure Anonymised Information Linkage}
\acrodef{id}[ID]{Intellectual Disability}
\acrodef{nhs}[NHS]{National Health Service}
\acrodef{tre}[TRE]{Trusted Research Environment}
\acrodef{pedw}[PEDW]{Patient Episode Database for Wales}
\acrodef{edds}[EDDS]{Emergency Department Dataset}
\acrodef{wdsd}[WDSD]{Welsh Demographic Service Dataset}
\acrodef{adde}[ADDE]{Annual District Death Extract}
\acrodef{gp}[GP]{General Practice}
\acrodef{wlgp}[WLGP]{Welsh Longitudinal General Practice}
\acrodef{ai}[AI]{Artificial Intelligence}
\acrodef{gdpr}[GDPR]{General Data Protection Regulation}
\acrodef{dhsc}[DHSC]{Department of Health and Social Care}

\acrodef{nir}[NIR]{Neural Information Retrieval}
\acrodef{plm}[PLM]{Pretrained Language Model}
\acrodef{drmm}[DRMM]{Deep Relevance Matching Model}
\acrodef{knrm}[KNRM]{Kernel-based Neural Ranking Model}
\acrodef{bertdot}[BERTdot]{BERT with Dot Productions}
\acrodef{ewc}[EWC]{Elastic Weight Consolidation}
\acrodef{ewcol}[EWCol]{Elastic Weight Consolidation in the Online Paradigm}
\acrodef{si}[SI]{Synaptic Intelligence}
\acrodef{mas}[MAS]{Memory Aware Synapses}
\acrodef{nr}[NR]{Naive Rehearsal}
\acrodef{gem}[GEM]{Gradient Episodic Memory}
\acrodef{clnir}[CLNIR]{Continual Learning Framework for Neural Information Retrieval}
\acrodef{std}[SD]{Standard Deviation}
\acrodef{mrr}[MRR]{Mean Reciprocal Rank}
\acrodef{fwt}[FWT]{Forward Transfer}
\acrodef{bwt}[BWT]{Backward Transfer}
\acrodef{ld}[LD]{Learning Disability}
\acrodef{ons}[ONS]{Office for National statistics}
\acrodef{lsoa}[LSOA]{Lower-layer Super Output Areas}
\acrodef{ltc}[LTC]{Long Term Condition}
\acrodef{wimd}[WIMD]{Welsh Index of Multiple Deprivation}
\acrodef{lsoa}[LSOA]{Lower layer Super Output Areas}
\acrodef{los}[LOS]{Length Of hospital Stay}
\acrodef{ckd}[CKD]{Chronic Kidney Disease}
\acrodef{iqr}[IQR]{Interquartile Range}
\acrodef{bmi}[BMI]{Body Mass Index}
\acrodef{icd10}[ICD-10]{International Classification of Diseases version 10}
\acrodef{fnr}[FNR]{False Negative Rate}
\acrodef{tpr}[TPR]{True Positive Rate}
\acrodef{fpr}[FPR]{False Positive Rate}
\acrodef{balacc}[BAL ACC]{Balanced Accuracy}
\acrodef{svm}[SVM]{Support Vector Machines}
\acrodef{knn}[KNN]{k-Nearest Neighbors}
\acrodef{xgboost}[XGBoost]{eXtreme Gradient Boosting}
\acrodef{rf}[RF]{Random Forest}
\acrodef{lr}[LR]{Logistic Regression}
\acrodef{gboost}[GBoost]{Gradient Ooosting}
\acrodef{histgboost}[HISTGBoost]{Histogram-based Gradient Boosting}
\acrodef{eg}[EG]{Exponentiated Gradient}
\acrodef{nn}[NN]{Neural Network}
\acrodef{roc}[ROC]{Receiver Operating Characteristic}
\acrodef{auc}[AUC]{Area Under the \ac{roc} Curve}
\acrodef{mltc}[MLTC]{Multiple Long Term Condition}
\acrodef{physical}[PHYSICAL]{Physical exercise}
\acrodef{alf}[ALF]{Anonymised Linking Field}
\acrodef{ssd}[SSD]{Study Start Date}
\acrodef{sed}[SED]{Study End Date}

\acrodef{igrp}[IGRP]{Information Governance Review Panel}
\acrodef{opcs}[OPCS]{Office of Population Censuses and Surveys}
\acrodef{opcs4}[OPCS-4]{\ac{opcs} Classification of Interventions and Procedures version 4}

\acrodef{amd}[AMD]{Acute Macular Degeneration}
\acrodef{ibd}[IBD]{Inflammatory Bowel Disease}
\acrodef{ms}[MS]{Multiple Sclerosis}
\acrodef{pvd}[PVD]{Peripheral Vascular Disease}

\usepackage{doi}

\title{Equitable Length of Stay Prediction for Patients with Learning Disabilities and Multiple Long-term Conditions Using Machine Learning}

\author{ \href{https://orcid.org/0000-0002-4742-3102}{\includegraphics[scale=0.06]{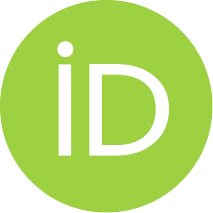}\hspace{1mm}Emeka ~Abakasanga}\\
	Department of Computer Science\\
	Loughborough University\\
	Loughborough, LE11 3TU, UK \\
	\texttt{e.abakasanga@lboro.ac.uk} \\
	\And
	\href{https://orcid.org/0000-0002-9883-0381}{\includegraphics[scale=0.06]{orcid.pdf}\hspace{1mm}Rania ~Kousovista} \\
	Department of Computer Science\\
	Loughborough University\\
	Loughborough, LE11 3TU, UK  \\
        \And
	\href{https://orcid.org/0000-0002-4663-6907}{\includegraphics[scale=0.06]{orcid.pdf}\hspace{1mm}Georgina ~Cosma} \\
	Department of Computer Science\\
	Loughborough University\\
	Loughborough, LE11 3TU, UK  \\
        \And
	{  Ashley ~Akbari} \\
	Population Data Science\\
        Swansea University Medical School\\
    Swansea, Wales, UK \\
        \And
	{  Francesco ~Zaccard} \\
 Diabetes Research Centre\\
	University of Leicester\\
	Leicester, LE1 7RH, UK \\
        \And
	{  Navjot ~Kaur} \\
	School of Design and Creative Arts\\
	Loughborough University\\
	Loughborough, LE11 3TU  \\
        \And
	{  Danielle ~Fitt} \\
	Population Data Science\\
        Swansea University Medical School\\
        Swansea, Wales, UK \\
        \And
	{  Gyuchan T. ~Jun} \\
	School of Design and Creative Arts\\
	Loughborough University\\
	Loughborough, LE11 3TU  \\
        \And
	{  Reza ~Kiani} \\
	Leicestershire Partnership NHS Trust \\
        Leicester, UK
        \And
	{  Satheesh ~Gangadharan} \\
	Leicestershire Partnership NHS Trust \\
        Leicester, UK \\
}

\renewcommand{\shorttitle}{\textit{arXiv} Template}

\hypersetup{
pdftitle={A template for the arxiv style},
pdfsubject={q-bio.NC, q-bio.QM},
pdfauthor={David S.~Hippocampus, Elias D.~Striatum},
pdfkeywords={First keyword, Second keyword, More},
}

\begin{document}
\maketitle

\begin{abstract}
People with learning disabilities have a higher mortality rate and premature deaths compared to the general public,  as reported in published research in the UK and other countries. This study analyses hospitalisations of 9\,618 patients identified with learning disabilities and long-term conditions for the population of Wales using electronic health record (EHR) data sources from the SAIL Databank. We describe the demographic characteristics, prevalence of long-term conditions, medication history, hospital visits, and lifestyle history for our study cohort, and apply machine learning models to predict the length of hospital stays for this cohort. The random forest (RF) model achieved an Area Under the Curve (AUC) of 0.759 (males) and 0.756 (females), a false negative rate of 0.224 (males) and 0.229 (females), and a balanced accuracy of 0.690 (males) and 0.689 (females). After examining model performance across ethnic groups, two bias mitigation algorithms (threshold optimization and the reductions algorithm using an exponentiated gradient) were applied to minimise performance discrepancies. The threshold optimizer algorithm outperformed the reductions algorithm, achieving lower ranges in false positive rate and balanced accuracy for the male cohort across the ethnic groups. This study demonstrates the potential of applying machine learning models with effective bias mitigation approaches on EHR data sources to enable equitable prediction of hospital stays by addressing data imbalances across groups.

\end{abstract}


\section{Introduction}
There are approximately 1.1 million adults aged 18 years and older living with a \ac{ld} in the UK, including over 54\,000 individuals from Wales\cite{mencap2020}. Existing sources show that individuals with \ac{ld} often experience poorer physical and mental health, as well as higher rates of \acp{mltc} and avoidable mortality compared to those without \ac{ld} \cite{mencapInequalities, white2022learning, Tyrer2022, Tyrer2022-re, Tyrer2021-zw, Tyrer2020MortalityPA, heslop2013confidential,carey2016health}. This demographic presents unique needs and challenges that impact their hospitalisations \cite{hatton2016learning}. 

Effectively managing healthcare resources while ensuring optimal patient outcomes poses particular challenges for patients with \ac{ld}. An important outcome of interest for patients with \ac{ld} is a reliable prediction of the \ac{los} of their admission and the underlying factors that could influence their \ac{los} \cite{stone2022systematic}. Predicting the \ac{los} can lead to enhanced healthcare services and further initiate proactive measures to prevent prolonged \ac{los}.  A recent study conducted in the UK discovered that at any given time, approximately 2\,000 patients with \ac{ld} and/or autism in long-stay hospitals have been hospitalised, with over half having spent over 2 years in hospital care. This includes 350 \ac{ld} patients who have been in long-stay hospitals for more than a decade \cite{ince2022we}.  The extended \ac{los} in their study was attributed to either the patient's personal characteristics or limitations of the system supporting them. Another study found that other general factors contributing to prolonged hospital stay for patients who are medically fit for discharge included hospital-acquired infections, falls, and other medical errors\cite{rodziewicz2020medical}. Conversely, there is also a downside to patients being discharged prematurely, as it may result in readmissions or, in severe cases, preventable deaths \cite{alper2017hospital, Alexandre}. Although these studies were carried out on the general population, the conclusion and outcome may still apply to people with \ac{ld} with \acp{mltc}. 

The above-mentioned studies highlight the importance of proactively managing patient discharges as early as possible during their hospitalisation to optimise the \ac{los}.
This paper presents a \ac{los} machine-learning prediction model of patients with \ac{ld} and \acp{mltc} in Wales. The model utilises a dataset of 9\,618 patients with \ac{ld} and a total of 62\,243 hospital admission records. Each admission record was associated with one or more of 39 \acp{ltc} (see Supplementary \ref{tab: ltc_list} for the full list of \acp{ltc}). As \ac{ml} increasingly influences decisions in healthcare, growing interest emerges in assessing and ensuring balanced predictions across sensitive groups\cite{Mittermaier2023-au}, and therefore bias mitigation techniques are employed to improve the performance of the model across ethnic groups.
The contributions of this paper are as follows.

\begin{itemize} [noitemsep]
\item Analysis of hospitalised Welsh \ac{ld} patient population - provided statistics on demographics, prevalence of 39 \acp{ltc}, previous hospitalisation data including prior admissions, episodes, days, and condition prevalence.
\item Identification of primary conditions treated during hospitalisation and the prevalent \acp{ltc} for hospitalised patients with \ac{ld}, along with admission rates per patient.
\item Identification of prevalent \acp{ltc} linked to prolonged hospital stays ($\geq$ 129 days).
\item Statistical analysis to identify factors associated with hospital stays $\geq$ 4 days using non-parametric tests.
\item Development and evaluation of machine learning models to predict whether a patient's \ac{los} would be $<$ 4 days or $\geq$ 4 days, using patient data available up to the first 24 hours of admission.
\item Assessment of model performance differences across ethnic groups. Application and comparison of two bias mitigation algorithms - threshold optimization and reductions algorithm using an exponentiated gradient.
\item Demonstrated potential of applying ML with effective bias mitigation on electronic health records data to promote equitable prediction across groups when predicting LOS.
\end{itemize}

\section*{Results}
\label{sec: results}

The \ac{los} is defined as the number of days an inpatient is hospitalised during a single admission \cite{stone2022systematic}. For each admission, the \ac{los} value was obtained by calculating the difference between the admission and discharge dates. A \ac{los} threshold $\psi$ was obtained by taking the ceiling value of the mean \ac{los} across all extracted admission records, excluding the \hyperref[sec: LOS_DESC]{`outliers'}. With the mean \ac{los} of 3.015 days (standard deviation of 4.064 days), $\psi$ was set at $\psi=\lceil3.015\rceil=4$ days. Therefore the \ac{los} threshold ($\psi$) was set to at least 4 days hospitalised.
Throughout the remaining paper, a `long stay' refers to hospitalisation with \ac{los}$\ge$4 days, while a `short stay' denotes hospitalisation with \ac{los} < 4 days. The term `long-stay rate', used in this study, refers to the percentage of admissions that lasted for at least 4 days, given by the formula (\ref{losrate}).
\begin{equation} \label{losrate}
    \text{Long-stay rate}= \frac{\text{Number of admissions with \ac{los}} \geq 4}{\text{Total number of admissions}}
\end{equation}

\subsection*{Dataset Description}
\begin{figure}[t]
    \centering
     \includegraphics[width=\textwidth]{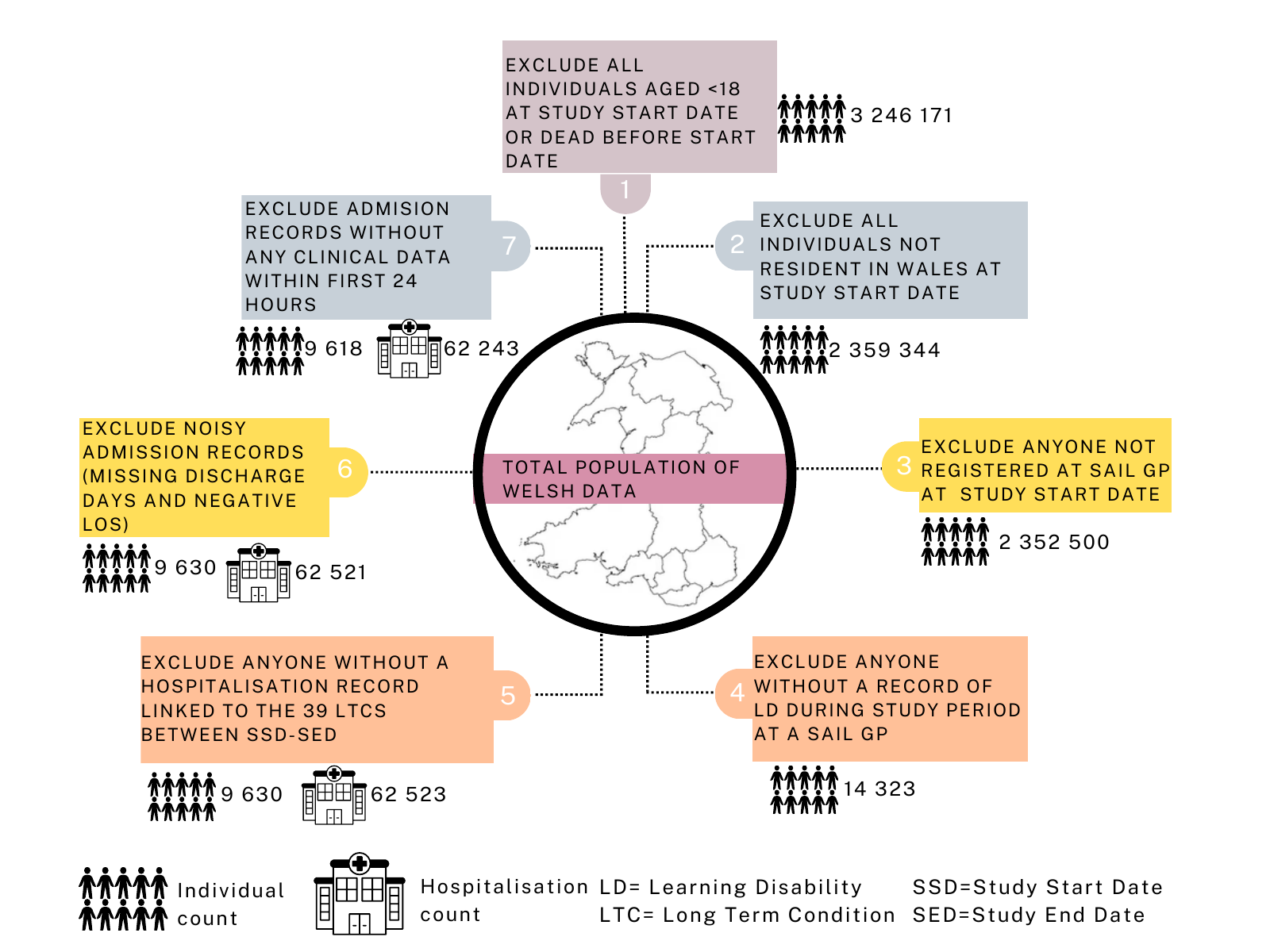}
         \caption{Consort Flow Diagram.}
         \label{fig:CohortDiag}
\end{figure}
\paragraph{\textbf{SAIL database.}} 
 This study utilised \ac{ehr} data sources of hospital admissions for patients with \ac{ld}, contained in the \ac{sail} Databank, the national \ac{tre} for Wales, enabling the use of anonymised individual-level, population-scale, linked data sources \cite{lyons2009sail,ford2009sail, rodgers2012protecting}. \ac{sail} partners with the \ac{nhs} and Welsh government to organise routinely collected longitudinal health and administrative data for $\sim$5 million Welsh residents, accessed securely under strict conditions compliant with the \ac{gdpr}. Specifically for this study the \ac{wlgp} data containing information on primary care \ac{gp} records, the \ac{pedw} containing information on secondary care inpatient hospitalisation admissions, the \ac{wdsd} containing patient demographic, residency, and registration history, and the \ac{adde} containing mortality records from the \ac{ons} were accessed. An \ac{alf} facilitates longitudinal anonymised linkage through all data sources in SAIL\cite{jones2014case}. Data captured within primary care is via read codes. The \ac{icd10} codes capture diagnosis, and \ac{opcs4} captures operations in hospital admissions.

\label{sec: inclusion_criteria}
\paragraph{\textbf{Inclusion and exclusion criteria.}} 
The study focused on Welsh residents aged 18 years or older identified with \ac{ld} during the period from January 1, 2000, to December 31, 2021, as depicted in Figure \ref{fig:CohortDiag}: Exclusions were implemented for individuals younger than 18 years, those not residing in Wales, individuals not registered with a \ac{sail} \ac{gp} at the study start date, and those without \ac{ld}. This process led to a total of 14\,323 unique patients during the data extraction phase. Subsequently, patients without any hospital admission records between the study start and end dates were omitted. The resulting admissions data were then associated with 39 \acp{ltc} (Supplementary Table \ref{tab: ltc_list}), documented primarily through Read codes in primary care and \ac{icd10} codes in secondary care \cite{benson2011history, cartwright2013icd}. The dataset comprised 62\,523 unique admissions from 9\,630 \ac{ld} patients, spanning the period from January 2000 to December 2021. Exclusions were applied to records with missing discharge dates and negative length of stay (\ac{los}), resulting in a cohort of 62\,521 unique admissions from 9\,630 \ac{ld} patients. Finally, admissions with no data within the first 24 hours were excluded, yielding the final cohort of 9\,618 unique patients (4\,929 males, 4\,689 females) with 62\,243 unique admissions (32\,275 males, 29\,968 females). Refer to Table \ref{Tab:cohort_desc} for further details on demographic distribution. Additionally, Table \ref{Tab:variables} provides a comprehensive list of all variables extracted for each patient across their admissions.

\subsection*{\textbf{Dataset Demographic Description}}

\paragraph{\textbf{Age group.}} Age was categorised into seven groups, as shown in Table \ref{Tab:cohort_desc}. Patient and admission counts generally followed a normal distribution across age groups for both sexes, with fewer admissions in ages above $60-69$ and below $30$. Possible factors for lower admission rates in ages above $60-69$ include home care reducing hospital need and mortality (see Table \ref{tab: Mortality_cohort} for mortality statistics on this study cohort). The statistics on mortality obtained in this study are consistent with the life expectancy statistics (66 and 67 years for \ac{ld} males and females, respectively)\cite{NHSDigital_2020}. Notably, ages above $60-69$ exhibited higher rates of long stays, similar for both sexes.

\paragraph{\textbf{\acs{wimd}.}}Patients’ socioeconomic status was described using the \ac{wimd} version 2019 for Wales\cite{Statistics_for_Wales}, an area-level weighted index across seven deprivation domains assigned based on the individual’s residence using their \ac{lsoa} version 2011, with each \ac{lsoa} representing an area of $\sim 1\,500$ people.  The seven deprivation domains include income, employment, health and disability, education skills and training, barriers to housing and services, living environment, and crime. \ac{wimd} quintiles were used, categorising the area of residence from 1 (most deprived) to 5 (least deprived) as shown in Table \ref{Tab:cohort_desc}. Patients without an \ac{lsoa} or associated \ac{wimd} quintile were grouped as ‘Unknown’. The highest representation of hospital admissions was from the most deprived quintile (quintile 1), comprising $\sim 28\%$ of admissions for males and $\sim 27\%$ of admissions for females, indicating heightened hospital demand with increasing deprivation. This aligns with research done on the general population, attributing higher admission rates in impoverished areas to factors such as inadequate allied healthcare and local resources and potential underuse of community medical resources\cite{cournane2015social, aljuburi2013socio}.  Table \ref{tab: Mortality_cohort} also shows higher mortality rates with increasing deprivation for the study cohort. This finding is consistent with a recently published study on the impact of deprivation on mortality \cite{Tyrer2021-zw}.

\paragraph{\textbf{Ethnic group.}} 
 Ethnic groups were classified using the \ac{ons}, UK categorisation. The methodology by Akbari et al.\cite{Akbari2022} was used to extract and harmonise the ethnic group details from the various data sources available to the project. The cohort was not uniformly represented (i.e. unbalanced) in terms of ethnic groups: $\sim 73\%$ of patients ($\sim 79\%$ of admissions) and $\sim 74\%$ of patients ($\sim 80\%$ of admissions) were from the `White' group for the male and the female sex respectively (see Table \ref{Tab:cohort_desc}). Within the cohort, $\sim 25\%$ of males and $\sim 24\%$ of females had no ethnic group records, classified as `Unknown' (19.24\% and 17.53\% of male and female admissions, respectively). The remaining ethnic groups (Black, Asian, Mixed, Other) each represented $<3\%$ of patients and admissions for both sexes. Long stay rates varied widely across the ethnic groups, with the `Black' group having the highest rate (M: 52\% and F: 65.9\%; M denotes male sex and F denotes female sex) and the `Other' group the lowest (M: 34.2\% and F: 30.4\%). A similar finding was observed in a study on inpatient discharges for the general patient population in the United States which revealed that Black patients had significantly longer \acp{los} compared to other groups\cite{Ghosh2021}.

{\subsection*{What were the primary conditions for hospital admissions of patients with \ac{ld}?}  The \ac{pedw} data includes a variable `diag\_num'- a number used to identify the position of diagnosis assigned to a patient during a unique admission. Value `1' relates to the primary \ac{icd10} Diagnostic Code which is the main condition treated or investigated during the relevant episode of healthcare. Values > 1 relate to secondary \ac{icd10} diagnostic codes. 
 Analysis of 18\,541 admissions of men and 17\,587 admissions of women with \ac{ld} from the last 10 years of study duration (January 2011-December 2021) showed cancer as the primary condition for admission, with 1\,703 (9.2\%) male and 2\,149 (12.2\%) female admissions. The subsequent top primary conditions differed by sex respectively, as shown in Figure \ref{fig:Condition}: epilepsy, chronic pneumonia, chronic airway diseases, and mental illness for males; versus chronic kidney disease, epilepsy, chronic pneumonia, and chronic airway diseases for females. Tables \ref{tab: primary_cond_male} and \ref{tab: primary_cond_female} detail the top 10 primary conditions for both sexes including admission counts, patient numbers, and admission rates per patient. The high \ac{std} values in the admission rate for some conditions indicate the variation in admission rates per individual, as some patients may be admitted more times than others. The admission rates are strongly influenced by the number of \acp{mltc} across individuals.

\subsection*{What were the prevalent conditions in hospital admissions for patients with \ac{ld}?}
 Further 2011-2021 admission analysis was made on the \ac{pedw} data for the prevalent conditions treated or investigated during unique hospital admissions. This includes both primary and secondary diagnostic codes. The analysis revealed epilepsy as the most commonly treated \acp{ltc} during admissions, present in $29.4\%$ of male and $24.1\%$ of female admissions (Figure \ref{fig:Condition}). The next most prevalent conditions in the male group were diabetes, chronic airway diseases, mental illness, and cancer, while the female group had chronic airway diseases, diabetes, thyroid disorders, and mental illness as the next most prevalent conditions after epilepsy. Figure \ref{fig:Condition} also indicates slightly higher rates of epilepsy, diabetes, and mental illness admissions in males compared to females, and higher rates of chronic airway diseases in females compared to males. Supplementary Table \ref{tab:2011_2021_conditions} provides the ranking of common \acp{ltc} treated during hospital admissions for the stated period.

\begin{figure}
     \centering
     \begin{subfigure}[b]{0.48\textwidth}
         \centering
         \includegraphics[width=\textwidth]{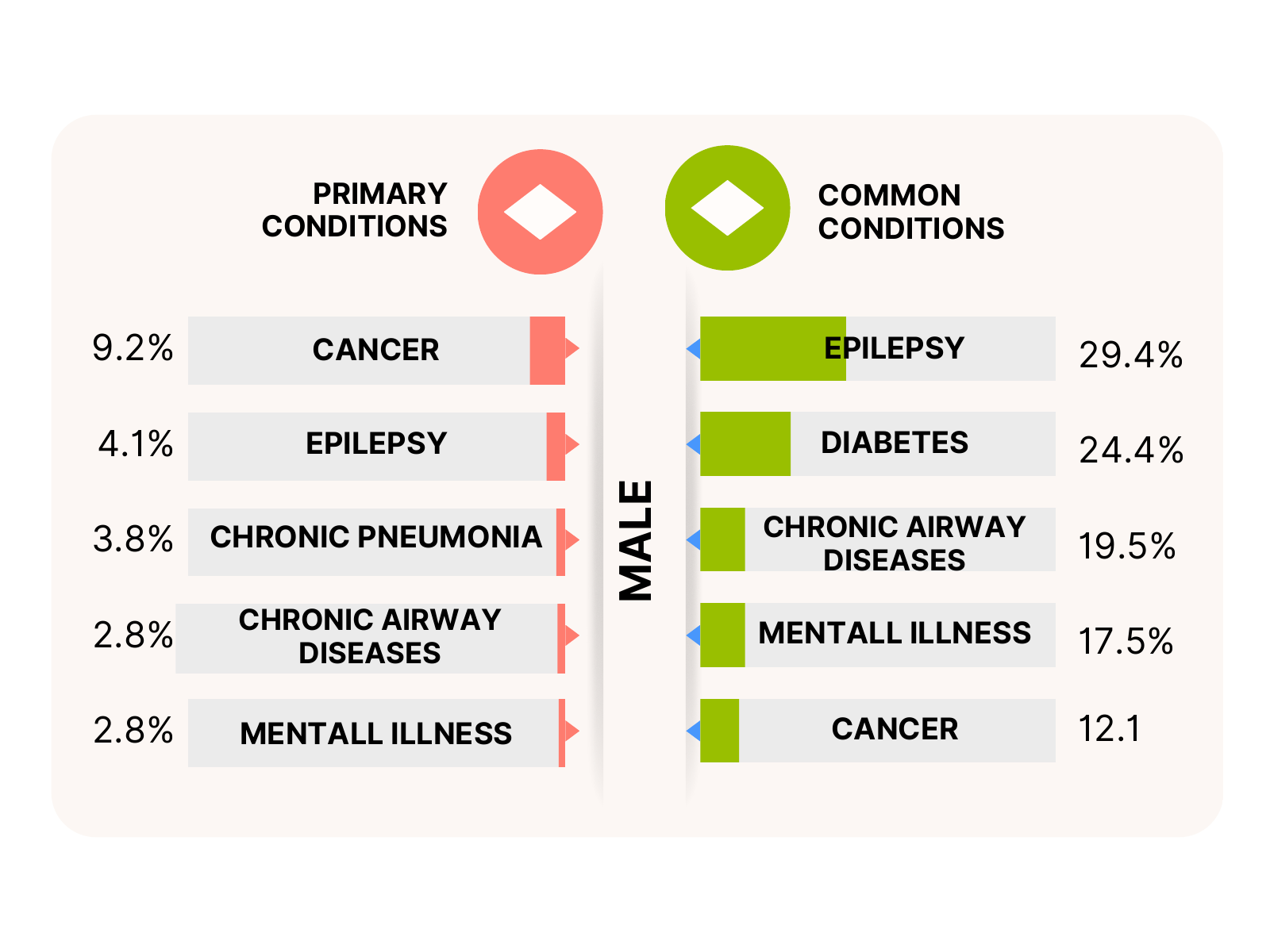}
         \caption{}
         \label{fig:Condition_male}
     \end{subfigure}
     \hfill
     \begin{subfigure}[b]{0.48\textwidth}
         \centering
         \includegraphics[width=\textwidth]{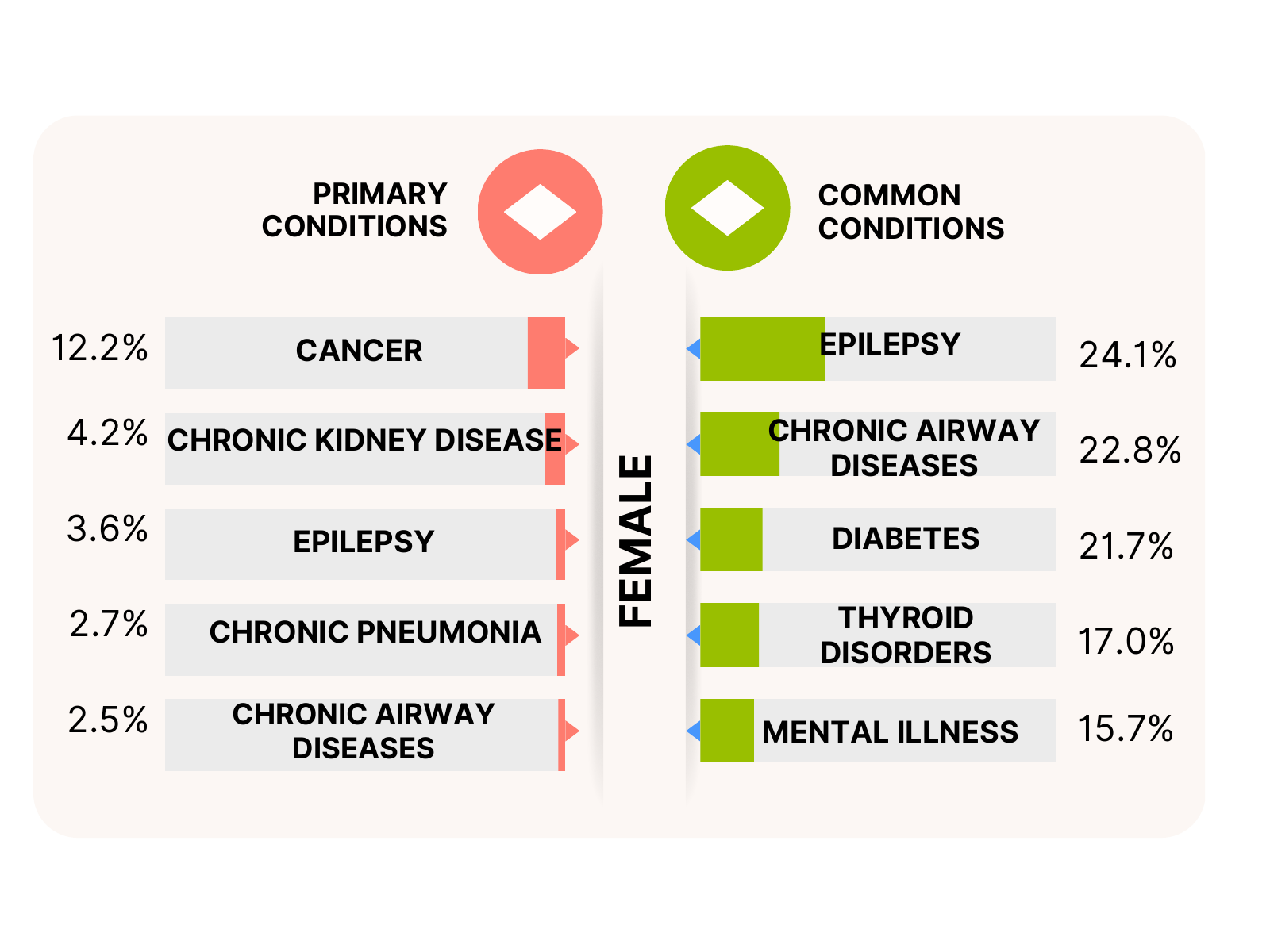}
         \caption{}
         \label{fig:Condition_female}
     \end{subfigure}
     \caption{ {The top five primary conditions and top five prevalent (common) conditions treated or investigated during hospitalisations for the (a) males and (b) females with \ac{ld} and \acp{mltc}. The primary conditions indicate the main condition treated or investigated during the admission, and the prevalent conditions indicate the most frequently treated or investigated conditions (includes both primary and secondary diagnostic codes) across all unique hospitalisations of patients with \ac{ld} and \acp{mltc}.}}
     \label{fig:Condition}
     \end{figure}

\subsection*{What is the distribution of the \ac{los} across patients?}
\label{sec: LOS_DESC}
The \ac{los} for all extracted patients' hospitalisation records from birth ranged from a minimum of 0 days (indicating no overnight stay during admission) to over 5\,000 days. The combined male and female group comprised 67\,377 admissions. The median hospitalisation was $Q_2=2$ days with first and third quartiles of 0 days ($Q_1$) and 7 days ($Q_3$), respectively, giving an \ac{iqr} of $Q_3-Q_1$. This is illustrated using the box plot in Figure~\ref{fig: box_plot}, which includes lower and upper whiskers. All admissions with \ac{los} days below the lower whiskers or above the upper whiskers are described as \textit{outliers}. The lower whisker was the smallest \ac{los} value in days greater than $Q_1-1.5\cdot \ac{iqr}$, equal to $Q_1$ ($0$ days). The upper whisker was obtained as the largest \ac{los} value in days less than $Q_3+1.5\cdot \ac{iqr}$, obtained at 17 days. 

\paragraph{Outliers.}\label{sec: outliers}   For the entire hospitalisation records of patients from birth, outliers (admissions with LOS > 17 days) were further analysed to understand patterns among admissions with the most extended stays. Hence, quartile values from boxplots of outlier records were obtained: $Q_1=24$ days, $Q_2=36$ days, $Q_3=66$ days, and an \ac{iqr} of 42 days. The upper whisker was the largest hospital stay under $Q_3+1.5\cdot \text{IQR}$, obtained at 129 days. Consequently, admissions with LOS $\geq$ 129 days were numerically evaluated. This amounted to 934 unique admissions. The majority of these admissions are related to mental illness. Generally, hospitalisations with very long \ac{los} are common for mental health admissions, especially for those with challenging behaviours and autism/personality disorders posing safety risks \cite{dorning2015focus,Siddiqui2018-pt}.
Figure \ref{fig: los129days} further illustrates these findings, depicting the common conditions for stays $\geq$ 129 days.  For these admissions with \ac{los} $\geq$ 129, the most common condition was mental illness and epilepsy, followed by diabetes, dementia, and cerebral palsy. Table~\ref{tab:Admission129days} provides a full breakdown of related \acp{ltc} for admissions with \ac{los} $\geq129$ days. Figure \ref{fig: los129days_age} further depicts, for all admissions with \ac{los} $\geq 129$ days, the age distribution for admissions involving mental illness and without mental illness. For the latter, age was normally distributed and skewed towards the older age groups. Given the study cohort includes patients with \acp{mltc}, most patients admitted for a primary condition also experienced hospital episodes associated with other secondary conditions.

\begin{figure}[t]
     \centering
     \begin{subfigure}[b]{0.45\textwidth}
         \centering
         \includegraphics[width=\textwidth]{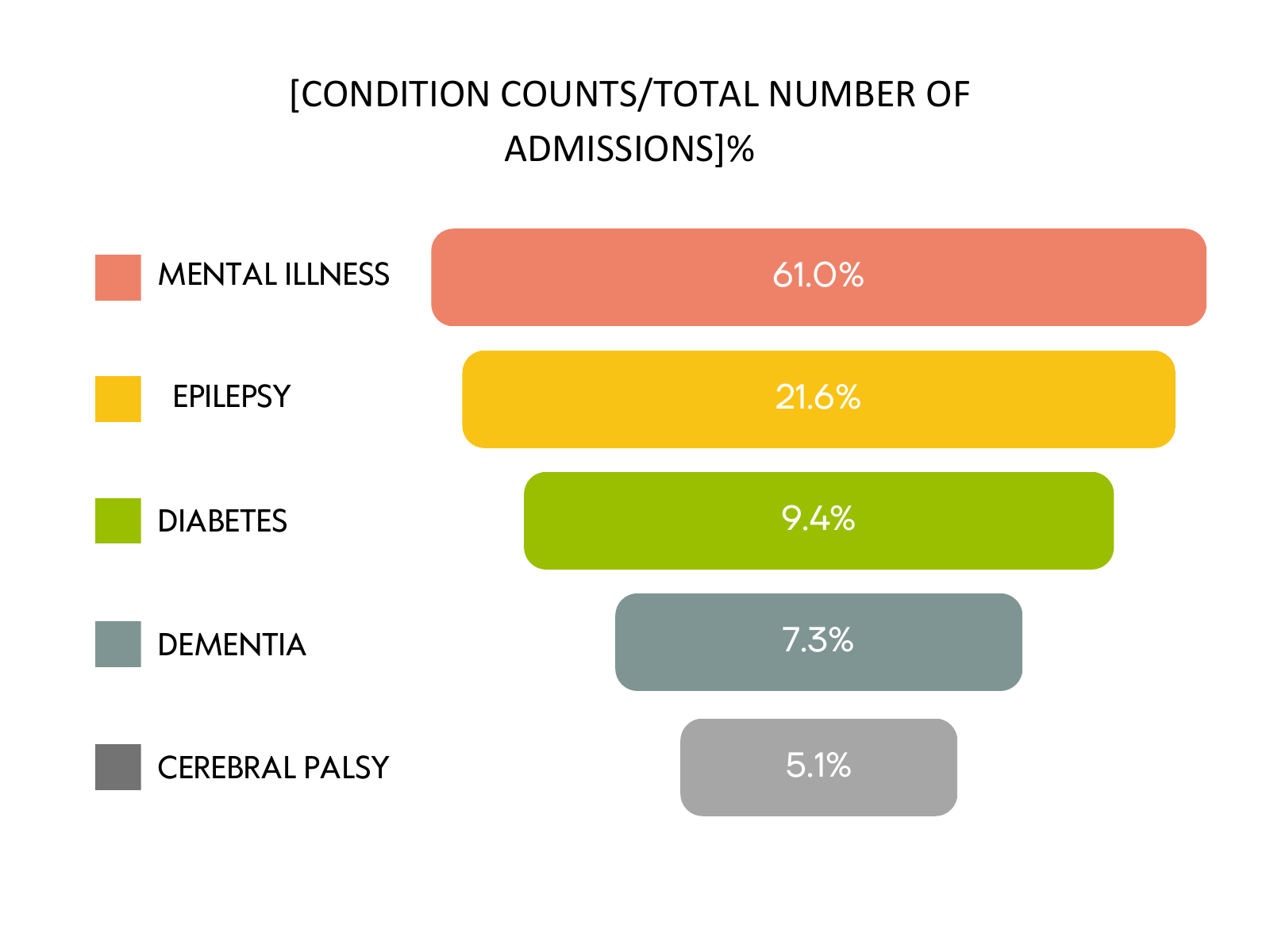}
         \caption{}
         \label{fig: los129days}
     \end{subfigure}
     \hfill
     \begin{subfigure}[b]{0.45\textwidth}
         \centering
         \includegraphics[width=\textwidth]{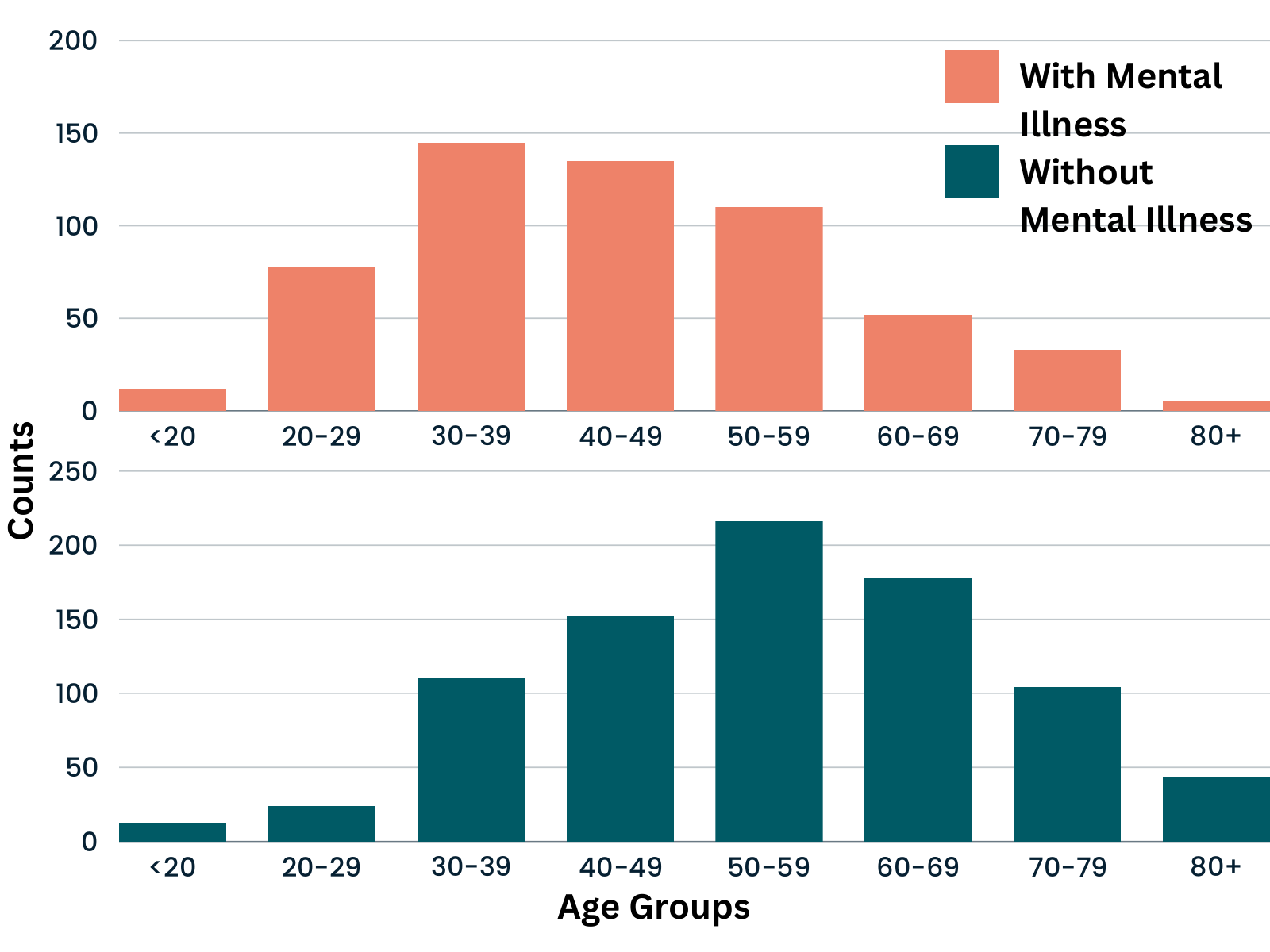}
         \caption{}
         \label{fig: los129days_age}
     \end{subfigure}
     \caption{Distribution of patients with \ac{los} $\ge$ 129 days. (a) The $4$ most common conditions and (b) age distribution of patients with and without mental illness.   }\label{fig:los_above_129}
     \end{figure}
     
\subsection*{What factors were prevalent in hospital stays $\geq 4$ days?}
 Analysis was conducted prior to the machine learning (ML) predictions to identify common trends among admissions with \ac{los} $\geq 4$ days within the study duration. The extracted data was analysed by sex (male and female) across all admissions in the study duration. The one-sample chi-square and binomial tests were applied to all categorical features to test the null hypothesis that the variable categories occur with equal probability. Results stated in Table \ref{Tab: chisquareBinomial} rejected this hypothesis for all categorical variables.

Excluding `unknown' groups, Table~\ref{tab:DistLOSClass} revealed $\geq 4$ day stays were predominantly in patients aged $\geq 50$ years, from more deprived socioeconomic quintiles, obese, and less physically active, compared to patients with short stays (\ac{los} < 4 days). This observation was similar for both sexes.  In addition, females with prescribed antipsychotic, antidepressant, or anti-manic/anti-epileptic medications were also seen to have more long stay admissions with \ac{los} $\geq 4$ days, compared to female admissions with \ac{los} < 4 days. Other factors influencing $\geq 4$ day stays include the primary long-term condition for admission and  \ac{mltc} counts, which can increase hospital episodes in a single admission. Patients in this study cohort had between $1$ and $21$ comorbid conditions per person. Figures~\ref{fig: avrg_con_count} and \ref{fig: avrg_con_count_gndr} illustrate the average \ac{mltc} counts by age group for combined and separate sexes, showing a linear rise with age. Further analysis was also conducted on previous hospitalisation data that included the number of previous admissions and hospital episodes, cumulative hospital days from past admissions, and the number of \acp{mltc} from previous admissions (See Table \ref{Tab:variables}). Patients with a higher number of \acp{mltc}, cumulative hospital days from past admissions, and long-term conditions treated in previous admissions were more likely to be hospitalised for $\geq$ 4 days compared to patients with stays less than 4 days. Patients with more frequent prior admissions and higher counts of previous hospital episodes tended to have short \ac{los} in their current admission. This is illustrated in Figures \ref{fig: pointMale}-\ref{fig: pointFemale2}. Those patients with \ac{los} <1 day were most likely to be attending routine appointments rather than emergency visits.

\begin{figure}
     \centering
     \begin{subfigure}[b]{0.45\textwidth}
         \centering
         \includegraphics[width=\textwidth]{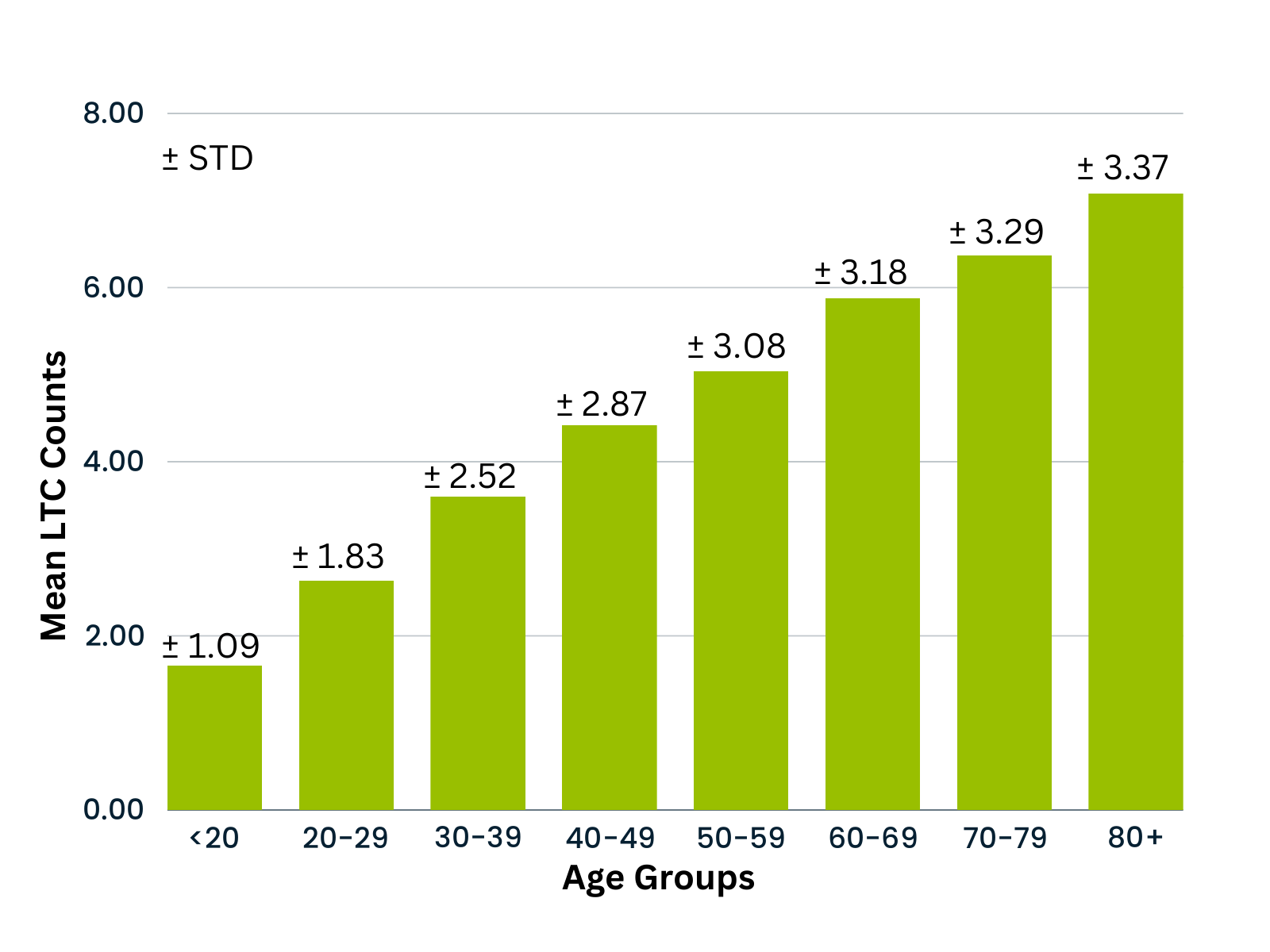}
         \caption{}
         \label{fig: avrg_con_count}
     \end{subfigure}
     \hfill
     \begin{subfigure}[b]{0.45\textwidth}
         \centering
         \includegraphics[width=\textwidth]{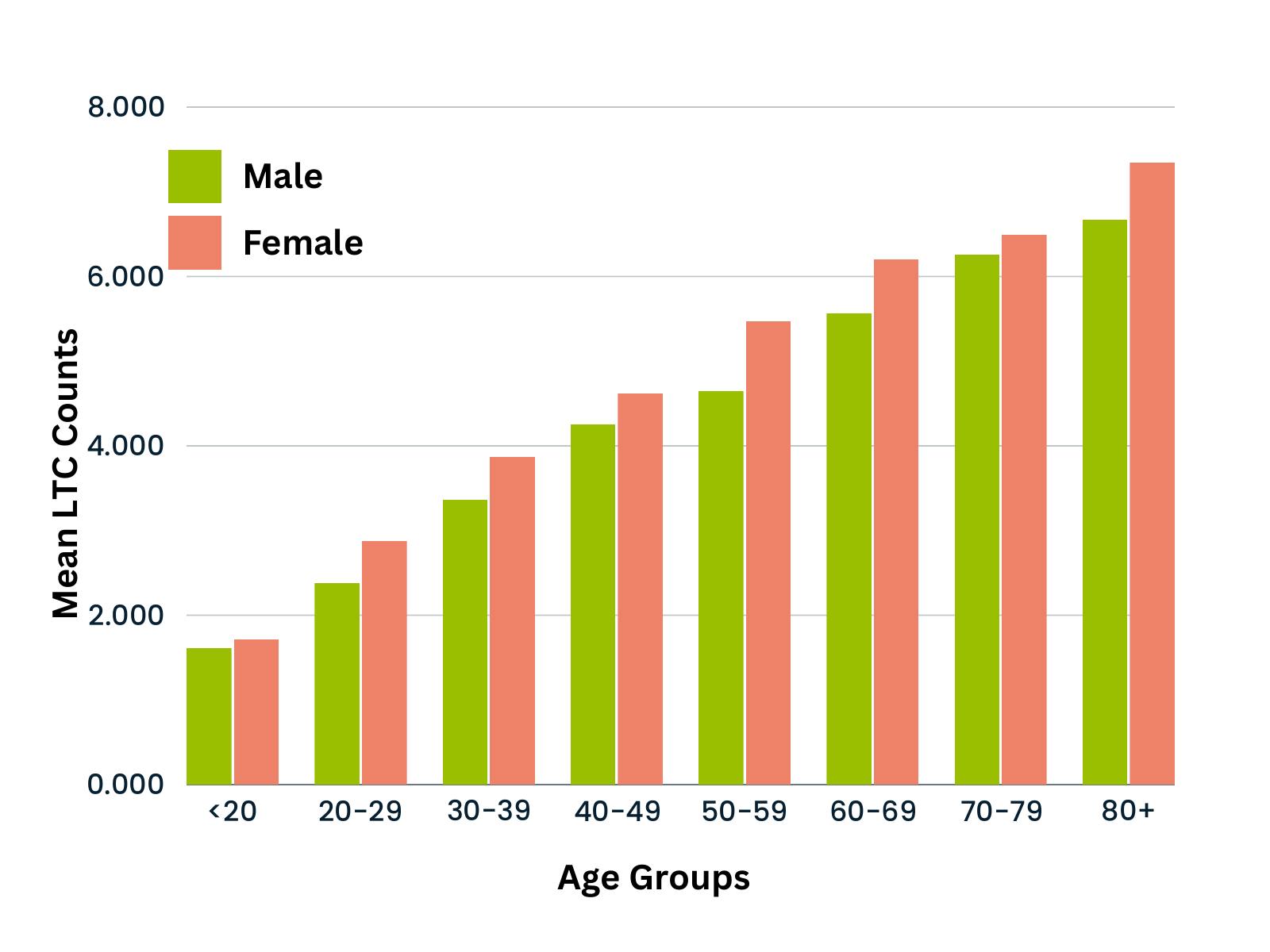}
         \caption{}
         \label{fig: avrg_con_count_gndr}
     \end{subfigure}
     \caption{Distribution of mean \ac{mltc} count across age groups for (a) combined sexes, and (b) individual sexes.  }\label{fig:avrg_con_count}
     \end{figure}

\subsection*{Is there a co-linear relationship between numerical features in the dataset?}

A correlation test explored relationships between the numerical features (columns) in the dataset. Correlation analysis aids \ac{ml} model building by detecting redundant inputs, simplifying interpretations, and improving target prediction performance. Table \ref{Tab:variables} describes all features for predicting \ac{los}. The Kolmogorov-Smirnov test assessed feature normality before selecting an appropriate correlation analysis. As shown in Table \ref{Tab:Normality}, all features had non-normal distributions. Consequently, the non-parametric Spearman's rank correlation coefficient was evaluated. Figures \ref{fig: male_corr} and \ref{fig: female_corr} illustrate no statistically significant associations between any input and the outcome (\textit{LOSClass}) for male and female cohorts. However, several input variable pairs exhibited collinearity, with correlation coefficients exceeding $\pm$0.5 for both groups. This study found that for the highly correlated variable pairs, their interaction provided useful information to the model for the cohort examined.

\subsection*{Can machine learning predict the \ac{los} across hospital admissions?}
Several classification models were developed for the prediction of the \ac{los} using the selected features described in Table \ref{Tab:variables}. The models were designed to predict whether a patient's admission would have \ac{los} $<$4 days or $\geq$4 days. Predictions were made using available patient data up until the first 24 hours of admission. The classification performance of each model is presented in Table \ref{Tab:classifiers}. Additionally, Figures \ref{fig:roc_male} and \ref{fig:roc_female} depict the \ac{roc} curve for each classifier, along with the \ac{auc} and optimal \ac{roc} point values. An optimal \ac{roc} point refers to the point on a \ac{roc} curve that provides the best balance between the \ac{tpr} and the \ac{fpr} for a given classification model.
\mbox{}\\

\paragraph{Classification performance.} In this study, the \ac{histgboost} and \ac{rf} classifiers showed the best performance compared to the other models. The \ac{histgboost} classifier achieved higher \ac{auc} (M: 0.771, F: 0.773) and balanced accuracy (M: 0.701, F: 0.705). This was followed by the \ac{xgboost} classifier (\ac{auc} = 0.763 (M), 0.761 (F); balanced accuracy = 0.695 (M), 0.692 (F)) and the \ac{rf} classifier (\ac{auc} = 0.759 (M), 0.756 (F); balanced accuracy = 0.690 (M), 0.689 (F)). Regarding the \ac{fnr} (indicating patients predicted to be discharged early when a longer (i.e., $\geq$ 4 days) hospital stay is required), the \ac{rf} classifier returned lower values compared to the other models (\ac{fnr} = 0.224 (M), 0.229 (F)). The \ac{fnr} value was approximately 7\% lower for the \ac{rf} classifier than the \ac{histgboost} model for the male group and $\sim 6\%$ lower than the \ac{xgboost} model for the female group.  The \ac{rf} demonstrated optimal performance for both the male and female groups as shown by its low \acp{fnr} and high balanced accuracy. The low \ac{fnr} indicates fewer missed cases where patients require longer hospitalisation $\ge$ 4 days, while the balanced accuracy shows high predictive performance. 

\begin{figure}
     \centering
     \begin{subfigure}[b]{0.47\textwidth}
         \centering
         \includegraphics[width=\textwidth]{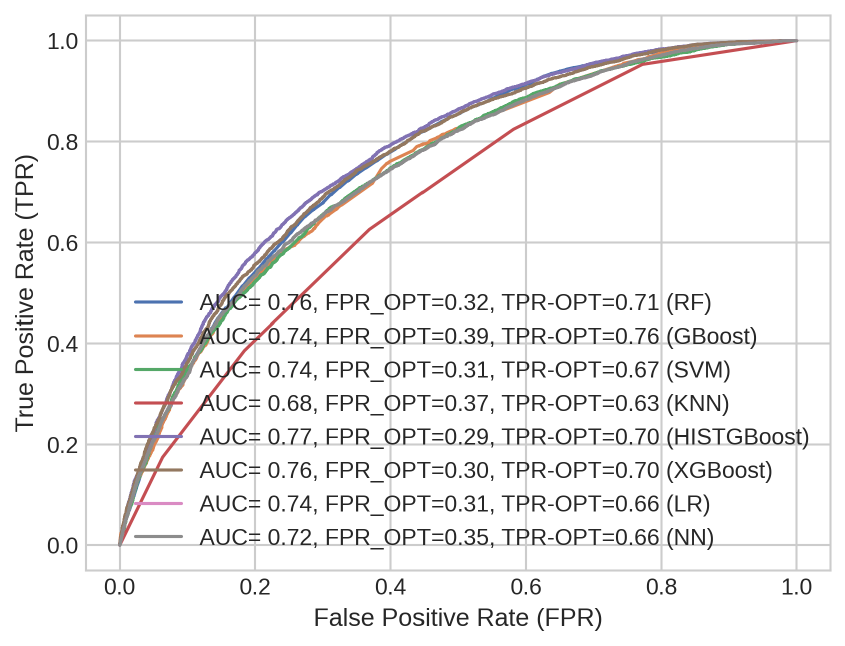}
         \caption{}
         \label{fig:roc_male}
     \end{subfigure}
     \hfill
     \begin{subfigure}[b]{0.47\textwidth}
         \centering
         \includegraphics[width=\textwidth]{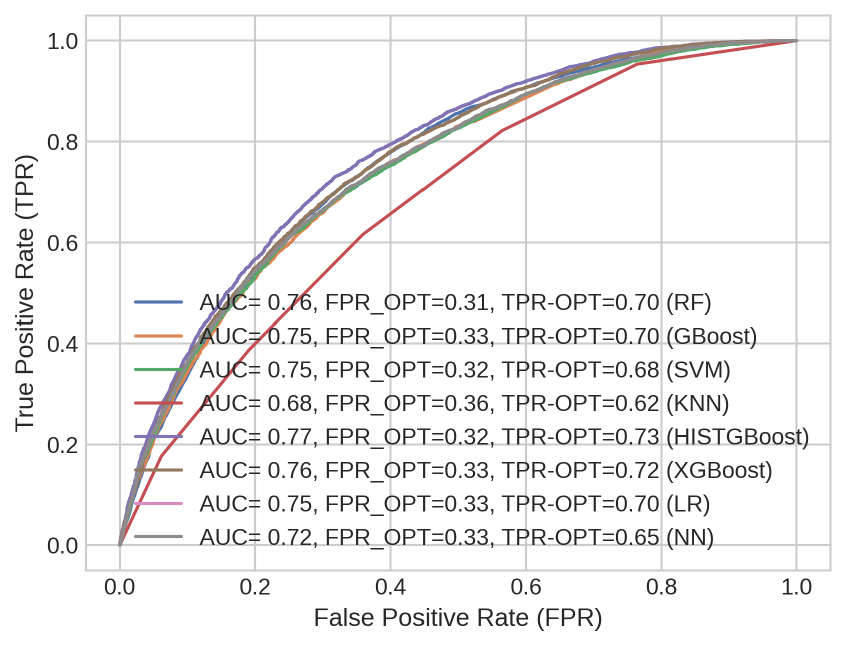}
         \caption{}
         \label{fig:roc_female}
     \end{subfigure}
     \caption{\ac{roc} curves across models indicating their optimal points (\ac{fpr} and \ac{tpr}) for (a) male and (b) female cohort.   }
     \label{fig:ROC}
     \end{figure}

\paragraph{Is the performance of the best model consistent across different train/test combinations?}

The selected \ac{rf} model was further evaluated using 10 randomly selected distinct train/test combinations to assess its performance across different samples. The mean performance (with \ac{std}) across all 10 iterations is provided in Table \ref{tab: Iterations_performance} for the male and female cohorts. The mean value obtained for each parameter across sexes is similar to the performance detailed in Table \ref{Tab:classifiers} for the male and female cohort, and the \ac{std} is less than 0.007 for all parameters as shown in Table \ref{tab: Iterations_performance}. This suggests the \ac{rf} model achieves consistent performance across different data samples, indicating it has good generalizability and is not overfitting on the given training set. The low \ac{std} also shows the model yields reliable and stable predictions.

\subsection*{How did the best model perform across ethnic groups?}
The \ac{rf} model which showed optimal performance was further evaluated to address potential performance bias associated with the ethnic groups. Tables \ref{tab:rf_ethnicityMale} and \ref{tab:rf_ethnicityFemale} provide an overview of the \ac{rf} model's performance segmented by ethnic groups for the male and female groups respectively.

\paragraph{Males.} For males, the `Other' ethnic group exhibited the highest \ac{fnr} at 33.3\% followed by the `Black' group with \ac{fnr}  of 27.3\%. The `Black' group had the lowest balanced accuracy at 59.7\%. In contrast, the `Asian' group demonstrated the lowest \ac{fnr} of 19.5\%, and the `White' group had the highest balanced accuracy of 69.2\%. Overall, there was an approximate 13.8\% range in the \ac{fnr} and a 9.5\% range in balanced accuracy across the considered ethnic groups. The model's performance for the `White' ethnic group closely resembled the overall model performance (detailed in Table \ref{Tab:classifiers} for the \ac{rf} model), most likely because the majority of the training data (79\%) originated from the `White' group. The `White' group constituted 79\% of the entire male cohort. 

\paragraph{Females.} For females, the model's performance for the `White' group closely matched the overall classifier performance, attributable to the high representation of the `White' group (80.6\% of training, 80\% of total data). The model underperformed in predicting \ac{los} for the `Black' group with the lowest balanced accuracy (66.7\%), and highest \ac{fpr} (50\%). Both the `Black' and `Other' groups had the lowest representation in the extracted data (0.15\% and 0.19\% respectively), suggesting insufficient data for optimal modelling. The `Other' group, despite having a low representation of admission records, demonstrated improved performance with the lowest \ac{fnr} (11.1\%) and the highest balanced accuracy (77.8\%). This suggests the classifier was able to effectively model outcomes for the `Other' ethnic group given the available training samples. Overall, for females, there was a performance range in the \ac{fnr}, and balanced accuracy of approximately 12\%, and 11.1\% respectively across ethnic groups.

\subsection*{Can consistency in \ac{ml} performance be achieved across ethnic groups?}
In the context of this study, an ideally fair model would exhibit consistent performance in predicting the \ac{los} across ethnic groups. To improve the fairness of the \ac{los} prediction models, two bias mitigation algorithms: one post-processing (threshold optimizer) and one in-processing (reductions approach with exponentiated gradient) were empirically investigated. Both approaches aimed to minimise the range of each performance metric across the ethnic groups. Each bias mitigation algorithm was assessed and compared to the unmitigated model to determine its effectiveness.

\paragraph{In-Processing (reductions with \ac{eg}).} 
Tables \ref{tab:rf_comp_Male} and \ref{tab:rf_comp_Female} overview the \ac{eg} reductions performance. This approach worked best in reducing the \ac{fnr} for the males by 9\% compared to the unmitigated model. However, the performance range was not optimized for the \acp{fpr} and balanced accuracies for both sexes (See Tables \ref{tab:Diff_female}, \ref{tab:Diff_male}). 

\paragraph{Post-processing (threshold optimizer).}
Tables \ref{tab:rf_comp_Male} and \ref{tab:rf_comp_Female} depict the performance of the threshold-optimized model. For females, compared to the unmitigated model, the threshold optimizer reduced the \ac{fnr} range across ethnic groups by 3.6\% (See Table \ref{tab:Diff_female}). However, the range in the \ac{fpr} and balanced accuracy increased by 9.8\% and 5.6\% respectively across ethnic groups compared to the unmitigated model. Specifically, the optimizer did not substantially improve the model fairness across the female group. The threshold optimizer yielded better performance in males, reducing the range in the \ac{fpr}, and the balanced accuracy values across ethnic groups by 7.4\% and 1.8\% respectively. However, the range in \ac{fnr} increased slightly by 1.7\% for the threshold optimizer compared to the unmitigated model. Particularly, the unmitigated classifier had a lower \ac{fnr} range across the ethnic groups than the optimizer for males (See Table \ref{tab:Diff_male}).

In summary, although the fairness goal of equal performance across ethnic groups was not fully met, the post-processing threshold optimizer approach was more effective at improving performance uniformity across ethnic groups compared to the reduction with exponentiated gradient.

\section*{Discussion}
\label{sec: discussion}
\begin{tcolorbox}[colback=gray!10,colframe=gray!40!black,title=Overview of findings]
\begin{itemize}[noitemsep]
\item Analysed electronic health records of 9\,618 patients with \ac{ld} and \acp{mltc} in Wales, examining 62\,243 hospital admissions.
\item Cancer was the top primary condition for hospital admissions in both males and females with \ac{ld}. Epilepsy was the most commonly co-occurring condition across all admissions between January 2011 and December 2021.
\item Hospital stays lasted a median of 2 days, with an \ac{iqr} of 0-7 days. Stays exceeding 129 days were commonly related to mental illness.
\item Common factors associated with patients with hospital stays $\geq$ 4 days included: age $\ge$ 50 years, higher socioeconomic deprivation, obesity, low physical activity; and a higher number of \acp{mltc}, cumulative hospital days from past admissions, and long-term conditions treated during previous admissions.
\item A random forest machine learning model achieved an \ac{auc} of 0.759 (males) and 0.756 (females) in predicting the length of stay using data up to the first 24 hours of admission.
\item Before bias mitigation, the model demonstrated performance discrepancies across ethnic groups. Two bias mitigation approaches were tested, with the threshold optimizer outperforming the reductions approach in minimising some performance differences across groups.
\end{itemize}
\end{tcolorbox}

\noindent \ac{nhs} England has a Reducing Length of Stay (RLoS) program\cite{NHSLos} that aims to improve patient care and experience by avoiding unnecessary delays in hospital discharges. Prolonged hospital stays can increase patient risk of infections, falls, sleep deprivation, and physical/mental decline. In 2018, a national ambition was set to reduce long hospital stays (21+ days) by 25-40\% by 2020. The RLoS program was established to provide strategic direction and support local delivery of this goal. Reducing stays improves patient outcomes, increases urgent and emergency care (UEC) system capacity, and frees up beds. The RLoS program has launched a national campaign to communicate the importance of timely discharges, established discharge tracking to identify delays, and published guidance on managing the length of stay. Individuals with \ac{ld} often experience poorer health outcomes compared to the general population\cite{carey2016health}, which also impacts their hospital stays. Therefore, accurately predicting the length of stay can inform care planning to optimise resource allocation and prevent unnecessarily long stays or premature hospital discharge \cite{stone2022systematic}.\\

\noindent Our study proposes \ac{ml} models to predict the hospital \ac{los} for patients with \ac{ld} while ensuring fairness across ethnic groups. As \ac{ml} increasingly guides healthcare decisions, ensuring algorithmic fairness across protected attributes such as ethnic groups is crucial.  The extracted data exhibited imbalanced representation across ethnic groups - records for underrepresented minority patients were limited, while ethnic group information was missing entirely for many patients. This resulted in notable performance discrepancies for the \ac{rf} model across groups, before the application of bias mitigation strategies. Specifically, \acp{fnr} (failing to predict long stays) varied widely and were worse for minority females. Insufficient minority representation in the training data is a primary factor that can negatively impact model performance for these groups. Therefore, bias mitigation techniques were employed to improve uniformity in machine learning performance across groups.  Increasing the completeness of data quality and consistency of recording of ethnic group data could result in fairer modelling, providing more examples for the models to learn patterns specific to minority populations. 

\section*{Methods}
\label{sec: methods}

\label{sec: data processing}

\subsection*{Stage I: Extracting and preparing the data for ML}

\paragraph{Step 1: Extracting the inputs.} The model utilises a dataset of 9\,618 patients with \ac{ld} (see Table \ref{Tab:cohort_desc}) and a total of 62\,243 hospital admission records to predict each patient's \ac{los} (i.e. target variable). For each admission, data relating to the patient's health up until the first 24 hours of admission were extracted to be applied as inputs to the \ac{ml} models. As shown in Table \ref{Tab:variables}, this data includes variables describing: the patient's lifestyle and history (\ac{bmi}, smoking, alcohol consumption, physical exercise, autism); prior 1-year and 3-year hospitalisation data (previous admissions and hospital episodes, cumulative hospital days from past admissions, and the number of \acp{mltc} from previous admissions); prescribed antipsychotic, antidepressant, and anti-manic/anti-epileptic medications (Table \ref{tab: med_list} provides the medications list); and other variables from first 24 hours of current admission-indicating the prevalence of the 39 \acp{ltc} (see Supplementary \ref{tab: ltc_list} for the full list of \acp{ltc}). Age group was also included as an input into the model (i.e. as a predictor) because \ac{mltc} counts increase with age as shown in (Fig. \ref{fig:avrg_con_count}) and impact on the length of stay (Fig. \ref{fig:los_above_129}). 

\paragraph{Step 2: Extracting the target variable.} The \ac{los} variable indicating the number of hospital days was replaced with a binary target variable ({LOSClass}) for machine learning classification purposes, as follows: values with a \ac{los} $\geq$ 4 days where replaced with 1, and values with \ac{los} <4 days where replaced with 0 as indicated in Table \ref{Tab:variables}. 

\subsection*{Stage II: Preprocessing the inputs for \ac{ml}}
\paragraph{Step 1: Data preprocessing.} 
Table \ref{Tab:variables} describes the \ac{ml} model input and target variables, and how they were preprocessed for \ac{ml} training and analysis. Table \ref{Tab:variables} also shows the demographic variables and these were not applied as inputs to the \ac{ml} model (except age group) but were utilised for describing the cohort and for bias analysis of \ac{ml} model performance.

\paragraph{Preprocessing longitudinal risk factors.}
\Ac{bmi}, alcohol consumption, smoking, physical activity, and medications are longitudinal variables, gathered, and coded by \acp{gp} or practice nurses. These risk factors change over time across patients. Hence, the variable {`BMI'} was coded categorically using the patient's \ac{bmi} value documented closest to the admission date. `ALCOHOL\_HISTORY' and `SMOKING\_HISTORY' were coded categorically based on alcohol and smoking intake history, respectively, up to admission (see Figures \ref{fig: alcohol} and \ref{fig: smoking}). `PHYSICAL' was coded categorically using the patient's physical activity status (i.e., if the patient engages in light or regular exercise) documented nearest admission. The `MEDICATIONS' variable was coded as binary for intake history at each admission and assumed lifetime use from the first prescription.

\paragraph{\textbf{Missing values.}} Missing values for categorical variables were classified into `unknown' categories for their respective variables, allowing models to utilise the partial information from observations with missing data rather than discarding or imputing them during preprocessing. There were no missing data for the numerical variables.

\paragraph{Step 2: Convert categorical variables to one-hot encoded variables.} \ac{ml} models require numerical inputs/target variable. To enable this, the categorical input variables were converted into numeric representations via one-hot encoding. The steps for one-hot encoding a categorical variable using the variable `PHYSICAL' as an example, are as follows:
\begin{enumerate}[noitemsep]
    \item Identify unique categories and count per variable, e.g. {`PHYSICAL'} has 3 categories: Yes, No, Unknown (Table \ref{Tab:variables})
    \item Create a new binary variable for each category. For each admission, only one of the new binary variables is `hot' (1), indicating its category. This is illustrated by example in Tables \ref{tab:first_table} and \ref{tab:second_table} for the variable `PHYSICAL' recorded over 3 admissions. 
    
    \item Concatenate these new binary columns to the dataset. Hence, from the example, instead of one 3-value variable, `PHYSICAL' is now encoded into three separate binaries that \ac{ml} models can easily use.
\end{enumerate}

\paragraph{Step 3: Data normalisation.} Z-score normalisation was applied to all numerical input variables in Table \ref{Tab:variables}. Through this, each numerical variable was centred to have zero mean and unit variance. The standardised data retains the skewness and kurtosis shape properties of the original dataset.

\subsection*{Stage III: Model training and evaluation methodology}
\label{TrainTest}
\paragraph{Step 1: Initial train and test split.} The preprocessed dataset comprising the model inputs was split into two separate sets: a training set and a test set, utilising a stratified 50-50 partition. Hence, the training and test set each had 16\,121 samples for the male dataset and 15\,094 samples for the female group. 

\paragraph{Step 2: Downsampling of the training set.} 
To ensure a balanced representation across the two classes of the target variable (long stay: {LOSClass=1} and short stay: {LOSClass=0}), the majority class (short stay) in the training set was downsampled. Specifically, the number of training samples belonging to the short-stay class was reduced to match the number of examples from the smaller, long-stay class. Applying this balanced downsampling serves to mitigate potential modelling inefficiencies caused by class imbalance, where models may ignore or not properly learn underrepresented classes. Also, by harmonising class distributions, downsampling can improve model evaluation metrics related to average performance across classes, such as balanced accuracy, as both classes are weighted and assessed equally. Tables \ref{tab: Traintest_male} and  \ref{tab: Traintest_female} show the demographic distribution of the training and test data applied in classifying the \ac{los} for males and females respectively.

\paragraph{Step 3: Model training and testing.}

\begin{enumerate}[noitemsep]
    \item  To identify an optimal ML model for predicting hospitalisation duration ({LOSClass}), eight classifiers were evaluated including \ac{lr}, \ac{svm}, \ac{rf}, \ac{gboost}, \ac{histgboost}, \ac{xgboost}, \ac{knn} and a sequential \ac{nn} model. Table \ref{tab:Class_param} indicates the parameter configurations set for each classifier. Tables \ref{tab: NN_male} and \ref{tab: NN_female} detail the parameter configurations for the \ac{nn}.
    
    \item  After the optimal model (\ac{rf}) was selected, it was further evaluated using repeated random train/test splits to scrutinise the model's generalizability. Specifically, the dataset was randomly split into dedicated 50\% sized training sets and 50\% sized testing sets a total of 10 times. This generated 10 distinct train/test set combinations, allowing for evaluation across different data partitions. For each of the 10 train/test splits, the corresponding training set was downsampled as described in step 2 of \hyperref[TrainTest]{Stage III} and used to train the model. The model performance was evaluated at each iteration using its corresponding test set. Finally, the evaluation performance across the 10 iterations with different dataset splits was averaged to assess generalisability.
    
\end {enumerate}

\paragraph{Step 4: Performance evaluation metrics.} Several evaluation metrics were employed to evaluate each classifier's performance on the test set. Let $|TP|$ denote the number of unique admissions for which the long stay (i.e. {LOSClass=1}) was correctly classified; $|TN|$ be the number of short stays ({LOSClass=0}) correctly classified; $|FP|$ be the number of short stays incorrectly classified as long stays; $|FN|$ be the number of long stays incorrectly classified as short stays; $|P|$ be the total number of long-stay admissions, where $|P|=|TP|+|FN|$; and $|N|$ represent the total number of short-stay admissions, where $|N|=|TN|+|FP|$. The following metrics were utilised to evaluate the performance of the \ac{ml} models. 

\begin{align}
&\text{True positive rate (TPR)} = \frac{\text{|TP|}}{\text{|TP| + |FN|}} \in (0,1) \\
&\text{True negative rate (TNR)} = \frac{\text{|TN|}}{\text{|TN| + |FP|}} \in (0,1) \\
&\text{Balanced accuracy} = \frac{\text{TPR + TNR}}{2} \in (0,1).
\end{align}

The closer the values of the abovementioned metrics are to 1 the better the performance of the model. The \ac{fnr} and \ac{fpr}  are given by the expression:

\begin{align}
    &\text{FNR}= 1-\text{TPR}=\frac{\text{|FN|}}{\text{|TP|+|FN|}}, \in(0,1),\\
    &\text{FPR}=1-\text{TNR}= \frac{\text{|FP|}}{\text{|TN|+|FP|}}, \in(0,1),
\end{align}
The closer the values of the \ac{fnr} and the \ac{fpr} are to 0 the better the performance of the model. \\

\noindent Another important evaluative measure is the \ac{roc} curve, which plots the \ac{tpr} against the \ac{fpr} at different threshold values. This creates a curve from (0,0) to (1,1). The \ac{auc} measures the two-dimensional area beneath this curve. Greater \ac{auc} shows better performance in predicting the long (\ac{los}$\geq$ 4 days) and short (\ac{los}$<$ 4 days) hospital stays.\\

\noindent The evaluations also refer to the performance range which is obtained by taking the difference between the maximum and minimum values for each metric across ethnic groups. For example, in Table \ref{tab:rf_comp_Female}, the false negative rate (FNR) with the exponentiated gradient (EG) algorithm has a maximum of 0.333 for the `Black' group and a minimum of 0.226 for the `Unknown' group. Hence, its range is calculated as:
\[\textrm{Performance range}_{\textrm{FNR}} = \max\{\textrm{FNR}\} - \min\{\textrm{FNR}\}\]
The closer the performance range is to 0 the better the performance of the model.

\subsection*{Stage IV: Bias analysis and mitigation} 
To check for fairness across ethnic groups, the optimal random forest model was analysed by ethnic groups. Maximising overall accuracy can disproportionately impact certain groups. Therefore, two bias mitigation techniques (reductions with \ac{eg} and threshold optimizer) were evaluated to balance performance metrics across ethnic groups:
 
\paragraph{Reductions with \acl{eg}}\cite{agarwal2018reductions}. The reductions algorithm was applied during training to limit performance ranges across ethnic groups. A fairness constraint was defined on \ac{fnr} parity, requiring the \ac{fnr} range between ethnic groups to be at most 0.2. The model was trained and evaluated to measure unfairness based on this constraint. The \ac{eg} algorithm assigned weights to training instances that reduce the overall violation of the fairness constraint, with higher weights to instances contributing more to unfairness. The weighted training data was used to retrain the model to focus more on instances contributing to unfairness. These steps were repeated 10 times, updating instance weights iteratively to reduce bias while maintaining accuracy across ethnic groups. At each iteration, the algorithm updates the model parameters by considering the gradient of an objective function incorporating both predictive balanced accuracy and fairness constraints. The model parameters are then updated to effectively adjust the model to reduce bias while maintaining predictive balanced accuracy.

\paragraph{Threshold optimizer.} To check for performance discrepancies across demographic groups, the post-processing threshold optimizer approach\cite{hardt2016equality} was utilised. Specifically, the threshold optimizer tuned the decision boundary of the random forest classifier to achieve parity in the balanced accuracy metric between ethnic groups subject to enforcing constraints on the \acp{fnr} per group. This optimized the fairness-balanced accuracy trade-off solution without needing to modify the underlying \ac{ml} model or training procedure. The threshold optimizer takes an already trained model and fits a transformation function to the model's outputs to satisfy certain fairness constraints. This approach allows for mitigating unfairness when developers have no control over the training process of the model, which may occur due to practical limitations or considerations around security or privacy.

\subsection*{Ethical approval and Data availability statement.} 
 The anonymised individual-level data sources used in this study are available in the \ac{sail} Databank at Swansea University, Swansea, UK, but as restrictions apply, they are not publicly available. All proposals to use SAIL data are subject to review by the independent \ac{igrp}. Before any data can be accessed, approval must be given by the \ac{igrp}. The \ac{igrp} gives careful consideration to each project to ensure proper and appropriate use of \ac{sail} data. When access has been granted, it is gained through a privacy-protecting safe haven and remote access system referred to as the \ac{sail} Gateway. \ac{sail} has established an application process to be followed by anyone who would like to access data via SAIL at: \url{https://www.saildatabank.com/application-process/} 

\noindent  This project was approved by the \ac{igrp} (\ac{igrp} Project: 1375). All research was conducted in compliance with the \ac{gdpr} guidelines, which safeguards all health-related data in the medical field. Individual written patient consent was not required for this study.

\section*{Acknowledgements}
Data-driven machinE-learning aided stratification and management of multiple long-term COnditions in adults with intellectual disabilitiEs (DECODE) project (NIHR203981) is funded by the NIHR AI for Multiple Long-term Conditions (AIM) Programme. The views expressed are those of the author(s) and not necessarily those of the NIHR or the Department of Health and Social Care.
\noindent This work uses data provided by patients and collected by the \ac{nhs} as part of their care and support. We would also like to acknowledge all data providers who make anonymised data available for research.

\section*{Author contributions statement}
Writing original draft: E.A.; Conceptualisation and design: E.A., G.C., S.G.; Data preparation: E.A., R.K., A.A., F.Z., N.K., D.F., G.C., S.G.; Data curation: E.A., R.K., A.A., D.F.; Funding acquisition: S.G., G.J., G.C., A.A., F.Z.; Analysis and modelling: E.A., G.C.; Interpretation of data: E.A., G.C., S.G.; Writing review and editing: E.A., G.C., R.K, R-z.K., F.Z., A.A., S.G., G.J.; Approving final version of manuscript: all authors.

\section*{Corresponding authors}
Correspondence to Georgina Cosma g.cosma@lboro.ac.uk

\section*{Ethics declaration}
\paragraph{Competing interests.} The authors declare no competing interests.

\section*{Additional information}
The list of Read and \ac{icd10} codes of conditions and medications can be found here: https://github.com/gcosma/DECODELengthofStayML}
The machine learning code can also be found in the above repository. This repository will be made public once the paper is published. 
\label{LastPageOfMain}

\clearpage

\pagenumbering{arabic}
\fancyfoot[r]{\small\sffamily\bfseries\thepage/\pageref{LastPageOfSupplementary}}

\newpage
\setcounter{page}{1}
\renewcommand{\thepage}{S\arabic{page}}
\renewcommand{\thesection}{S\arabic{section}}
\renewcommand{\thetable}{S\arabic{table}}
\renewcommand{\thefigure}{S\arabic{figure}}

\setcounter{figure}{0}

\newcommand{\sectionbreak}{\clearpage\phantomsection}
\section*{Supplementary Files}
\label{supplementary}
\par

{\raggedright\sffamily\bfseries\fontsize{20}{25}\selectfont Equitable Length of Stay Prediction for Patients with Learning Disabilities and Multiple Long-term Conditions Using Machine Learning \par}
\vskip10pt

{\raggedright\sffamily\bfseries\fontsize{12}{12}\usefont{OT1}{phv}{b}{n} Emeka Abakasanga\textsuperscript{1}, Rania Kousovista\textsuperscript{1}, Georgina Cosma*\textsuperscript{1}, Ashley Akbari\textsuperscript{2}, 
Francesco Zaccardi\textsuperscript{3},
Navjot Kaur\textsuperscript{3,4},
Danielle Fitt\textsuperscript{2},
Gyuchan Thomas Jun\textsuperscript{4},
Reza Kiani\textsuperscript{5},
and Satheesh Gangadharan\textsuperscript{5} \par}
\vskip18pt
{\raggedright\sffamily\fontsize{10}{12}\usefont{OT1}{phv}{m}{n}
    \textsuperscript{1}Loughborough University, Computer Science, School of Science, Loughborough, LE11 3TU, UK \\
    \textsuperscript{2}Population Data Science, Swansea University Medical School, Faculty of Medicine, Health \& Life Science, Swansea University, Swansea, Wales, UK \\
    \textsuperscript{3}University of Leicester, Diabetes Research Centre,  Leicester, LE1 7RH, UK \\
    \textsuperscript{4}Loughborough University, School of Design and Creative Arts, Loughborough, LE11 3TU, UK \\
    \textsuperscript{5}Leicestershire Partnership NHS Trust, Leicester, UK \\
    \textsuperscript{*}Correspondence: Georgina Cosma: g.cosma@lboro.ac.uk 
}
\vskip18pt%

\begin{abstract}
People with learning disabilities have a higher mortality rate and premature deaths compared to the general public,  as reported in published research in the UK and other countries. This study analyses hospitalisations of 9\,618 patients identified with learning disabilities and long-term conditions for the population of Wales using electronic health record (EHR) data sources from the SAIL Databank. We describe the demographic characteristics, prevalence of long-term conditions, medication history, hospital visits, and lifestyle history for our study cohort, and apply machine learning models to predict the length of hospital stays for this cohort. The random forest (RF) model achieved an Area Under the Curve (AUC) of 0.759 (males) and 0.756 (females), a false negative rate of 0.224 (males) and 0.229 (females), and a balanced accuracy of 0.690 (males) and 0.689 (females). After examining model performance across ethnic groups, two bias mitigation algorithms (threshold optimization and the reductions algorithm using an exponentiated gradient) were applied to minimise performance discrepancies. The threshold optimizer algorithm outperformed the reductions algorithm, achieving lower ranges in false positive rate and balanced accuracy for the male cohort across the ethnic groups. This study demonstrates the potential of applying machine learning models with effective bias mitigation approaches on EHR data sources to enable equitable prediction of hospital stays by addressing data imbalances across groups.

\end{abstract}

\begin{center}
{\footnotesize
\setlength{\tabcolsep}{2pt}
\begin{longtable}{@{}l@{}}
\caption{List of \acp{ltc} considered in this study.  }
\label{tab: ltc_list}\\
\toprule
CONDITIONS                  \\* \midrule
\endfirsthead
\multicolumn{1}{c}%
{{\bfseries Table \thetable\ continued from previous page}} \\
\toprule
CONDITIONS                  \\* \midrule
\endhead
\bottomrule
\endfoot
\endlastfoot
ANAEMIA                     \\
BARRETTS OESOPHAGUS          \\
BRONCHIECTASIS              \\
CANCER                      \\
CARDIAC ARRHYTHMIAS          \\
CEREBRAL PALSY               \\
CHRONIC CONSTIPATION           \\
CHRONIC DIARRHOEA              \\
CHRONIC AIRWAY DISEASES   \\
CHRONIC ARTHRITIS          \\
CHRONIC PAIN CONDITIONS   \\
CHRONIC PNEUMONIA            \\
CIRRHOSIS                   \\
CHRONIC KIDNEY DISEASE                         \\
CORONARY HEART DISEASE        \\
DEMENTIA                    \\
DIABETES                    \\
DYSPHAGIA                   \\
EPILEPSY                    \\
HEARING LOSS                 \\
HEART FAILURE                \\
HYPERTENSION                \\
\acf{ibd}                         \\
INSOMNIA                    \\
INTERSTITIAL LUNG DISEASE     \\
MENOPAUSAL AND PRE-MENOPAUSAL \\
MENTAL ILLNESS               \\
MS                          \\
NEUROPATHIC PAIN             \\
OSTEOPOROSIS                \\
PARKINSONS                  \\
POLYCYSTIC OVARY SYNDROME                        \\
PSORIASIS                   \\
\acf{pvd}                         \\
REFLUX DISORDERS           \\
STROKE                      \\
THYROID DISORDERS            \\
TOURETTE                    \\
VISUAL IMPAIRMENT            \\* \bottomrule
\end{longtable}
}
\end{center}

\begin{center}
    {\footnotesize
    \setlength{\tabcolsep}{2pt}
\begin{longtable}{@{}lllllll@{}}
\caption{Demographic description of the cohort.  }
\label{Tab:cohort_desc}\\
\toprule
\multirow{2}{*}{}  & \multicolumn{3}{l}{Male}                                   & \multicolumn{3}{l}{Female}                                 \\* \cmidrule(l){2-7} 
                   & Patients (\%) & Unique Admissions (\%) & Long-stay rate \% & Patients (\%) & Unique Admissions (\%) & Long-stay rate \% \\* \midrule
\endfirsthead
\multicolumn{7}{c}%
{{\bfseries Table \thetable\ continued from previous page}} \\
\toprule
\multirow{2}{*}{}  & \multicolumn{3}{l}{Male}                                   & \multicolumn{3}{l}{Female}                                 \\* \cmidrule(l){2-7} 
                   & Patients (\%) & Unique Admissions (\%) & Long-stay rate \% & Patients (\%) & Unique Admissions (\%) & Long-stay rate \% \\* \midrule
\endhead
\bottomrule
\endfoot
\endlastfoot
Total              & 4929          & 32275                  & 38.885            & 4689          & 29968                  & 38.798            \\* \midrule
\multicolumn{7}{l}{Age}                                                                                                                      \\* \midrule
\textless{}30      & 159           & 1565                   & 36.741            & 119           & 1525                   & 37.967            \\
30-39              & 594 (12.051)  & 4883 (15.129)          & 32.521            & 531 (11.324)  & 4171(13.918)           & 33.277            \\
40-49              & 1065 (21.607) & 6954 (21.546)          & 35.620            & 965 (20.58)   & 6103(20.365)           & 32.591            \\
50-59              & 1229 (24.934) & 7677 (23.786)          & 38.205            & 1083 (23.097) & 6840(22.824)           & 35.146            \\
60-69              & 1025 (20.795) & 6556 (20.313)          & 42.145            & 934 (19.919)  & 6361(21.226)           & 38.610            \\
70-79              & 605 (12.274)  & 3631 (11.25)           & 44.643            & 640 (13.649)  & 3404(11.359)           & 54.083            \\
80+                & 252 (5.113)   & 1009 (3.126)           & 58.771            & 417 (8.893)   & 1564(5.219)            & 62.084            \\* \midrule
\multicolumn{7}{l}{Ethnic   group}                                                                                                           \\* \midrule
Asian              & 79 (1.603)    & 576 (1.785)            & 46.528            & 64 (1.365)    & 564 (1.882)            & 35.461            \\
Black              & 19 (0.385)    & 75 (0.232)             & 52.000            & 14 (0.299)    & 44 (0.147)             & 65.909            \\
Other              & 19 (0.385)    & 76 (0.235)             & 34.211            & 18 (0.384)    & 56 (0.187)             & 30.357            \\
Unknown            & 1225 (24.853) & 6208 (19.235)          & 39.030            & 1124 (23.971) & 5253 (17.529)          & 40.034            \\
White              & 3587 (72.773) & 25340 (78.513)         & 38.650            & 3469 (73.982) & 24051 (80.256)         & 38.576            \\* \midrule
\multicolumn{7}{l}{WIMD}                                                                                                                     \\* \midrule
1 (Most Deprived)  & 1227 (24.893) & 9071 (28.105)          & 37.813            & 1207 (25.741) & 8177 (27.286)          & 38.963            \\
2                  & 990 (20.085)  & 6947 (21.524)          & 38.491            & 943 (20.111)  & 6855 (22.874)          & 35.799            \\
3                  & 791 (16.048)  & 4943 (15.315)          & 41.958            & 770 (16.421)  & 4660 (15.55)           & 42.253            \\
4                  & 721 (14.628)  & 4184 (12.964)          & 41.276            & 700 (14.929)  & 4433 (14.792)          & 39.274            \\
5 (Least Deprived) & 511 (10.367)  & 3068 (9.506)           & 39.309            & 494 (10.535)  & 2811 (9.38)            & 39.772            \\ 
Unknown            & 689 (13.978)  & 4062 (12.586)          & 35.426            & 575 (12.263)  & 3032 (10.117)          & 38.226            \\* \bottomrule
\end{longtable}
\par
\noindent Remark: Long-stay rate defined in formula \eqref{losrate}.\\ The `Patients' column across age groups takes the latest age group for each unique patient.
 }
\end{center}

\begin{center}
{\footnotesize
\setlength{\tabcolsep}{2pt}
\begin{longtable}{p{1.7cm}p{4.5cm}p{4.5cm}p{4cm}}

\caption{Variables used for the \acf{ml} training and analysis.  }
\label{Tab:variables} \\
\toprule
\multicolumn{1}{ l}{\textbf{}}	  &  \multicolumn{1}{ l}{\textbf{Variable name}}		   &  \multicolumn{1}{ l}{\textbf{Description}}		                            & \multicolumn{1}{ l}{\textbf{Variable type}}\\ \midrule 
\endfirsthead

\multicolumn{4}{c}%
{{\bfseries \tablename\ \thetable{} -- continued from previous page}} \\
\toprule \multicolumn{1}{ l}{\textbf{}} & \multicolumn{1}{ l}{\textbf{Variable name}} & \multicolumn{1}{ l}{\textbf{Description}} & \multicolumn{1}{ l}{\textbf{Variable type}} \\ \midrule 
\endhead


\multicolumn{4}{c}{Input variables to \ac{ml} models }     \\ \midrule
{Patient lifestyle and history}		            &  {\ac{bmi}}		&  {\ac{bmi} value documented closest to the admission date}		&  {Categorical Variable: (pre-obesity, obesity class I, obesity class III, normal weight, underweight,
and `Unknown' )}		\\  
{}		&  {SMOKING\_HISTORY}		            &  {History of smoking prior to admission}		&  {Categorical Variable ( Yes, No, `Unknown' )}		\\  
{}		&  {ALCOHOL\_HISTORY}		            &  {history of alcohol intake prior to admission}		&  {Categorical Variable (Yes, No, `Unknown')}		\\  
{}		&  {PHYSICAL}		                    &  {Patient does light/regular exercise}		&  {Categorical Variable (Yes, No, `Unknown')}		\\    
{}		&  {AUTISM}		&  {Indicates if patient is autistic}		&  {Binary Variable (Yes/No)}		\\     
{Prior \newline hospitalization \newline data}		&  {NUM\_PRVADMISSION\_1RY}		&  {Number of hospital admissions in the past year from admission date}		&  {Numeric variable}		\\  
{}		&  {NUM\_PRVEPISODES\_1RY}		        &  {Total hospital episodes from all 1 year prior admissions}		&  {Numeric variable}		\\  
{}		&  {NUM\_PRVCOMORBID\_1RY}		        &  {Frequency of \acp{ltc} during previous admissions over past 1 year}		&  {Numeric variable}		\\  
{}		&  {NUM\_PRVADMISSION\_3RY}		        &  {Number of hospital admissions in the past 3 years from admission date}		&  {Numeric variable}		\\  
{}		&  {NUM\_PRVEPISODES\_3RY}		        &  {Total hospital episodes from all 3 years prior admissions}		&  {Numeric variable}		\\  
 {}		&  {NUM\_PRVCOMORBID\_3RY}		        &  {Frequency of \acp{ltc} during previous admissions over past 3 years}		&  {Numeric variable}		\\  
{}		&  {NUM\_PRVHOSPITAL\_DAYS\_1YR}		&  {Cumulative hospital days 1 year prior to admissions}		&  {Numeric variable}		\\  
{}		&  {NUM\_PRVHOSPITAL\_DAYS\_3YR}		&  {Cumulative hospital days 3 years prior to admissions}		&  {Numeric variable}		\\   
{Prescriptions}		&  {MEDICATIONS}		& {History of antipsychotic, antidepressant or  anti-manic / anti-epileptic medications}		&  {Binary variable (Yes/No)}		\\   
{Admission data}		&  {TOTAL\_COMORBIDITY}		&  {Total number of multiple \ac{ltc} by patient as at $24$ hours after admission date}		&  {Numeric variable}		\\  
{}		&  {NUMEPISODES\_24HRS}		&  {Total number of clinical episodes within 24 hours of admission}		&  {Numeric variable}		\\  
{}		&  {NUMCOMORBIDITIES\_24HRS}		&  {Number of \acp{ltc} linked to hospital episodes in the first $24$ hours of admission}		&  {Numeric variable}		\\  
{}		&  {COND}		& {Indicates which of the $64$ \acp{ltc} were linked to the patients' episodes in the first 24 hours of admission}		&  {Binary Variable for each condition. 60 \acp{ltc} for males and 62 for females}		\\ \midrule
\multicolumn{4}{c}{Target variable}     \\ \midrule
{Target \newline variable}		&  {LOSClass}		&  {Indicator of \ac{los} }		&  {Binary Variable 0: \ac{los}<4 and 1: \ac{los} $\geq 4$}		\\ \midrule 
\multicolumn{4}{c}{Demographic variables (for analysis)}     \\ \midrule
{Demographics}		&  AGEGRP\_AT\_ADMIS\_DT	&  {Age group of patient as at admission date }		&  {Categorical Variable (see table \ref{Tab:cohort_desc})}		\\  
{}		            &  {Ethnic group}		        &  {Ethnic group of patient}		                    &  {Categorical Variable (see table \ref{Tab:cohort_desc})}	\\  
{}	                &  {\ac{wimd}}		        &  {\acl{wimd} version 2019}		                                &  {Numeric variable on a scale of $1$ (most deprived) to $5$ (least deprived), and `Unknown' group}		\\ \bottomrule
\end{longtable}} 
\par
\noindent Remark: {`AUTISM'} is a neurodevelopmental disorder stated to occur from birth\cite{bonnet2018autism} but may be undiagnosed until adulthood. Therefore, this variable takes the value 1 for all the admission records of an autistic patient, regardless of when they were first diagnosed with autism. 
\end{center}

\begin{center}
{\footnotesize
\setlength{\tabcolsep}{2pt}
\begin{longtable}{lllll}
\caption{Mortality statistics of the cohort of hospitalized patients for the male and female sexes.}
\label{tab: Mortality_cohort}\\
\toprule
\multirow{2}{*}{}  & Male      &                         & Female    &                         \\ \cmidrule{2-5} 
                   & Mortality & In-hospital   mortality & Mortality & In-hospital   mortality \\ \midrule
\endfirsthead
\multicolumn{5}{c}%
{{\bfseries Table \thetable\ continued from previous page}} \\
\toprule
\multirow{2}{*}{}  & Male      &                         & Female    &                         \\ \cmidrule{2-5} 
                   & Mortality & In-hospital   mortality & Mortality & In-hospital   mortality \\ \midrule
\endhead
Total              & 1904      & 979                     & 1777      & 850                     \\ \midrule
\multicolumn{5}{l}{Age}                                                                        \\ \midrule
\textless{}30      & 26        & 10                      & 12        & 5                       \\
30-39              & 82        & 31                      & 66        & 29                      \\
40-49              & 231       & 111                     & 196       & 76                      \\
50-59              & 443       & 205                     & 337       & 147                     \\
60-69              & 517       & 281                     & 453       & 237                     \\
70-79              & 395       & 227                     & 381       & 191                     \\
80+                & 210       & 114                     & 332       & 165                     \\ \midrule
\multicolumn{5}{l}{WIMD}                                                                       \\ \midrule
1 (Most Deprived)  & 452       & 225                     & 437       & 216                     \\
2                  & 371       & 209                     & 378       & 187                     \\
3                  & 353       & 181                     & 312       & 143                     \\
4                  & 288       & 153                     & 298       & 133                     \\
5 (Least Deprived) & 196       & 104                     & 180       & 90                      \\
Unknown            & 244       & 107                     & 172       & 81             \\ \bottomrule        
\end{longtable}
}
\end{center}

\begin{center}
    {
    \footnotesize
\setlength{\tabcolsep}{2pt}

\begin{longtable}{p{3cm}p{1.7cm}p{1.5cm}p{2cm}p{2cm}p{2cm}p{2cm}p{2cm}}
\caption{Ranking of top 10 primary conditions treated during hospitalization between January 2011 to December 2021, for males with \ac{ld}.  }
\label{tab: primary_cond_male}\\
\toprule
\textbf{CONDITION}      & \textbf{ADMISSION COUNT} & \textbf{PATIENT COUNT} & \textbf{MEAN \newline ADMISSIONS \newline PER \newline PATIENT} & \textbf{\ac{std} OF \newline ADMISSIONS \newline PER   \newline PATIENT} & \textbf{MEDIAN ADMISSIONS PER \newline PATIENT} & \textbf{MODE \newline ADMISSIONS \newline PER \newline PATIENT}\\* \midrule
\endfirsthead
\multicolumn{7}{c}%
{{\bfseries Table \thetable\ continued from previous page}} \\
\toprule
\textbf{CONDITION}      & \textbf{ADMISSION COUNT} & \textbf{PATIENT COUNT} & \textbf{MEAN ADMISSIONS PER PATIENT} & \textbf{\ac{std} OF ADMISSIONS PER   PATIENT} & \textbf{MEDIAN ADMISSIONS PER PATIENT} & \textbf{MODE ADMISSIONS PER PATIENT} \\* \midrule
\endhead
CANCER                  & 1703                     & 349                   & 4.880                                & 7.567                                              & 2                                      & 1                                    \\
EPILEPSY                & 764                      & 328                   & 2.329                                & 2.599                                              & 1                                      & 1                                    \\
CHRONIC PNEUMONIA       & 713                      & 247                   & 2.887                                & 1.775                                              & 3                                      & 2                                    \\
CHRONIC AIRWAY DISEASES & 519                      & 178                   & 2.916                                & 5.897                                              & 1                                      & 1                                    \\
MENTAL ILLNESS          & 510                      & 193                   & 2.642                                & 2.880                                              & 1                                      & 1                                    \\
DIABETES                & 334                      & 145                   & 2.303                                & 2.462                                              & 1                                      & 1                                    \\
CORONARY HEART DISEASE  & 312                      & 176                   & 1.773                                & 1.289                                              & 1                                      & 1                                    \\
CHRONIC KIDNEY DISEASE                     & 269                      & 173                   & 1.555                                & 1.313                                              & 1                                      & 1                                    \\
PSORIASIS               & 261                      & 11                    & 23.727                               & 25.939                                             & *                                      & 1                                    \\
REFLUX DISORDERS        & 226                      & 173                   & 1.306                                & 0.780                                              & 1                                      & 1            \\ \bottomrule                       
\end{longtable}
    }
    
\end{center}

\begin{center}
{
\footnotesize
\setlength{\tabcolsep}{2pt}
\begin{longtable}{p{3cm}p{1.7cm}p{1.5cm}p{2cm}p{2cm}p{2cm}p{2cm}p{2cm}}
\caption{Ranking of top 10 primary conditions treated during hospitalization between January 2011 to December 2021, for females with \ac{ld}.  }
\label{tab: primary_cond_female}\\
\toprule
\textbf{CONDITION}      & \textbf{ADMISSION COUNT} & \textbf{PATIENT COUNT} & \textbf{MEAN \newline ADMISSIONS \newline PER \newline PATIENT} & \textbf{\ac{std} OF \newline ADMISSIONS \newline PER   \newline PATIENT} & \textbf{MEDIAN ADMISSIONS PER \newline PATIENT} & \textbf{MODE \newline ADMISSIONS \newline PER \newline PATIENT}\\* \midrule
\endfirsthead
\multicolumn{7}{c}%
{{\bfseries Table \thetable\ continued from previous page}} \\
\toprule
\textbf{CONDITION}      & \textbf{ADMISSION COUNT} & \textbf{PATIENT COUNT} & \textbf{MEAN ADMISSIONS PER PATIENT} & \textbf{\ac{std} OF ADMISSIONS PER   PATIENT} & \textbf{MEDIAN ADMISSIONS PER PATIENT} & \textbf{MODE ADMISSIONS PER PATIENT} \\* \midrule
\endhead
CANCER                  & 2149            & 386           & 5.567                       & 7.836                                     & 2                             & 1                           \\
CHRONIC KIDNEY DISEASE                     & 742             & 117           & 6.342                       & 51.548                                    & 1                             & 1                           \\
EPILEPSY                & 631             & 272           & 2.320                       & 2.849                                     & 1                             & 1                           \\
CHRONIC PNEUMONIA       & 471             & 177           & 2.661                       & 1.799                                     & 2                             & 2                           \\
CHRONIC AIRWAY DISEASES & 446             & 188           & 2.372                       & 3.008                                     & 1                             & 1                           \\
CHRONIC ARTHRITIS       & 386             & 176           & 2.193                       & 8.375                                     & 1                             & 1                           \\
MENTAL ILLNESS          & 349             & 179           & 1.950                       & 1.825                                     & 1                             & 1                           \\
REFLUX DISORDERS        & 259             & 212           & 1.222                       & 0.626                                     & 1                             & 1                           \\
DIABETES                & 203             & 89            & 2.281                       & 5.829                                     & 1                             & 1                           \\
IBD                     & 189             & 93            & 2.032                       & 4.228                                     & 1                             & 1  \\ \bottomrule                        
\end{longtable}
    
    }
\end{center}

\begin{center}
{\footnotesize
\setlength{\tabcolsep}{2pt}

\begin{longtable}{@{}lllllll@{}}
\caption{Common conditions treated during admission of patients between 2011-2021. }
\label{tab:2011_2021_conditions}\\
\toprule
   & MALE                      &       &          & FEMALE                      &       &          \\* \midrule
      & CONDITION                 & Count & Count \% & CONDITION                   & Count & Count \% \\* \midrule
\endfirsthead
\multicolumn{7}{c}%
{{\bfseries Table \thetable\ continued from previous page}} \\
\toprule
   & MALE                      &       &          & FEMALE                      &       &          \\* \midrule
      & CONDITION                 & Count & Count \% & CONDITION                   & Count & Count \% \\* \midrule
\endhead

1  & EPILEPSY                  & 5453  & 29.410   & EPILEPSY                    & 4238  & 24.097   \\
2  & DIABETES                  & 4519  & 24.373   & CHRONIC AIRWAY DISEASES   & 4002  & 22.755   \\
3  & CHRONIC AIRWAY DISEASES & 3619  & 19.519   & DIABETES                    & 3814  & 21.686   \\
4  & MENTAL ILLNESS             & 3252  & 17.540   & THYROID DISORDERS            & 2995  & 17.030   \\
5  & CANCER                    & 2243  & 12.098   & MENTAL ILLNESS               & 2763  & 15.710   \\
6  & CHRONIC KIDNEY DISEASE                       & 1933  & 10.426   & CANCER                      & 2575  & 14.641   \\
7  & CORONARY HEART DISEASE      & 1783  & 9.617    & CHRONIC KIDNEY DISEASE                         & 2104  & 11.963   \\
8  & THYROID DISORDERS          & 1660  & 8.953    & CHRONIC ARTHRITIS          & 1506  & 8.563    \\
9  & CARDIAC ARRHYTHMIAS        & 1465  & 7.901    & CARDIAC ARRHYTHMIAS          & 1250  & 7.108    \\
10 & CEREBRAL PALSY             & 1464  & 7.896    & CEREBRAL PALSY               & 1118  & 6.357    \\
11 & DEMENTIA                  & 983   & 5.302    & CORONARY HEART DISEASE        & 1091  & 6.203    \\
12 & CHRONIC ARTHRITIS        & 970   & 5.232    & DEMENTIA                    & 1019  & 5.794    \\
13 & CHRONIC PNEUMONIA          & 944   & 5.091    & OSTEOPOROSIS                & 816   & 4.640    \\
14 & HEART FAILURE              & 857   & 4.622    & REFLUX DISORDERS           & 808   & 4.594    \\
15 & REFLUX DISORDERS         & 851   & 4.590    & HEART FAILURE                & 648   & 3.685    \\
16 & HEARING LOSS               & 585   & 3.155    & CHRONIC PNEUMONIA            & 607   & 3.451    \\
17 & OSTEOPOROSIS              & 570   & 3.074    & ANAEMIA                     & 571   & 3.247    \\
18 & ANAEMIA                   & 491   & 2.648    & STROKE                      & 394   & 2.240    \\
19 & DYSPHAGIA                 & 467   & 2.519    & HEARING LOSS                 & 379   & 2.155    \\
20 & INSOMNIA                  & 434   & 2.341    & CHRONIC CONSTIPATION           & 365   & 2.075    \\
21 & VISUAL IMPAIRMENT          & 428   & 2.308    & \acf{ibd}                         & 363   & 2.064    \\
22 & STROKE                    & 406   & 2.190    & DYSPHAGIA                   & 343   & 1.950    \\
23 & PSORIASIS                 & 392   & 2.114    & VISUAL IMPAIRMENT            & 299   & 1.700    \\
24 & CHRONIC CONSTIPATION         & 336   & 1.812    & CHRONIC PAIN CONDITIONS   & 273   & 1.552    \\
25 & \acs{ibd}                       & 318   & 1.715    & INSOMNIA                    & 231   & 1.313    \\
26 & PARKINSONS                & 278   & 1.499    & MENOPAUSAL AND PRE-MENOPAUSAL & 228   & 1.296    \\
27 & PVD                       & 263   & 1.418    & PSORIASIS                   & 227   & 1.291    \\
28 & BRONCHIECTASIS            & 218   & 1.176    & CHRONIC DIARRHOEA              & 168   & 0.955    \\
29 & CIRRHOSIS                 & 177   & 0.955    & \acf{pvd}                         & 148   & 0.842    \\
30 & NEUROPATHICPAIN           & 170   & 0.917    & NEUROPATHICPAIN             & 145   & 0.824    \\
31 & CHRONIC DIARRHOEA            & 162   & 0.874    & PARKINSONS                  & 106   & 0.603    \\
32 & BARRETTS OESOPHAGUS        & 141   & 0.760    & CIRRHOSIS                   & 80    & 0.455    \\
33 & INTERSTITIAL LUNG DISEASE   & 104   & 0.561    & BARRETTS OESOPHAGUS          & 77    & 0.438    \\
34 & CHRONIC PAIN CONDITIONS & 95    & 0.512    & INTERSTITIAL LUNG DISEASE     & 75    & 0.426    \\
35 & HYPERTENSION              & 43    & 0.232    & BRONCHIECTASIS              & 72    & 0.409    \\
36 & TOURETTE                  & 29    & 0.156    & HYPERTENSION                & 44    & 0.250    \\
37 & ADDISONS DISEASE           & 12    & 0.065    & POLYCYSTIC OVARY SYNDROME                         & 37    & 0.210    \\
38 &                           &       &          & ADDISONS DISEASE             & 35    & 0.199   \\ \bottomrule
\end{longtable}
}
\end{center}

\begin{center}
    \begin{figure}[!htb]
    \centering
\includegraphics[width=0.7\textwidth]{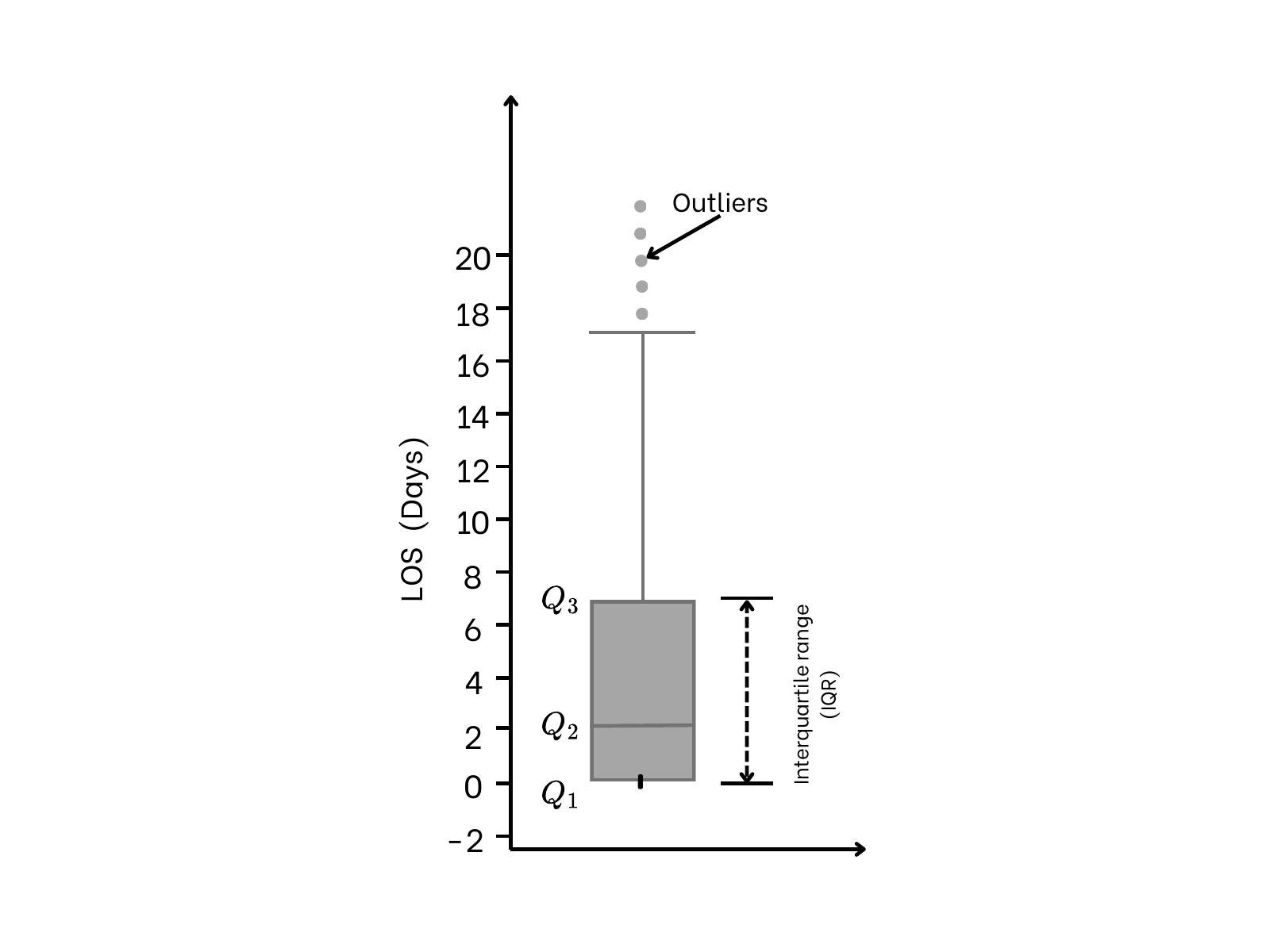}
    \caption{Box plot showing the distribution of \ac{los} across admission records for the combined male and female groups.
    Outliers in a box plot of all the unique \ac{los} days refer to days that fall significantly outside the range of the majority of the data. Specifically, outliers are defined based on the interquartile range and they help reveal patterns and anomalies that the bulk of the data could obscure.\\
     Remark: The dots indicating `Outliers' in this figure are illustrations (not real admission points), put in to indicate the presence of admissions above the upper whisker.}
    \label{fig: box_plot}
\end{figure}
\end{center}

\begin{center}
{\footnotesize
\setlength{\tabcolsep}{2pt}
\begin{longtable}{@{}lll@{}}
\caption{Top conditions for admissions with \ac{los} $\geq 129$ days. }
\label{tab:Admission129days}\\
\toprule
CONDITION                 & \% OCCURRENCE & ADMISSION COUNTS \\* \midrule
\endfirsthead
\multicolumn{3}{c}%
{{\bfseries Table \thetable\ continued from previous page}} \\
\toprule
CONDITION                 & \% OCCURRENCE & ADMISSION COUNTS \\* \midrule
\endhead
\bottomrule
\endfoot
\endlastfoot
MENTAL ILLNESS             & 61           & 570              \\
EPILEPSY                  & 21.6         & 202              \\
DIABETES                  & 9.4          & 88               \\
DEMENTIA                  & 7.3          & 68               \\
CEREBRAL PALSY            & 5.1          & 48               \\
CHRONIC KIDNEY DISEASE                       & 4.4          & 41               \\
THYROID DISORDERS         & 4.1          & 38               \\
CHRONIC AIRWAY DISEASES & 4            & 37               \\
CHRONIC ARTHRITIS        & 3.2          & 30               \\
DYSPHAGIA                 & 3.1          & 29               \\
CARDIAC ARRHYTHMIAS        & 2.5          & 23               \\
STROKE                    & 2.4          & 22               \\
CHRONIC PNEUMONIA          & 2.2          & 21               \\
CORONARY HEART DISEASE      & 2.2          & 21               \\
ANAEMIA                   & 2            & 19               \\
HEART FAILURE              & 1.9          & 18               \\
REFLUX DISORDERS         & 1.9          & 18               \\
HEARING LOSS               & 1.6          & 15               \\
OSTEOPOROSIS              & 1.5          & 14               \\
VISUAL IMPAIRMENT          & 1.3          & 12               \\
\acf{ibd}                       & 1.2          & 11               \\
INSOMNIA                  & 1.1          & 10               \\
CHRONIC CONSTIPATION         & 0.9          & 8                \\
CANCER                    & 0.9          & 8                \\* \bottomrule
\end{longtable}
}
\end{center}

\begin{center}
{\footnotesize
\setlength{\tabcolsep}{2pt}
\begin{longtable}{p{1cm}p{3.5cm}p{1.5cm}p{1cm}p{1.5cm}p{2cm}p{1cm}p{1.5cm}p{2cm}}
\caption{Chi-square and Binomial tests applied to determine if the categories of each multivariate variable occur with equal probabilities. All variables are described in Table \ref{Tab:variables}.}
\label{Tab: chisquareBinomial}\\
\toprule
\multicolumn{9}{c}{Hypothesis Test   Summary}    \\* \midrule                                                                                              \multirow{2}{*}{} & \multirow{2}{*}{Variable}                                                            & \multirow{2}{*}{Test}        & \multicolumn{3}{l}{Male}                           & \multicolumn{3}{l}{Female}                        \\* \cmidrule(l){4-9} 
                  &                                                                                             &                              & P-value$^a$  & Test stat  & Decision                      & P-value$^a$  & Test stat & Decision                      \\* \midrule                                                                                                    
\endfirsthead
\multicolumn{9}{c}
{{\bfseries Table \thetable\ continued from previous page}} \\
\toprule
\multicolumn{9}{c}{Hypothesis Test   Summary} \\* \midrule      \multirow{2}{*}{} & \multirow{2}{*}{Variable}                                                            & \multirow{2}{*}{Test}        & \multicolumn{3}{l}{Male}                           & \multicolumn{3}{l}{Female}                        \\* \cmidrule(l){4-9} 
                  &                                                                                             &                              & P-value$^a$  & Test stat  & Decision                      & P-value$^a$  & Test stat & Decision                      \\* \midrule                                                                                                                                                                  
\endhead
1                 & \ac{wimd}     & One-Sample   Chi-Square Test & 0.000 & 8440.207   & Reject hypothesis. & 0.000 & 8175.366  & Reject hypothesis. \\ 
2                 & Ethnic group                              & One-Sample   Chi-Square Test & 0.000 & 87412.885  & Reject hypothesis. & 0.000 & 84630.100 & Reject hypothesis. \\ 
3                 & AUTISM & One-Sample   Binomial Test   & 0.000 & 2833       & Reject hypothesis. & 0.000 & 27962.000 & Reject hypothesis. \\ 
4                 & ALCOHOL\_HISTORY                       & One-Sample   Chi-Square Test & 0.000 & 21268.570  & Reject hypothesis. & 0.000 & 30480.204 & Reject hypothesis. \\ 
5                 & SMOKING\_HISTORY                       & One-Sample   Chi-Square Test & 0.000 & 19438.225  & Reject hypothesis. & 0.000 & 23301.848 & Reject hypothesis. \\ 
6                 & MEDICATIONS & One-Sample   Binomial Test   & 0.000 & 14492      & Reject hypothesis. & 0.000 & 14207.000 & Reject hypothesis. \\ 
7                 & PHYSICAL                              & One-Sample   Chi-Square Test & 0.000 & 20305.609  & Reject hypothesis. & 0.000 & 20076.616 & Reject hypothesis. \\ 
8                 & BMI                                     & One-Sample   Chi-Square Test & 0.000 & 115339.196 & Reject hypothesis. & 0.000 & 98928.679 & Reject hypothesis. \\ 
9                 & AGEGRP\_AT\_ADMIS\_DT                   & One-Sample   Chi-Square Test & 0.000 & 18432.285  & Reject hypothesis. & 0.000 & 15325.888 & Reject hypothesis. \\ 
10                & LOSClass               & One-Sample   Binomial Test   & 0.000 & 12063      & Reject hypothesis. & 0.000 & 11138.000 & Reject hypothesis. \\* \midrule
\multicolumn{9}{l}{Hypothesis: The categories of each variable occur with equal probabilities.}\\* 
\multicolumn{9}{l}{a.   The significance level is .050.}                                                                       \\* \bottomrule
\end{longtable}
}  
\end{center}

\begin{center}
    {\footnotesize
\setlength{\tabcolsep}{2pt}

\begin{longtable}{@{}lllll@{}}
\caption{Distribution of LOSClass across variable categories.}
\label{tab:DistLOSClass}\\
\toprule
\multirow{3}{*}{} & \multicolumn{2}{l}{MALE}                                                           & \multicolumn{2}{l}{FEMALE}                                                         \\* \cmidrule(l){2-5} 
                  & \multirow{2}{*}{Admission Count} & \multirow{2}{*}{LOS\textgreater{}=4 \% (Count)} & \multirow{2}{*}{Admission count} & \multirow{2}{*}{LOS\textgreater{}=4 \% (Count)} \\
                  &                                  &                                                 &                                  &                                                 \\* \midrule
\endfirsthead
\multicolumn{5}{c}%
{{\bfseries Table \thetable\ continued from previous page}} \\
\toprule
\multirow{3}{*}{} & \multicolumn{2}{l}{MALE}                                                           & \multicolumn{2}{l}{FEMALE}                                                         \\* \cmidrule(l){2-5} 
                  & \multirow{2}{*}{Admission Count} & \multirow{2}{*}{LOS\textgreater{}=4 \% (Count)} & \multirow{2}{*}{Admission count} & \multirow{2}{*}{LOS\textgreater{}=4 \% (Count)} \\
                  &                                  &                                                 &                                  &                                                 \\* \midrule
\endhead
\multicolumn{5}{l}{Ethnic group}                                                                                                                                                               \\* \midrule
Asian             & 576                              & 0.465 (268)                                      & 564                              & 0.355 (200)                                      \\
Black             & 75                               & 0.52 (39)                                        & 44                               & 0.659 (29)                                       \\
Other             & 76                               & 0.342 (26)                                       & 56                               & 0.304 (17)                                       \\
Unknown           & 6208                             & 0.39 (2421)                                      & 5253                             & 0.4 (2101)                                       \\
White             & 25340                            & 0.387 (9807)                                     & 24051                            & 0.386 (9284)                                     \\* \midrule
\multicolumn{5}{l}{WIMD}                                                                                                                                     \\* \midrule
1 (Most Deprived)              & 9071                             & 0.378 (3429)                                     & 8177                             & 0.39 (3189)                                      \\
2                 & 6947                             & 0.385 (2675)                                     & 6855                             & 0.358 (2454)                                     \\
3                 & 4943                             & 0.42 (2076)                                      & 4660                             & 0.423 (1971)                                     \\
4                 & 4184                             & 0.413 (1728)                                     & 4433                             & 0.393 (1742)                                     \\
5 (Least Deprived)              & 3068                             & 0.393 (1206)                                     & 2811                             & 0.398 (1119)                                     \\
Unknown           & 4062                             & 0.354 (1438)                                     & 3032                             & 0.382 (1158)                                     \\* \midrule
\multicolumn{5}{l}{AUTISM}                                                                                                                                                                  \\* \midrule
0                 & 29399                            & 0.392 (11524)                                    & 28773                            & 0.391 (11250)                                    \\
1                 & 2876                             & 0.358 (1030)                                     & 1195                             & 0.316 (378)                                      \\* \midrule
\multicolumn{5}{l}{ALCOHOL\_HISTORY}                                                                                                                                                        \\* \midrule
0                 & 18380                            & 0.406 (7462)                                     & 20325                            & 0.377 (7663)                                     \\
1                 & 3734                             & 0.362 (1352)                                     & 1709                             & 0.377 (644)                                      \\
unknown           & 10161                            & 0.367 (3729)                                     & 7934                             & 0.419 (3324)                                     \\* \midrule
\multicolumn{5}{l}{SMOKING\_HISTORY}                                                                                                                                                        \\* \midrule
0                 & 16308                            & 0.394 (6425)                                     & 17563                            & 0.397 (6973)                                     \\
1                 & 12809                            & 0.377 (4829)                                     & 10325                            & 0.358 (3696)                                     \\
unknown           & 3158                             & 0.407 (1285)                                     & 2080                             & 0.462 (961)                                      \\* \midrule
\multicolumn{5}{l}{MEDICATIONS}                                                                                                                                                             \\* \midrule
0                 & 17298                            & 0.397 (6867)                                     & 14482                            & 0.392 (5677)                                     \\
1                 & 14977                            & 0.38 (5691)                                      & 15486                            & 0.384 (5947)                                     \\* \midrule
\multicolumn{5}{l}{PHYSICAL}                                                                                                                                                                \\* \midrule
0                 & 14765                            & 0.39 (5758)                                      & 13810                            & 0.399 (5510)                                     \\
1                 & 2541                             & 0.352 (894)                                      & 1988                             & 0.293 (582)                                      \\
unknown           & 14969                            & 0.394 (5898)                                     & 14170                            & 0.391 (5540)                                     \\* \midrule
\multicolumn{5}{l}{BMI}                                                                                                                                                                     \\* \midrule
normal weight     & 1084                             & 0.328 (356)                                      & 550                              & 0.396 (218)                                      \\
obesity class I   & 2869                             & 0.405 (1162)                                     & 3901                             & 0.366 (1428)                                     \\
obesity class III & 516                              & 0.357 (184)                                      & 961                              & 0.344 (331)                                      \\
pre-obesity       & 742                              & 0.41 (304)                                       & 687                              & 0.378 (260)                                      \\
underweight       & 483                              & 0.408 (197)                                      & 136                              & 0.449 (61)                                       \\
unknown           & 26581                            & 0.389 (10340)                                    & 23733                            & 0.393 (9327)                                     \\* \midrule
\multicolumn{5}{l}{AGEGRP\_AT\_ADMIS\_DT}                                                                                                                                                   \\* \midrule
0-20              & 86                               & 0.291 (25)                                       & 71                               & 0.366 (26)                                       \\
20-29             & 1479                             & 0.372 (550)                                      & 1454                             & 0.38 (553)                                       \\
30-39             & 4883                             & 0.325 (1587)                                     & 4171                             & 0.333 (1389)                                     \\
40-49             & 6954                             & 0.356 (2476)                                     & 6103                             & 0.326 (1990)                                     \\
50-59             & 7677                             & 0.382 (2933)                                     & 6840                             & 0.351 (2401)                                     \\
60-69             & 6556                             & 0.421 (2760)                                     & 6361                             & 0.386 (2455)                                     \\
70-79             & 3631                             & 0.446 (1619)                                     & 3404                             & 0.541 (1842)                                     \\
80+               & 1009                             & 0.588 (593)                                      & 1564                             & 0.621 (971)    \\ \bottomrule                                 
\end{longtable} 
    }
\end{center}

\begin{figure}[htp]
\centering
\includegraphics[width=.4\textwidth]{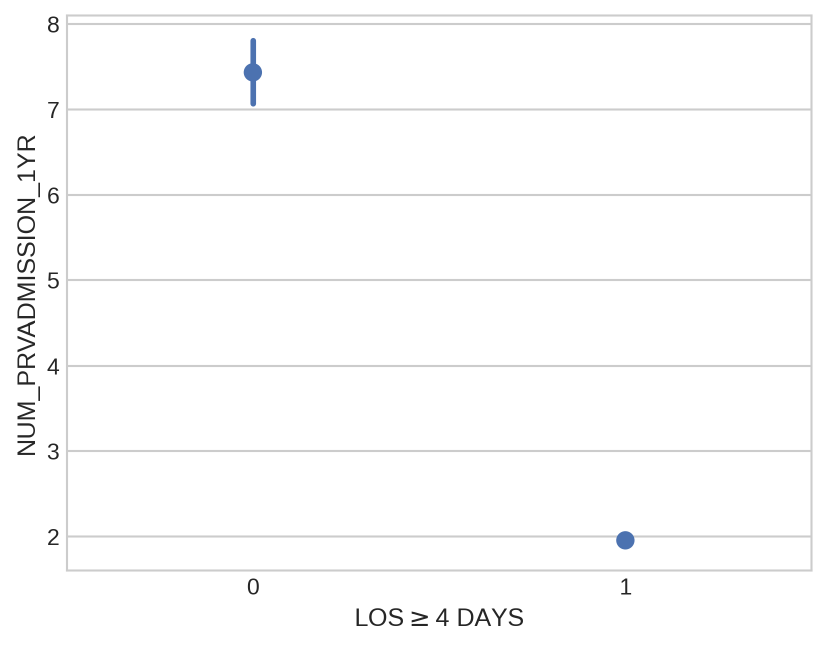}\quad
\includegraphics[width=.4\textwidth]{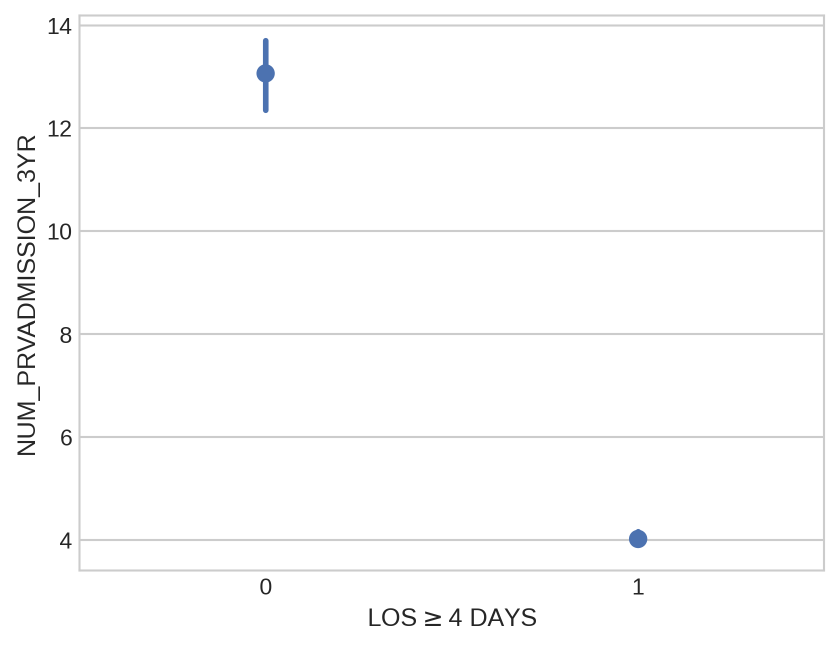}\quad

\medskip

\includegraphics[width=.4\textwidth]{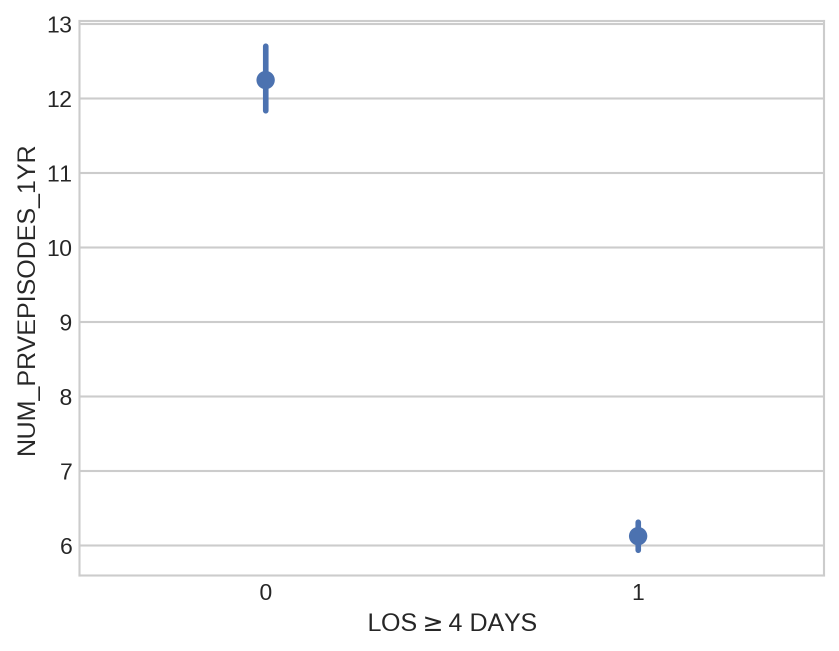}\quad
\includegraphics[width=.4\textwidth]{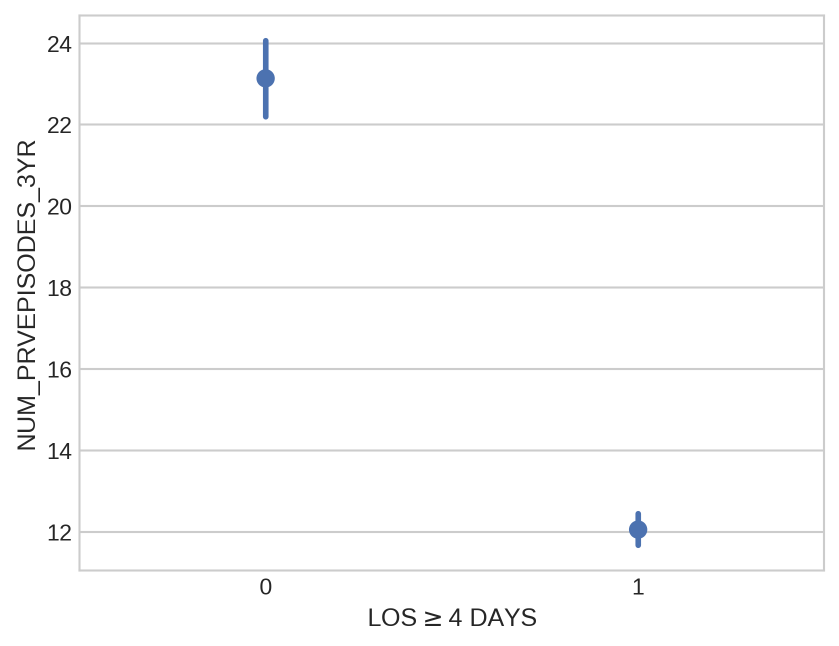}

\medskip

\includegraphics[width=.4\textwidth]{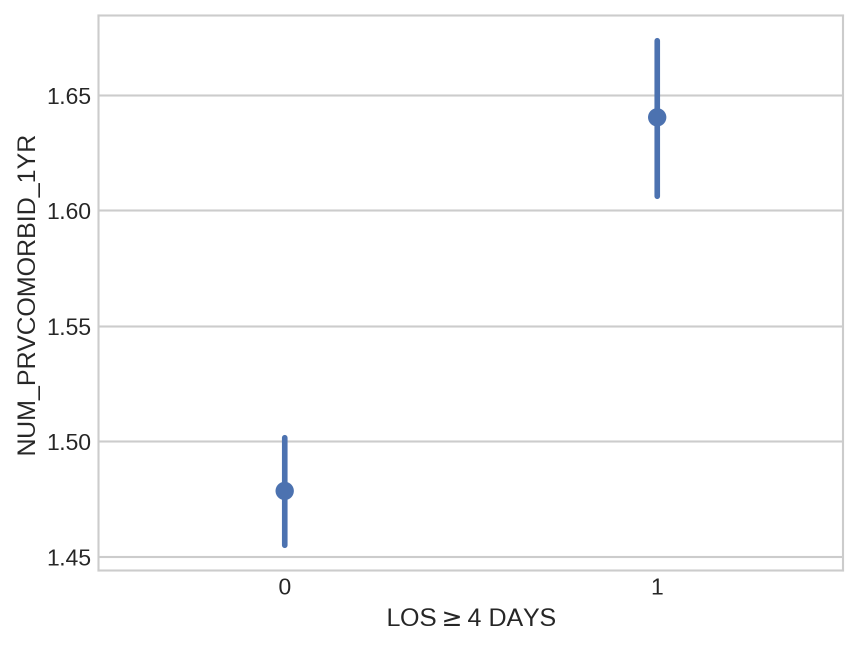}\quad
\includegraphics[width=.4\textwidth]{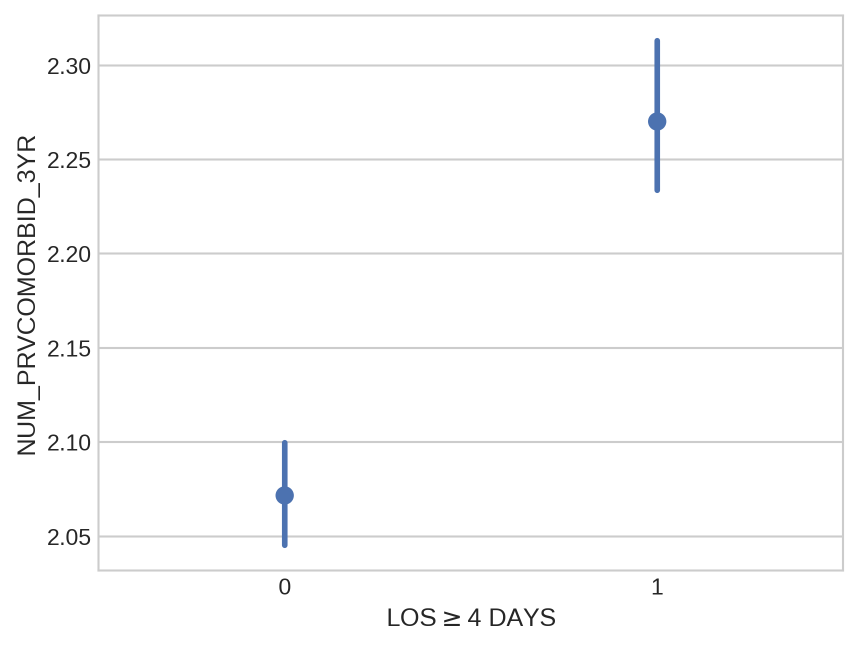}

\medskip

\includegraphics[width=.4\textwidth]{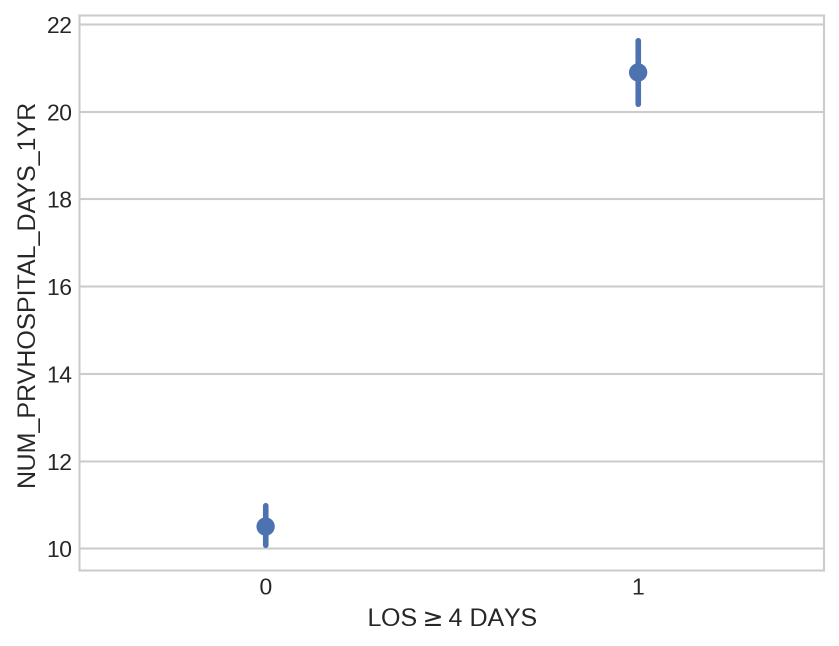}\quad
\includegraphics[width=.4\textwidth]{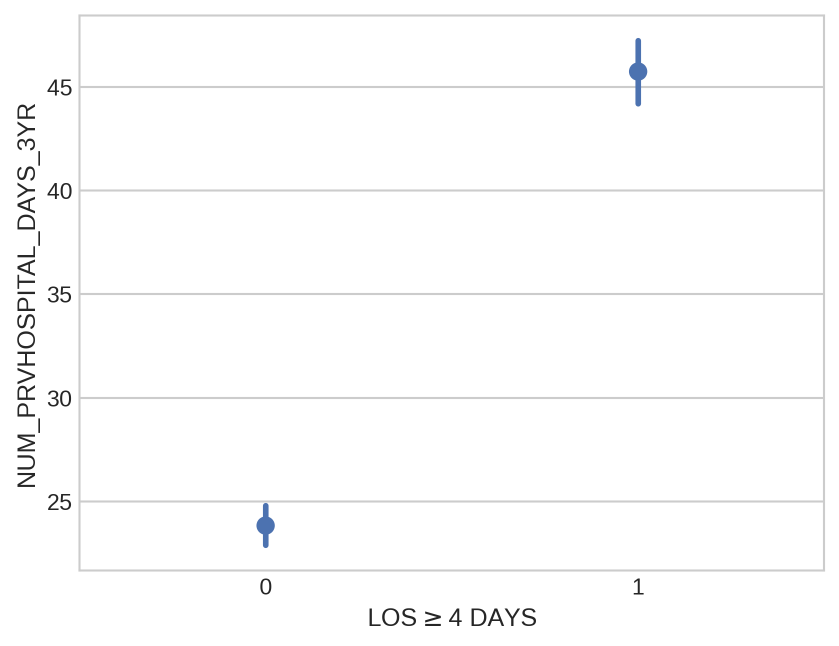}

\caption{Point plots showing the relationship between the numerical variables and \ac{los} $\geq 4 $ days for males. See Table \ref{Tab:variables} for description of variables.}
\label{fig: pointMale}
\end{figure}

\begin{figure}[t]
\centering
\includegraphics[width=.4\textwidth]{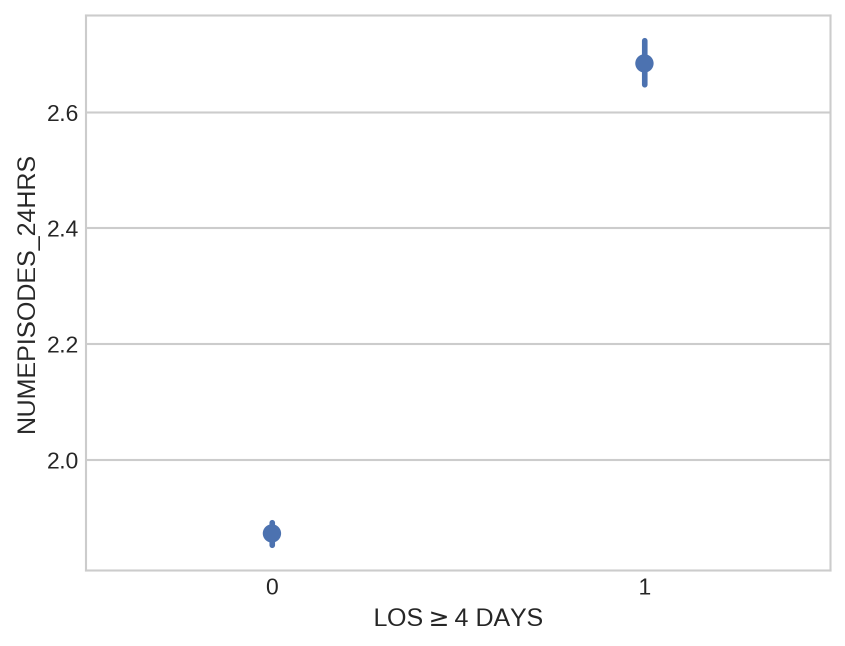}\quad
\includegraphics[width=.4\textwidth]{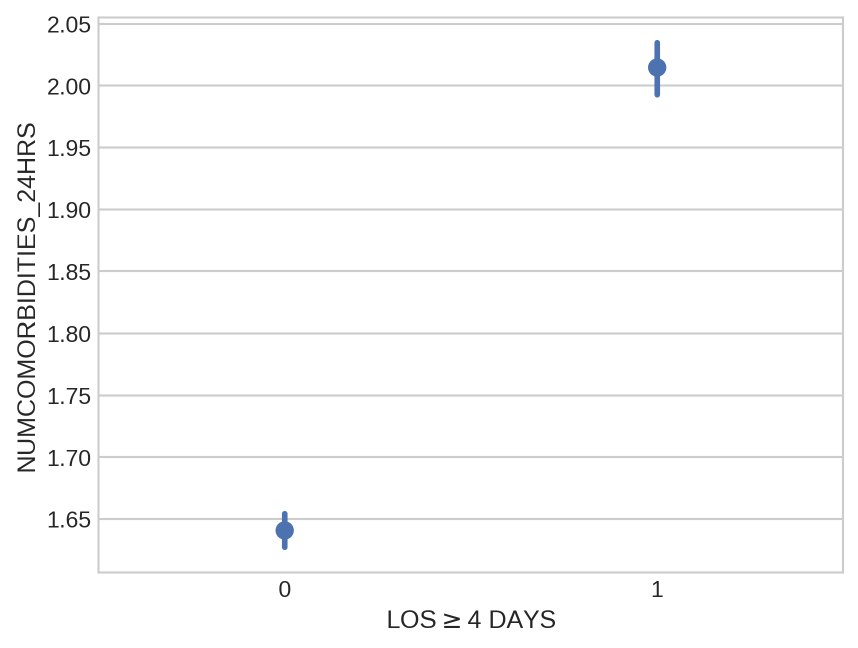}

\medskip

\includegraphics[width=.4\textwidth]{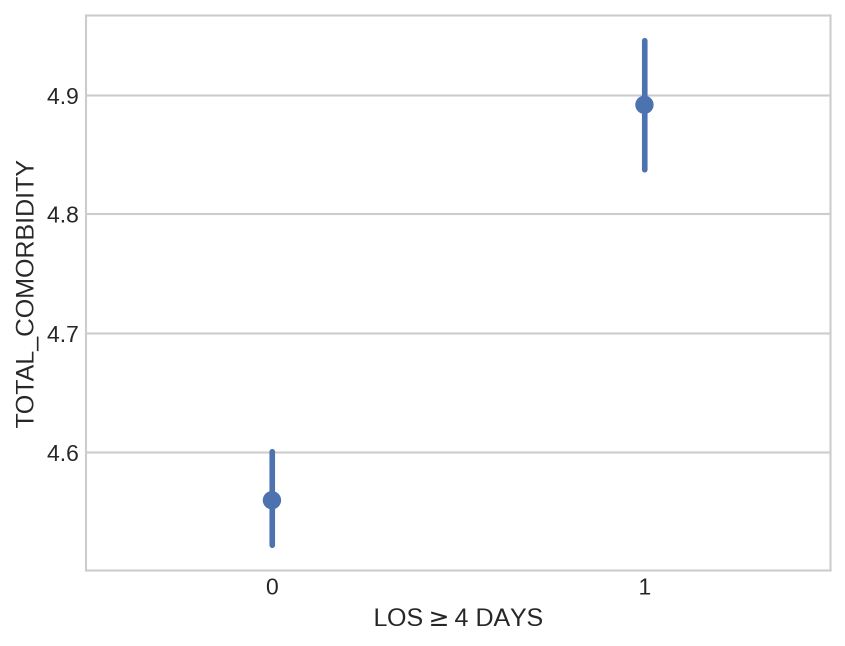}\quad

\caption{Point plots showing the relationship between the numerical variables and \ac{los} $\geq 4$ days for males. See Table \ref{Tab:variables} for description of variables.}
\label{fig: pointMale2}
\end{figure}

\begin{figure}[htp]
\centering
\includegraphics[width=.4\textwidth]{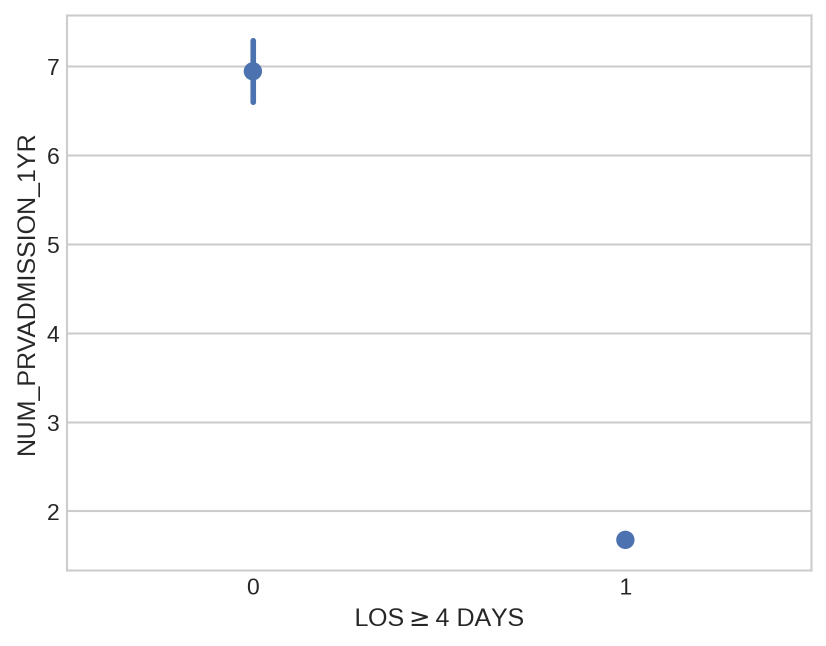}\quad
\includegraphics[width=.4\textwidth]{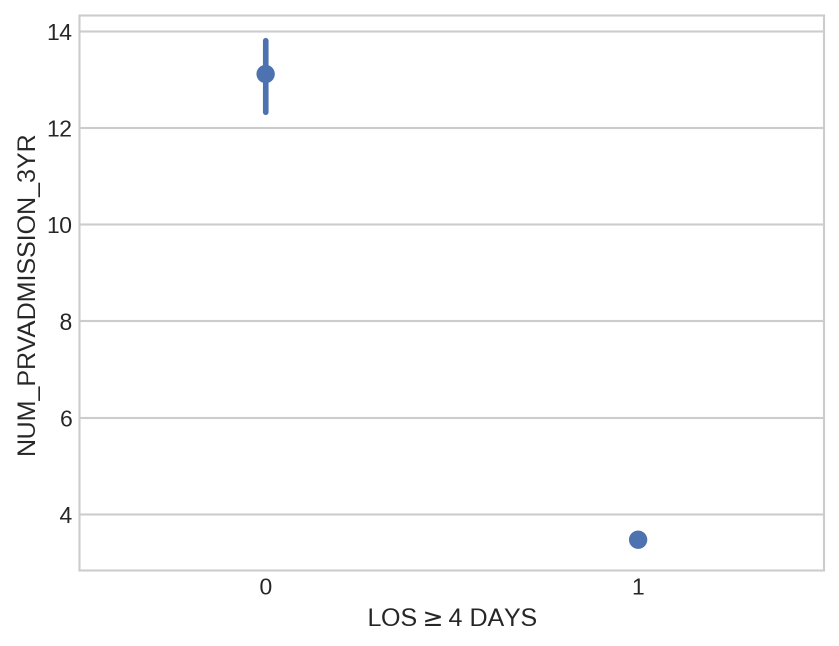}\quad

\medskip

\includegraphics[width=.4\textwidth]{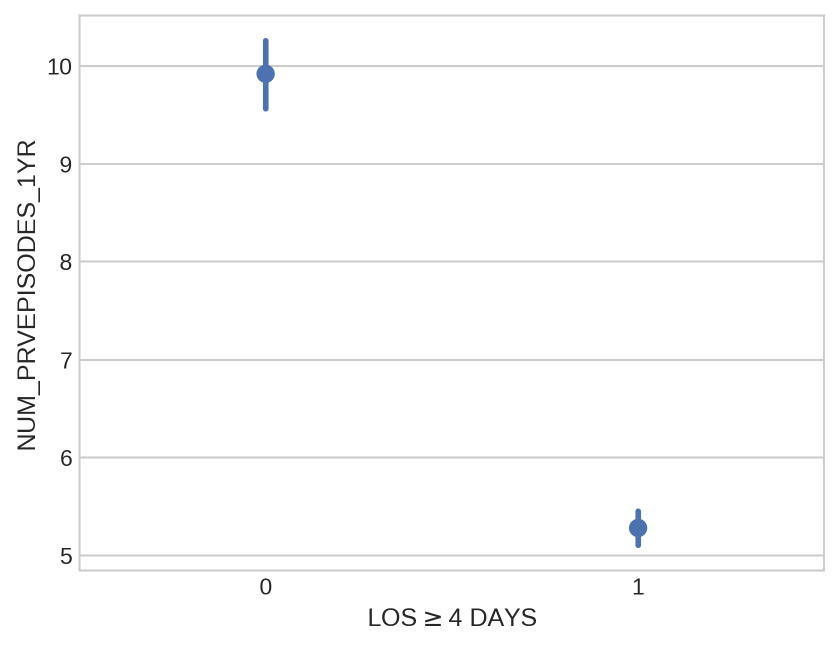}\quad
\includegraphics[width=.4\textwidth]{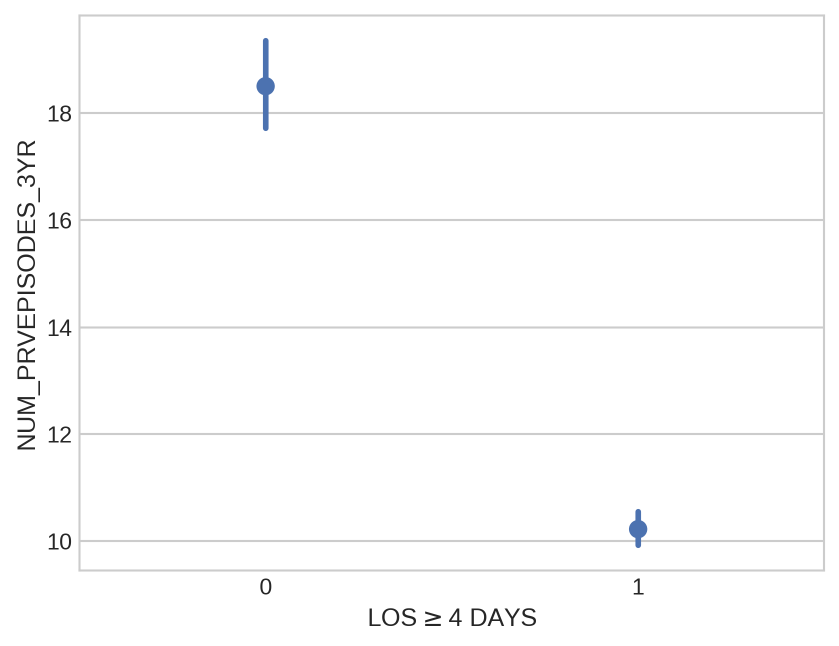}

\medskip

\includegraphics[width=.4\textwidth]{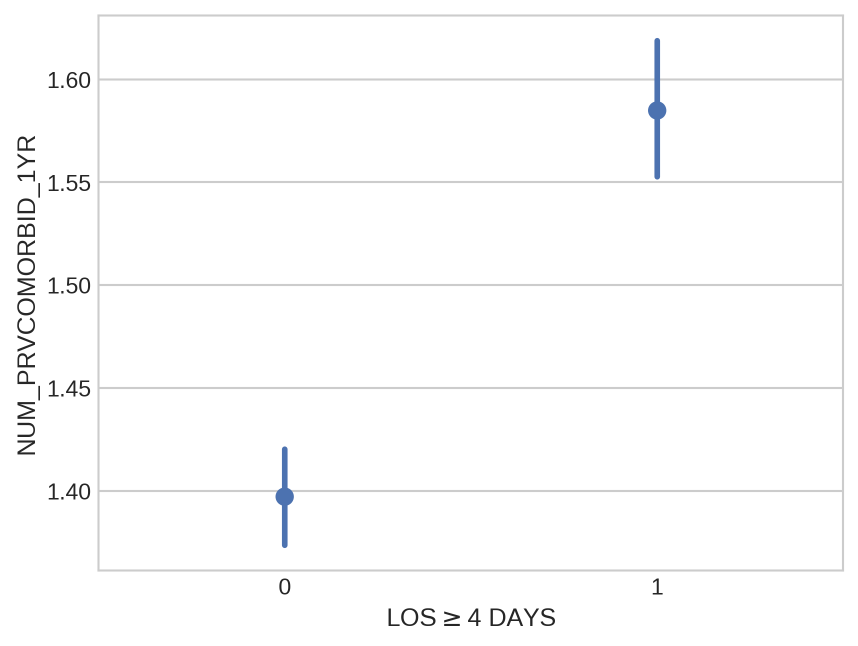}\quad
\includegraphics[width=.4\textwidth]{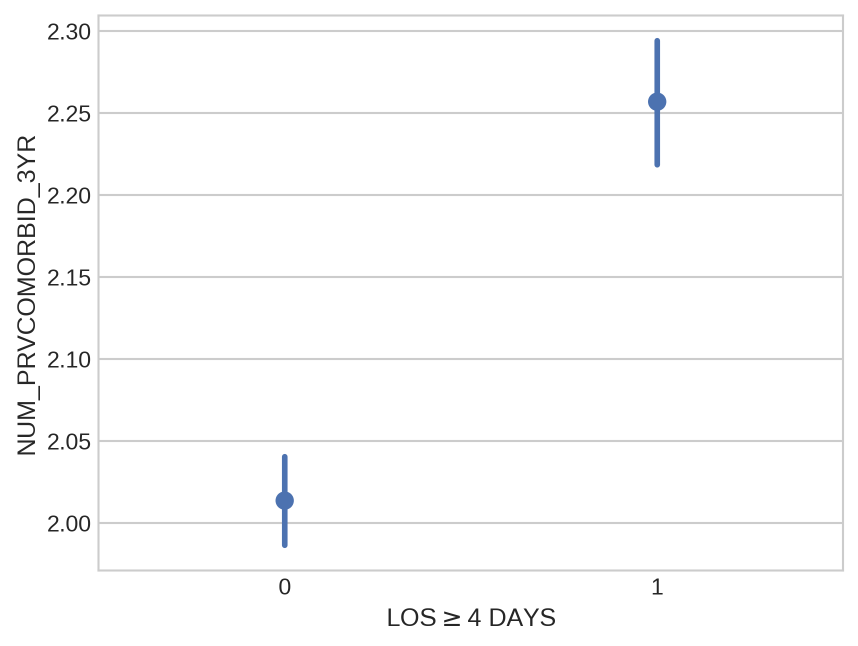}

\medskip

\includegraphics[width=.4\textwidth]{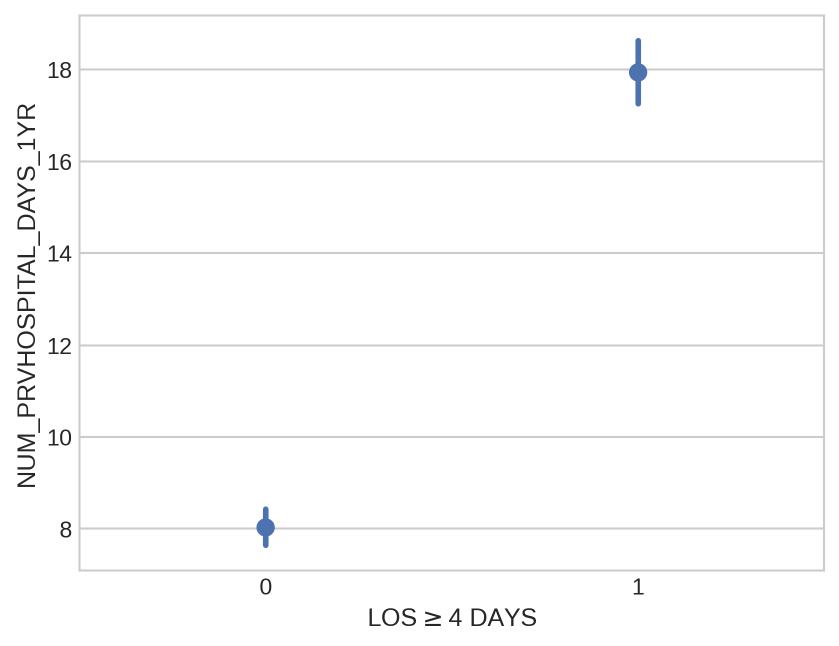}\quad
\includegraphics[width=.4\textwidth]{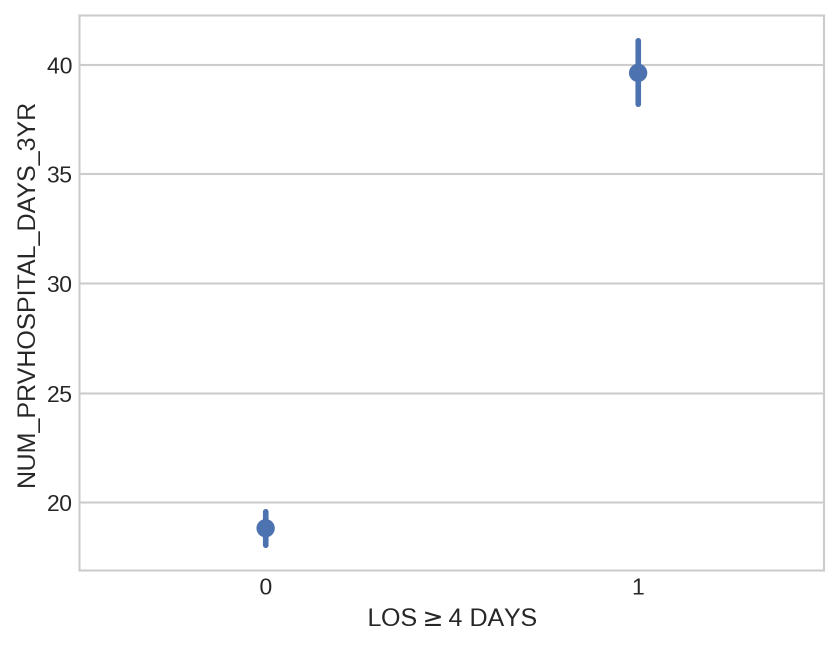}

\caption{Point plots showing the relationship between the numerical variables and \ac{los} $\geq 4 $ days for females. See Table \ref{Tab:variables} for description of variables.}
\label{fig: pointFemale}
\end{figure}

\begin{figure}[t]
\centering
\includegraphics[width=.4\textwidth]{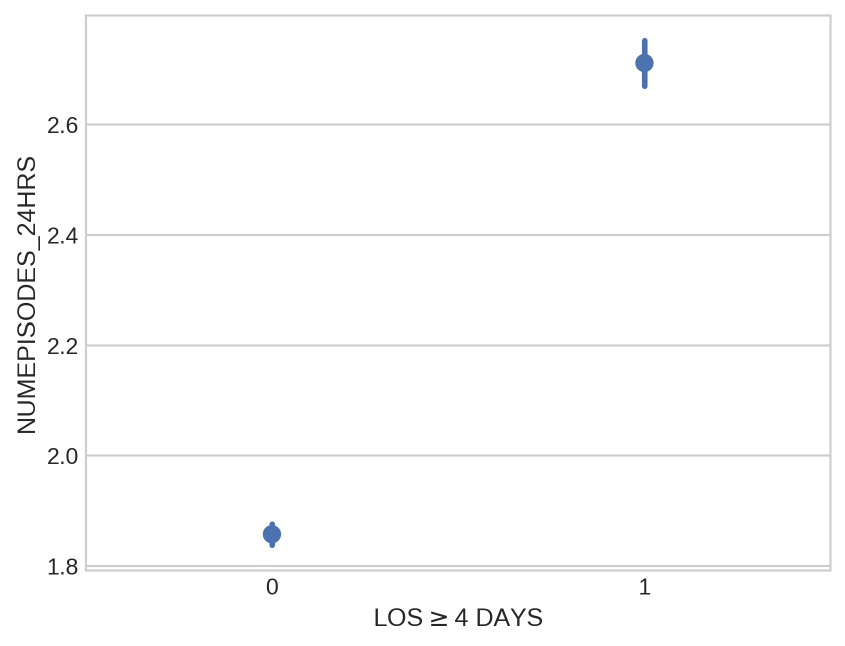}\quad
\includegraphics[width=.4\textwidth]{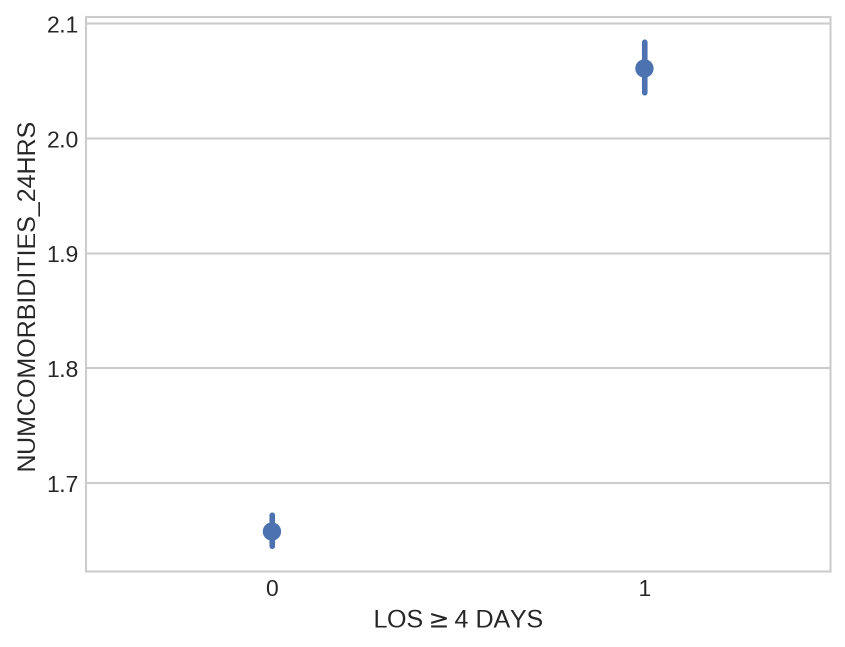}

\medskip

\includegraphics[width=.4\textwidth]{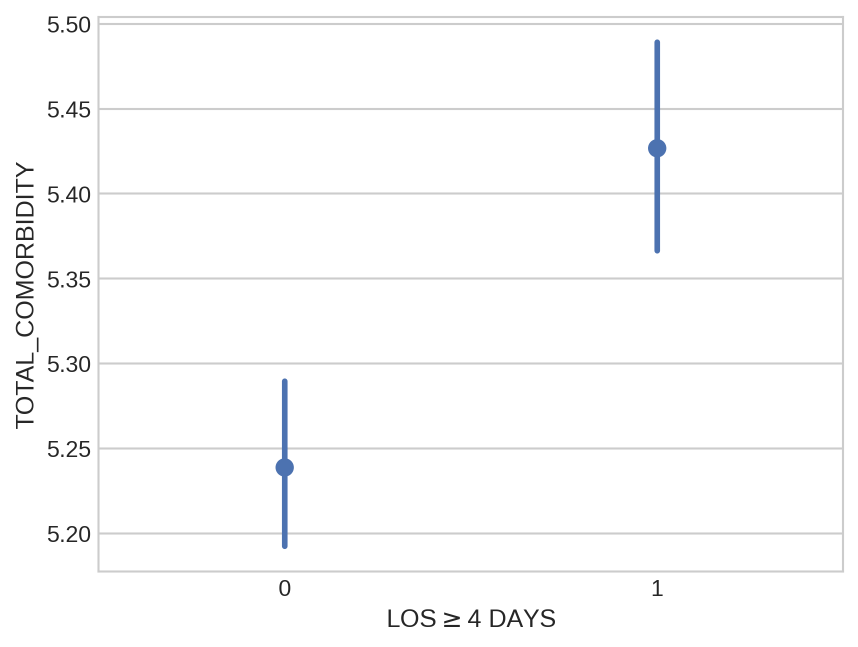}\quad

\caption{Point plots showing the relationship between the numerical variables and \ac{los} $\geq 4 $ days for females. See Table \ref{Tab:variables} for description of variables.}
\label{fig: pointFemale2}
\end{figure}

\begin{figure}[!htb]
    \centering
    \includegraphics[width=\textwidth]{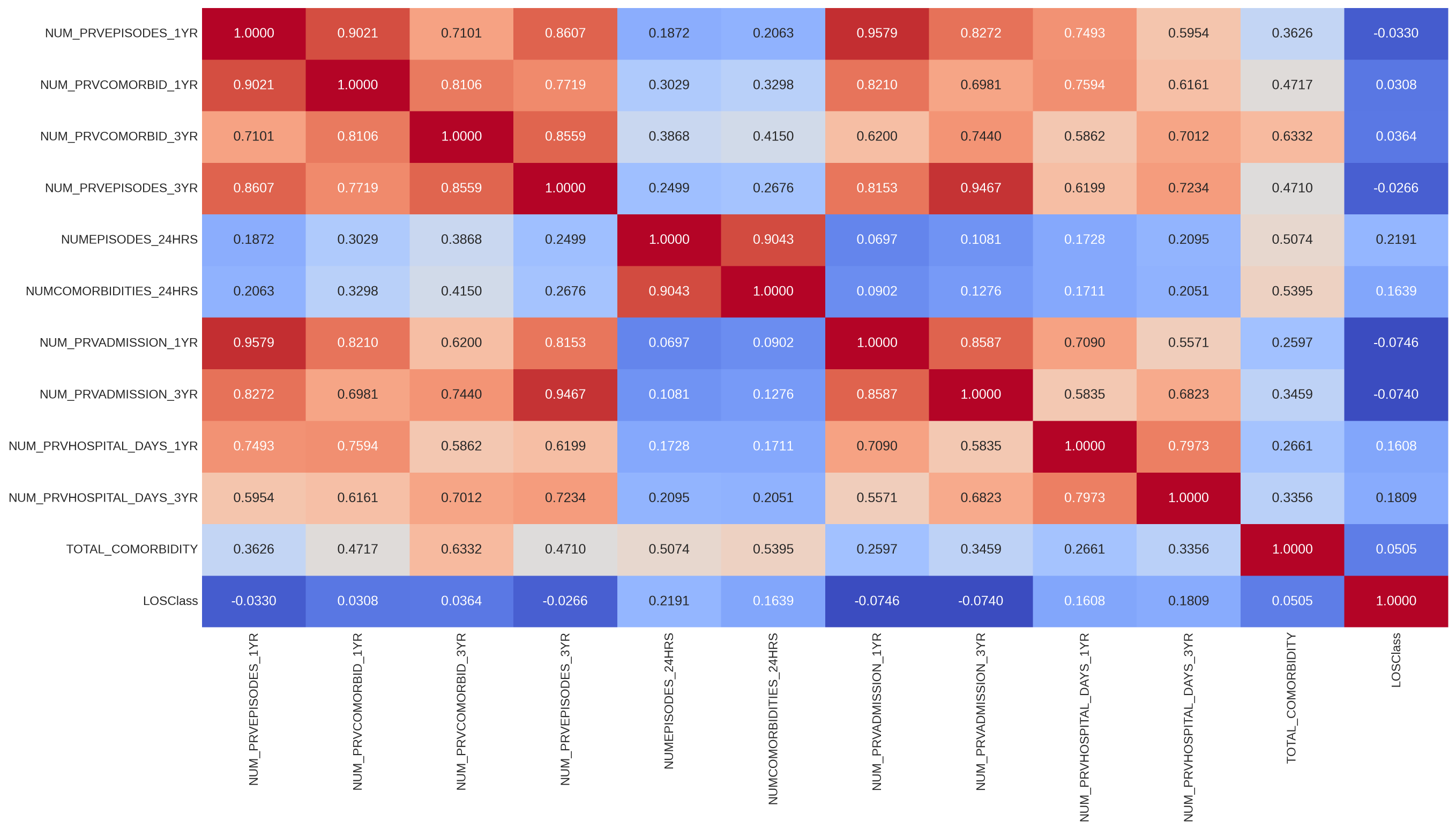}
    \caption{Feature correlation for males.}
    \label{fig: male_corr}
\end{figure}

\begin{figure}
         \centering
         \includegraphics[width=\textwidth]{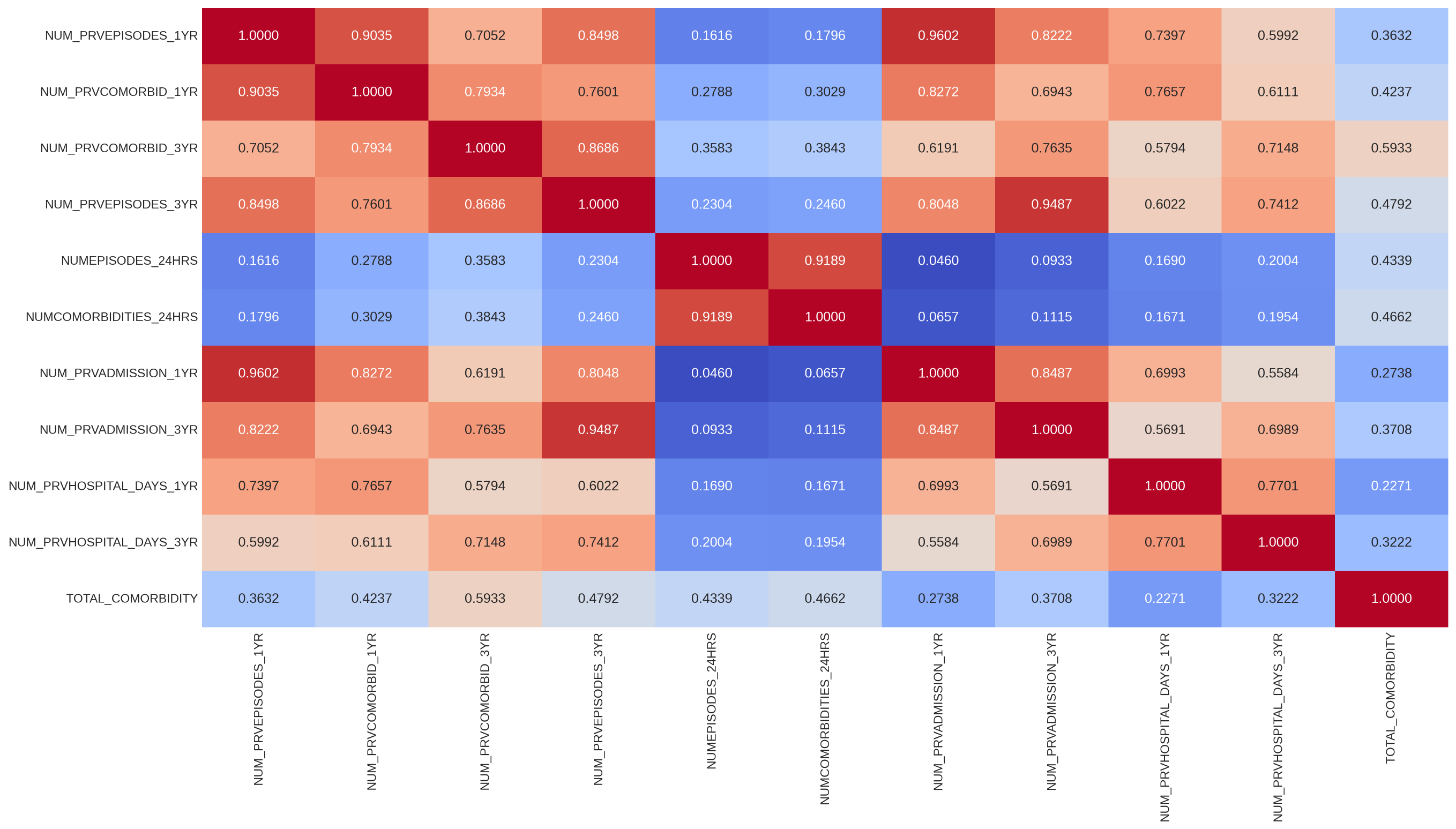}
         \caption{Feature correlation for females.}
         \label{fig: female_corr}
\end{figure}

\begin{center}
{
\footnotesize
\setlength{\tabcolsep}{2pt}
\begin{longtable}{@{}lllll@{}}
\caption{Results of normality tests applied to the dataset; KS denotes the Kolmogorov Smirnov test. The threshold for P-value is 0.05, hence, for P-value> 0.05, the normality test is positive otherwise, the independent variable is not normally distributed.  }
\label{Tab:Normality}\\
\toprule
\multirow{3}{*}{}           & \multicolumn{2}{l}{Male}              & \multicolumn{2}{l}{Female}            \\* \cmidrule(l){2-5} 
                            & \multicolumn{2}{l}{KS Normality Test} & \multicolumn{2}{l}{KS Normality Test} \\* \cmidrule(l){2-5} 
                            & Stat               & P-value             & Stat               & P-value             \\* \midrule
\endfirsthead
\multicolumn{5}{c}%
{{\bfseries Table \thetable\ continued from previous page}} \\
\toprule
\multirow{3}{*}{}           & \multicolumn{2}{l}{Male}              & \multicolumn{2}{l}{Female}            \\* \cmidrule(l){2-5} 
                            & \multicolumn{2}{l}{KS Normality Test} & \multicolumn{2}{l}{KS Normality Test} \\* \cmidrule(l){2-5} 
                            & Stat               & P-value             & Stat               & P-value             \\* \midrule
\endhead
NUM\_PRVEPISODES\_1YR       & 0.354              & 0.000            & 0.353              & 0.000            \\
NUM\_PRVCOMORBID\_1YR       & 0.222              & 0.000            & 0.217              & 0.000            \\
NUM\_PRVCOMORBID\_3YR       & 0.188              & 0.000            & 0.180              & 0.000            \\
NUM\_PRVEPISODES\_3YR       & 0.366              & 0.000            & 0.369              & 0.000            \\
NUMEPISODES\_24HRS          & 0.269              & 0.000            & 0.264              & 0.000            \\
NUMCOMORBIDITIES\_24HRS     & 0.324              & 0.000            & 0.313              & 0.000            \\
NUM\_PRVADMISSION\_1YR      & 0.397              & 0.000            & 0.399              & 0.000            \\
NUM\_PRVADMISSION\_3YR      & 0.401              & 0.000            & 0.412              & 0.000            \\
NUM\_PRVHOSPITAL\_DAYS\_1YR & 0.351              & 0.000            & 0.355              & 0.000            \\
NUM\_PRVHOSPITAL\_DAYS\_3YR & 0.345              & 0.000            & 0.349              & 0.000            \\
TOTAL\_COMORBIDITY          & 0.144              & 0.000            & 0.139              & 0.000 \\          \bottomrule
\end{longtable}
}
\end{center}

\newpage
\begin{center}
{\footnotesize
\setlength{\tabcolsep}{2pt}   
\begin{longtable}{@{}lllllllll@{}}
\caption{Comparison of the performance of classifiers used in predicting the \ac{los} for both male and female patients with \ac{ld}. }
\label{Tab:classifiers}\\
\toprule
\multirow{2}{*}{}     & \textbf{Male}  &       &       &          & \textbf{Female} &       &       &          \\* \cmidrule(l){2-9} 
                      & \textbf{AUC}   & \textbf{\ac{fnr}}   & \textbf{\ac{fpr}}   & \textbf{Balanced Accuracy }& \textbf{AUC}    & \textbf{\ac{fnr}}   & \textbf{\ac{fpr}}   & \textbf{Balanced Accuracy} \\* \midrule
\endfirsthead
\multicolumn{9}{c}%
{{\bfseries Table \thetable\ continued from previous page}} \\
\toprule
\multirow{2}{*}{}     & \textbf{Male}  &       &       &          & \textbf{Female} &       &       &          \\* \cmidrule(l){2-9} 
                      & \textbf{AUC}   & \textbf{\ac{fnr}}   & \textbf{\ac{fpr}}   & \textbf{Balanced Accuracy} & \textbf{AUC}    & \textbf{\ac{fnr}}   & \textbf{\ac{fpr}}   & \textbf{Balanced Accuracy} \\* \midrule
\endhead
\ac{lr} & 0.742 & 0.362 & 0.285 & 0.677    & 0.751  & 0.343 & 0.292 & 0.682    \\
RF                    & 0.759 & 0.224 & 0.396 & 0.690    & 0.756  & 0.229 & 0.392 & 0.689    \\
\ac{svm}                   & 0.742 & 0.420 & 0.243 & 0.669    & 0.747  & 0.376 & 0.266 & 0.679    \\
\ac{knn}                   & 0.679 & 0.375 & 0.369 & 0.628    & 0.681  & 0.385 & 0.359 & 0.628    \\
\ac{gboost}             & 0.742 & 0.399 & 0.264 & 0.668    & 0.747  & 0.401 & 0.251 & 0.674    \\
\ac{histgboost}              & 0.771 & 0.296 & 0.303 & 0.701    & 0.773  & 0.278 & 0.313 & 0.705    \\
\ac{xgboost}                & 0.763 & 0.284 & 0.326 & 0.695    & 0.761  & 0.286 & 0.331 & 0.692    \\
\ac{nn}                    & 0.716 & 0.496 & 0.210 & 0.647    & 0.723  & 0.473 & 0.233 & 0.647 \\ \midrule
\end{longtable}
}
\end{center}

\begin{center}
 {\footnotesize
\setlength{\tabcolsep}{2pt}
\begin{longtable}{p{2cm}p{2cm}p{2cm}p{2cm}p{2cm}p{2cm}p{2cm}p{2cm}p{2cm}}
\caption{Mean performance across the \ac{rf} classifier after training and testing over 10 random train/test sets.  }
\label{tab: Iterations_performance}\\
\toprule
\multicolumn{4}{l}{\textbf{MALE}}                                                                                                           & \multicolumn{4}{l}{\textbf{FEMALE}}                                                                                                         \\ \midrule
\textbf{AUC (\ac{std})} & \textbf{\ac{fnr} (\ac{std})} & \textbf{\ac{fpr} (\ac{std})} & \textbf{Balanced \newline Accuracy} \textbf{(\ac{std})} & \textbf{AUC (\ac{std})} & \textbf{\ac{fnr} (\ac{std})} & \textbf{\ac{fpr} (\ac{std})} & \textbf{Balanced \newline Accuracy} \textbf{(\ac{std})} \\* \midrule
\endfirsthead
\multicolumn{8}{c}%
{{\bfseries Table \thetable\ continued from previous page}} \\
\toprule
\multicolumn{4}{l}{\textbf{MALE}}                                                                                                           & \multicolumn{4}{l}{\textbf{FEMALE}}                                                                                                         \\ \midrule
\textbf{AUC (\ac{std})} & \textbf{\ac{fnr} (\ac{std})} & \textbf{\ac{fpr} (\ac{std})} & \textbf{Balanced Accuracy} \textbf{(\ac{std})} & \textbf{AUC (\ac{std})} & \textbf{\ac{fnr} (\ac{std})} & \textbf{\ac{fpr} (\ac{std})} & \textbf{Balanced Accuracy} \textbf{(\ac{std})} \\* \midrule
\endhead

0.758, \newline($\pm$ 0.003)                & 0.226, \newline($\pm$0.005)                & 0.394, \newline($\pm$0.007)                & 0.690,\newline ($\pm$0.003)                     & 0.762,\newline ($\pm$0.003)                & 0.217, \newline($\pm$0.005)                & 0.392,\newline ($\pm$0.006)                & 0.696,\newline ($\pm$0.002)                     \\* \bottomrule
\end{longtable}
    }
\end{center}

\begin{center}
    {\footnotesize
    \setlength{\tabcolsep}{2pt}
\begin{longtable}{@{}llll@{}}
\caption{Performance comparison of the unmitigated \ac{rf} classifier across Ethnic groups for males. The unmitigated model refers to the classifier without any bias mitigation algorithms applied. }
\label{tab:rf_ethnicityMale}\\
\toprule
\textbf{Ethnic group} & \textbf{\ac{fnr}}   & \textbf{\ac{fpr}}   & \textbf{Balanced Accuracy} \\* \midrule
\endfirsthead
\multicolumn{4}{c}%
{{\bfseries Table \thetable\ continued from previous page}} \\
\toprule
\textbf{Ethnic group} & \textbf{\ac{fnr}}   & \textbf{\ac{fpr}}   & \textbf{Balanced Accuracy} \\* \midrule
\endhead
\bottomrule
\endfoot
\endlastfoot
Asian        & 0.195 & 0.547 & 0.629             \\
Black        & 0.273 & 0.533 & 0.597             \\
Other        & 0.333 & 0.423 & 0.622             \\
Unknown      & 0.234 & 0.392 & 0.687             \\
White        & 0.222 & 0.394 & 0.692             \\* \bottomrule
\end{longtable}
    }
\end{center}

\begin{center}
{\footnotesize
\setlength{\tabcolsep}{2pt}
\begin{longtable}{@{}llll@{}}
\caption{Performance comparison of the unmitigated \ac{rf} classifier across Ethnic groups for females. The unmitigated model refers to the classifier without any bias mitigation algorithms applied. }
\label{tab:rf_ethnicityFemale}\\
\toprule
\textbf{Ethnic group} & \textbf{\ac{fnr}} & \textbf{\ac{fpr}} & \textbf{Balanced Accuracy} \\* \midrule
\endfirsthead
\multicolumn{4}{c}%
{{\bfseries Table \thetable\ continued from previous page}} \\
\toprule
\textbf{Ethnic group} & \textbf{\ac{fnr}} & \textbf{\ac{fpr}} & \textbf{Balanced Accuracy} \\* \midrule
\endhead
\bottomrule
\endfoot
\endlastfoot
Asian                 & 0.216        & 0.320        & 0.732                      \\
Black                 & 0.167        & 0.500        & 0.667                      \\
Other                 & 0.111        & 0.333        & 0.778                      \\
Unknown               & 0.219        & 0.398        & 0.692                      \\
White                 & 0.231        & 0.393        & 0.688                      \\* \bottomrule
\end{longtable}

}
\end{center}

\begin{center}
{\footnotesize
\setlength{\tabcolsep}{2pt}
\begin{longtable}{@{}llll@{}}
\caption{Performance comparison of the unmitigated \ac{rf} with threshold optimizer and exponentiated gradient across Ethnic groups for males. }
\label{tab:rf_comp_Male}\\
\toprule
\textbf{}             & \multicolumn{3}{r}{\textbf{Unmitigated RF Classifier Model}} \\* \midrule
\endfirsthead
\multicolumn{4}{c}%
{{\bfseries Table \thetable\ continued from previous page}} \\
\toprule

\endhead
\endfoot
\textbf{Ethnic group} & \textbf{\ac{fnr}}  & \textbf{\ac{fpr}} & \textbf{Balanced    Accuracy} \\* \midrule
Asian                 & 0.195         & 0.547        & 0.629                         \\
Black                 & 0.273         & 0.533        & 0.597                         \\
Other                 & 0.333         & 0.423        & 0.622                         \\
Unknown               & 0.234         & 0.392        & 0.687                         \\
White                 & 0.222         & 0.394        & 0.692                         \\* \midrule
\textbf{}             & \multicolumn{3}{r}{\textbf{ThresholdOptimizer}}              \\* \midrule
\textbf{Ethnic group} & \textbf{\ac{fnr}}  & \textbf{\ac{fpr}} & \textbf{Balanced Accuracy}    \\* \midrule
Asian                 & 0.297         & 0.399        & 0.652                         \\
Black                 & 0.318         & 0.467        & 0.608                         \\
Other                 & 0.333         & 0.385        & 0.641                         \\
Unknown               & 0.196         & 0.438        & 0.683                         \\
White                 & 0.179         & 0.453        & 0.684                         \\* \midrule
\textbf{}             & \multicolumn{3}{r}{\textbf{Reductions}}                      \\* \midrule
\textbf{Ethnic group} & \textbf{\ac{fnr}}  & \textbf{\ac{fpr}} & \textbf{Balanced Accuracy}    \\* \midrule
Asian                 & 0.219         & 0.581        & 0.600                         \\
Black                 & 0.227         & 0.733        & 0.520                         \\
Other                 & 0.267         & 0.346        & 0.694                         \\
Unknown               & 0.237         & 0.404        & 0.679                         \\
White                 & 0.228         & 0.390        & 0.691                         \\* \bottomrule
\end{longtable}
}
\end{center}

\begin{center}
{\footnotesize
\setlength{\tabcolsep}{2pt}
\begin{longtable}{@{}llll@{}}
\caption{Performance comparison of the unmitigated \ac{rf} with threshold optimizer and exponentiated gradient across Ethnic groups for females. }
\label{tab:rf_comp_Female}\\
\toprule
                      & \multicolumn{3}{r}{\textbf{Unmitigated RF Classifier Model}} \\* \midrule
\endfirsthead
\multicolumn{4}{c}%
{{\bfseries Table \thetable\ continued from previous page}} \\
\toprule
\textbf{Ethnic group} & \textbf{\ac{fnr}}   & \textbf{\ac{fpr}}  & \textbf{Balanced Accuracy}  \\* \midrule
\endhead
\textbf{Ethnic group} & \textbf{\ac{fnr}}   & \textbf{\ac{fpr}}  & \textbf{Balanced Accuracy}  \\* \midrule
Asian                 & 0.216          & 0.320         & 0.732                       \\
Black                 & 0.167          & 0.500         & 0.667                       \\
Other                 & 0.111          & 0.333         & 0.778                       \\
Unknown               & 0.219          & 0.398         & 0.692                       \\
White                 & 0.231          & 0.393         & 0.688                       \\* \midrule
                      & \multicolumn{3}{r}{\textbf{ThresholdOptimizer}}              \\* \midrule
\textbf{Ethnic group} & \textbf{\ac{fnr}}   & \textbf{\ac{fpr}}  & \textbf{Balanced Accuracy}  \\* \midrule
Asian                 & 0.176          & 0.349         & 0.737                       \\
Black                 & 0.167          & 0.500         & 0.667                       \\
Other                 & 0.111          & 0.222         & 0.833                       \\
Unknown               & 0.191          & 0.441         & 0.684                       \\
White                 & 0.195          & 0.437         & 0.684                       \\* \midrule
                      & \multicolumn{3}{r}{\textbf{Reductions}}                      \\* \midrule
\textbf{Ethnic group} & \textbf{\ac{fnr}}   & \textbf{\ac{fpr}}  & \textbf{Balanced Accuracy}  \\* \midrule
Asian                 & 0.206          & 0.314         & 0.740                       \\
Black                 & 0.167          & 0.500         & 0.667                       \\
Other                 & 0.111          & 0.278         & 0.806                       \\
Unknown               & 0.210          & 0.413         & 0.688                       \\
White                 & 0.226          & 0.399         & 0.687\\ \midrule                      
\end{longtable}
}
\end{center}

\begin{center}
{\footnotesize
\setlength{\tabcolsep}{2pt}
\begin{longtable}{@{}llll@{}}
\caption{Performance range for the unmitigated \ac{rf} model and bias-mitigated \ac{rf} models (threshold optimizer and reductions with exponentiated gradient) for males. }
\label{tab:Diff_male}\\
\toprule
                       & \ac{fnr} & \ac{fpr} & Balanced Accuracy \\* \midrule
\endfirsthead
\multicolumn{4}{c}%
{{\bfseries Table \thetable\ continued from previous page}} \\
\toprule
                       & \ac{fnr} & \ac{fpr} & Balanced Accuracy \\* \midrule
\endhead
\bottomrule
\endfoot
\endlastfoot
Unmitigated            & 0.138                 & 0.156                 & 0.095              \\
Threshold Optimizer    & 0.155                 & 0.082                 & 0.077              \\
Exponentiated Gradient & 0.048                 & 0.387                 & 0.174              \\* \bottomrule
\end{longtable}

}
\end{center}

\begin{center}
{\footnotesize
\setlength{\tabcolsep}{2pt}
\begin{longtable}{@{}llll@{}}
\caption{Performance range for the unmitigated \ac{rf} model and bias-mitigated \ac{rf} models (threshold optimizer and reductions with exponentiated gradient) for females. }
\label{tab:Diff_female}\\
\toprule
                       & \ac{fnr} & \ac{fpr} & balanced\_accuracy \\* \midrule
\endfirsthead
\multicolumn{4}{c}%
{{\bfseries Table \thetable\ continued from previous page}} \\
\toprule
                       & \ac{fnr} & \ac{fpr} & balanced\_accuracy \\* \midrule
\endhead
\bottomrule
\endfoot
\endlastfoot
Unmitigated            & 0.120                 & 0.180                 & 0.111              \\
Threshold Optimizer    & 0.084                 & 0.278                 & 0.167              \\
Exponentiated Gradient & 0.115                 & 0.222                 & 0.139              \\* \bottomrule
\end{longtable}
}
\end{center}
\begin{center}
{\footnotesize
\setlength{\tabcolsep}{2pt}
\begin{longtable}{l}
\caption{List of antipsychotic, antidepressant, and anti-manic/ anti-epileptic medications considered in this study. }
\label{tab: med_list}\\
\toprule
\textbf{MEDICATIONS}                               \\ \midrule
\endfirsthead
\multicolumn{1}{c}%
{{\bfseries Table \thetable\ continued from previous page}} \\
\toprule
\textbf{MEDICATIONS}                               \\ \midrule
\endhead
EPILIM                                             \\  
TEGRETOL                                           \\  
LAMOTRIGINE                                        \\  
LEVETIRACETAM                                      \\  
CARBAMAZEPINE                                      \\  
SODIUM VALPROATE                                   \\  
PHENYTOIN                                          \\  
TOPIRAMATE                                         \\  
GABAPENTIN                                         \\  
EPANUTIN                                           \\  
CLOBAZAM                                           \\  
CLONAZEPAM                                         \\  
PHENOBARBITAL                                      \\  
DEPAKOTE                                           \\  
PREGABALIN                                         \\  
KEPPRA                                             \\  
LAMICTAL                                           \\  
PRIMIDONE                                          \\  
VALPROIC ACID                                      \\  
PROCHLORPERAZINE                                   \\  
ZONISAMIDE                                         \\  
TOPAMAX                                            \\  
VIGABATRIN                                         \\  
LACOSAMIDE                                         \\  
MYSOLINE                                           \\  
RIVOTRIL                                           \\  
EPISENTA                                           \\  
OXCARBAZEPINE                                      \\  
RUFINAMIDE                                         \\  
ACETAZOLAMIDE                                      \\  
ETHOSUXIMIDE                                       \\  
PERAMPANEL                                         \\  
ESLICARBAZEPINE ACETATE                            \\  
SABRIL                                             \\  
ZONEGRAN                                           \\  
LYRICA                                             \\  
FLUOXETINE HYDROCHLORIDE                           \\  
NEURONTIN                                          \\  
ZARONTIN                                           \\  
TIAGABINE                                          \\  
On lithium                                         \\  
TRILEPTAL                                          \\  
DESITREND                                          \\  
ORLEPT                                             \\  
VIMPAT                                             \\  
MIRTAZAPINE                                        \\  
THIORIDAZINE                                       \\  
ZEBINIX                                            \\  
TAPCLOB                                            \\  
PAROXETINE HYDROCHLORIDE                           \\  
FRISIUM                                            \\  
QUETIAPINE                                         \\  
DIAMOX                                             \\  
VENLAFAXINE                                        \\  
SERTRALINE HYDROCHLORIDE                           \\  
CONVULEX                                           \\  
ESLICARBAZEPINE                                    \\  
RETIGABINE                                         \\  
CARBAGEN SR                                        \\  
EPIMAZ                                             \\  
LEVOMEPROMAZINE                                    \\  
TRAZODONE HYDROCHLORIDE                            \\  
CLOMIPRAMINE HYDROCHLORIDE                         \\  
AMITRIPTYLINE HYDROCHLORIDE   {[}ANTIDEPRESSANT{]} \\  
ZONISAMIDE                                         \\  
DOSULEPIN HYDROCHLORIDE                            \\  
PROMAZINE HYDROCHLORIDE                            \\  
OLANZAPINE                                         \\  
CHLORPROMAZINE HYDROCHLORIDE                       \\  
HALOPERIDOL {[}ANTIPSYCHOTIC{]}                    \\  
INOVELON                                           \\  
AMISULPRIDE                                        \\  
PHENOBARBITONE SODIUM                              \\  
GABITRIL                                           \\  
FYCOMPA                                            \\  
TERIL                                              \\  
EMESIDE                                            \\  
PALIPERIDONE                                       \\  
DULOXETINE                                         \\  
RISPERIDONE                                        \\ \bottomrule
\end{longtable}
}
\end{center}

{
\setlength{\tabcolsep}{2pt}

\begin{table}[!htb]
  
\begin{minipage}{.35\linewidth}
    \centering
    \footnotesize
    \caption{Raw Variable.}
    \label{tab:first_table}


\begin{tabular}{ c c } 
\toprule
\makecell{ \textbf{Admission ID}}  &  \textbf{PHYSICAL} \\   
\midrule
001 &  Yes\\
002 & No     \\
003 & Unknown    \\

\bottomrule
\end{tabular}
\end{minipage}
\begin{minipage}{.5\linewidth}
    \centering
    \footnotesize
    \caption{One-hot encoded variable.}
    \label{tab:second_table}


\begin{tabular}{ c c c c} 
    \toprule
    \makecell{ \textbf{Admission ID}}  & \textbf{ PHYSICAL\_Yes} & \textbf{PHYSICAL\_No} & \textbf{PHYSICAL\_Unknown} \\   
    \midrule
    001 &  1 & 0 & 0 \\
    002 &  0 & 1 & 0    \\
    003 &  0 & 0 & 1    \\
   
    \bottomrule
\end{tabular}
\end{minipage}
\end{table}
}

\section*{CATEGORIZATION FOR ALCOHOL STATUS PER PATIENT} 
\begin{figure}[!htb]
    \centering
    \includegraphics[width=0.9\textwidth]{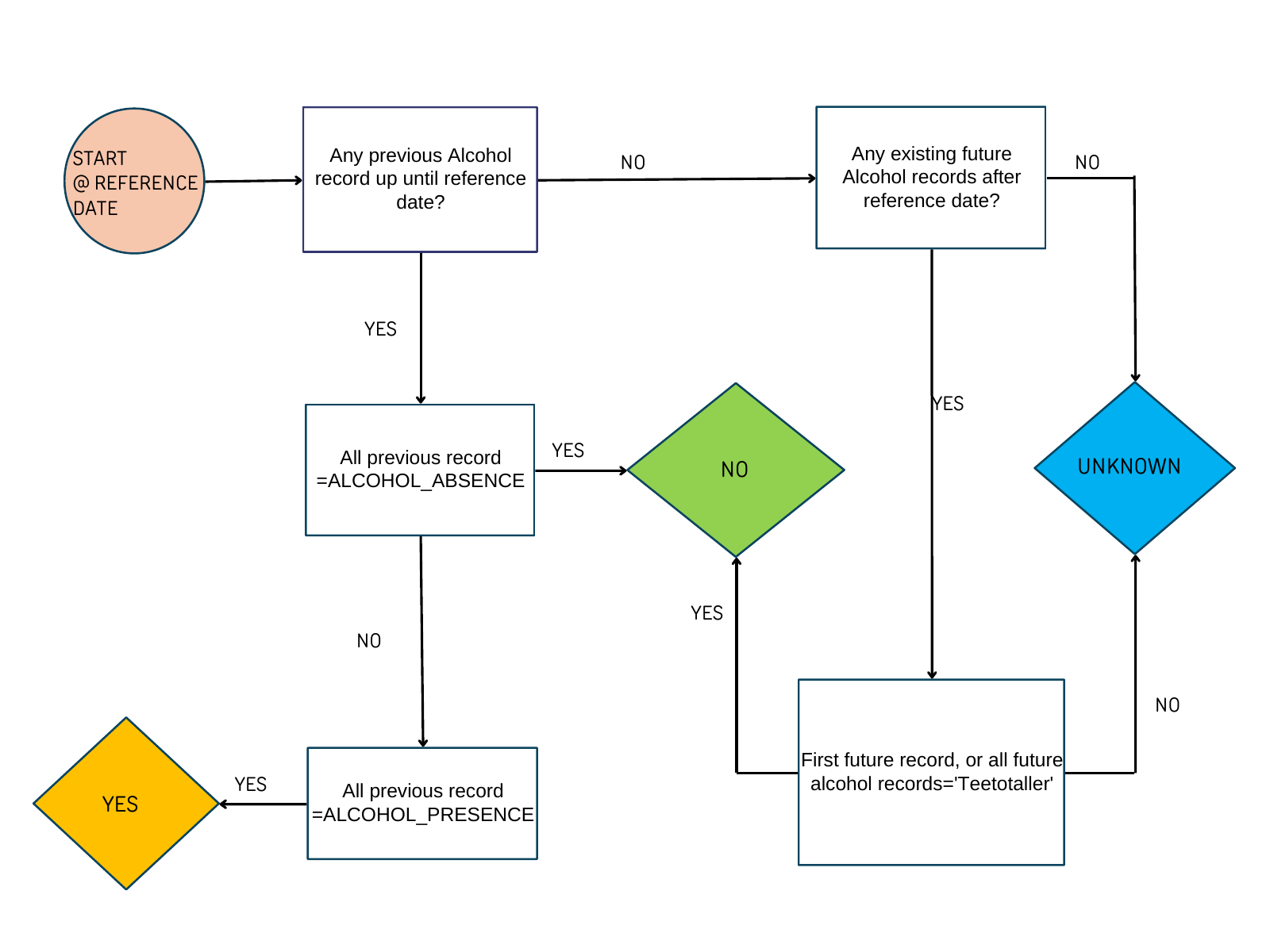}
    \caption{Flow chart diagram describing the algorithm for alcohol status at admission (reference date in the figure represents the admission date).}
    \label{fig: alcohol}
\end{figure}

To categorize the alcohol consumption history across patients as depicted in Figure \ref{fig: alcohol}, the following read code and \ac{icd10} code descriptions were classified into several sets as follows:  
\begin{itemize}
    \item NON\_DRINKER=['Teetotaller'] 

    \item CURRENT\_NON\_DRINKER=['Current non drinker'] 

    \item MODERATE\_DRINKER=['Ex-light drinker - (1-2u/day)', 'Drinks beer and spirits', 'Ex-trivial drinker (<1u/day)', 'Trivial drinker - <1u/day', Social drinker', 'Alcohol intake within recommended sensible limits', 'Beer drinker', 'Light drinker', 'Drinks wine', 'Moderate drinker', 'Spirit drinker'] 

    \item HEAVY\_DRINKER=['Mental and behavioural disorders due to use of alcohol','[X]Mental and behavioural disorders due to use of alcohol: psychotic disorder', 'Chronic alcoholism in remission','Alcohol-induced chronic pancreatitis', '[X]Mental and behavioural disorders due to use of alcohol: harmful use','[X]Mental and behavioural disorders due to use of alcohol', 'Alcohol withdrawal delirium','[X]Alcohol withdrawal-induced seizure','Chronic alcoholism NOS','Alcoholic hepatitis', 'Under care of community alcohol team','[X]Mental and behavioural disorders due to use of alcohol: dependence syndrome', 'Continuous chronic alcoholism','Alcohol dependence syndrome NOS','Alcoholic fatty liver','Very heavy drinker','Chronic alcoholism', 'Alcoholic liver damage unspecified','Very heavy drinker - >9u/day','Binge drinker','Alcohol intake above recommended sensible limits', 'Alcohol abuse monitoring','Hazardous alcohol use','Heavy drinker','Harmful alcohol use','Alcoholic cirrhosis of liver', 'Alcohol misuse','Acute alcoholic intoxication in alcoholism',]

    \item DRINKER=['Feels should cut down drinking', 'Declined referral to specialist alcohol treatment service', 'Alcohol dependence syndrome', 'Current non drinker', 'H/O: alcoholism', 'Alcohol withdrawal syndrome',] 

    \item FORMER\_DRINKER=['Ex-light drinker - (1-2u/day)','Ex-heavy drinker - (7-9u/day)','Ex-very heavy drinker-(>9u/d)', 'Ex-trivial drinker (<1u/day)'] 

    \item ALCOHOL\_ABSENCE=NON\_DRINKER + MODERATE\_DRINKER+ CURRENT\_NON\_DRINKER 

    \item ALCOHOL\_PRESENCE= HEAVY\_DRINKER + DRINKER + FORMER\_DRINKER 
\end{itemize}

\section*{CATEGORIZATION FOR SMOKING HISTORY PER PATIENT} 
\begin{figure}[!htb]
    \centering
    \includegraphics[width=\textwidth]{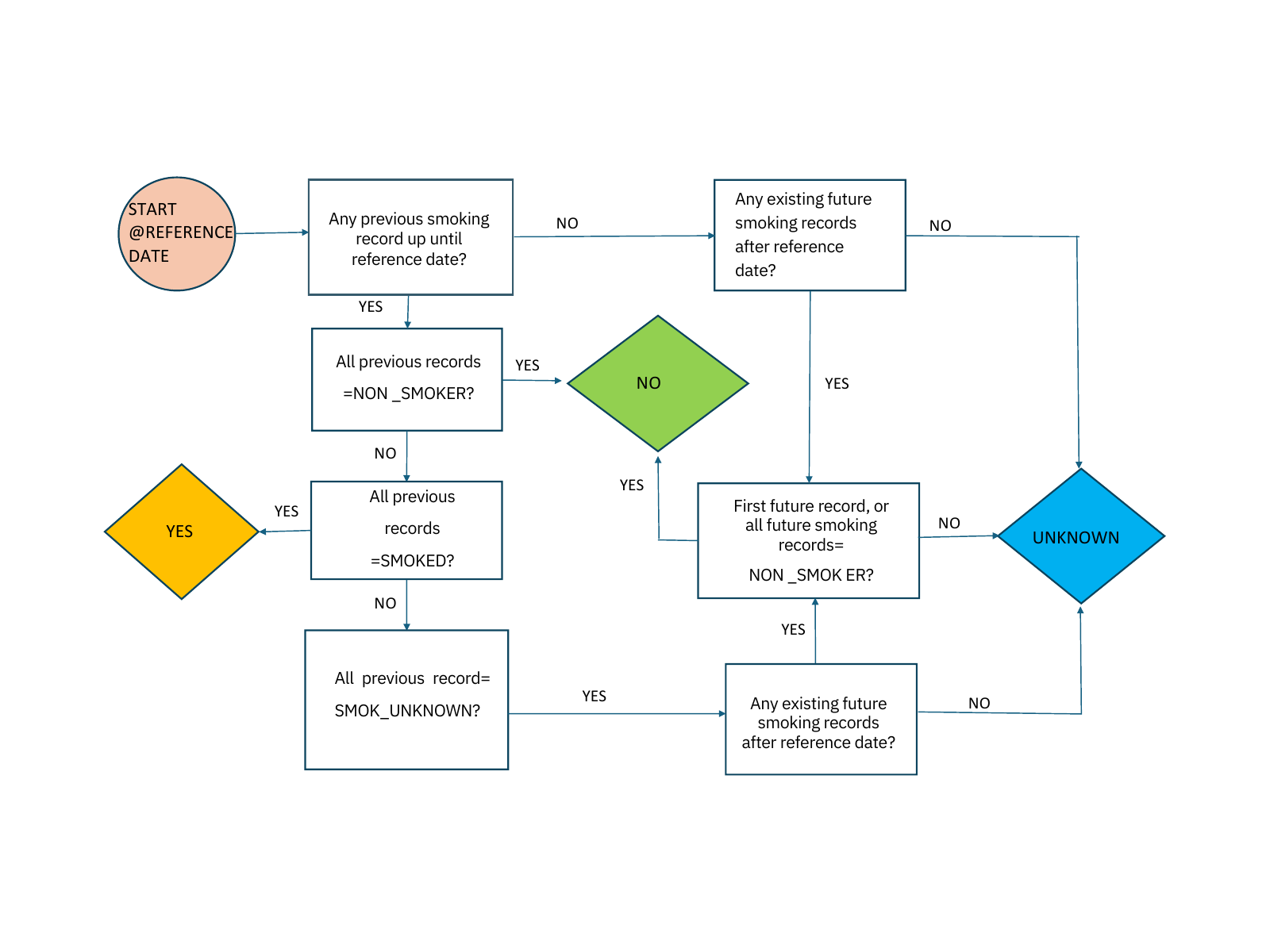}
    \caption{Flow chart diagram describing the algorithm for coding smoking history at admission (reference date in the figure represents the admission date).}
    \label{fig: smoking}
\end{figure}

To categorize the smoking history across patients as depicted in Figure \ref{fig: smoking}, the following readcode descriptions were classified into several sets as follows: 
\begin{itemize}
    \item NON\_SMOKER=['Never smoked tobacco','Current non-smoker'] 

    \item EX\_SMOKER=['Ex roll-up cigarette smoker','Ex cigar smoker','Ex pipe smoker','Ex-very heavy smoker (40+/day)', 'Ex-cigarette smoker','Ex-heavy smoker (20-39/day)','Ex smoker', 'Current non-smoker','Stopped smoking', 'Ex-smoker - amount unknown'] 

    \item SMOKER=['Smoking status at 52 weeks','Smoking restarted','Smoking status between 4 and 52 weeks','Smoking status at 4 weeks', 'Current smoker annual review - enhanced services administration','Minutes from waking to first tobacco consumption', 'Smoking cessation programme start date','Recently stopped smoking','Smoking free weeks','Smoking reduced', 'Negotiated date for cessation of smoking','Smoking started','Keeps trying to stop smoking','Failed attempt to stop smoking', 'Smoking cessation milestones','Not a passive smoker','Ready to stop smoking','Passive smoker','Thinking about stopping smoking', 'Pipe smoker','Trivial smoker - < 1 cig/day','Cigarette smoker','Current smoker','Rolls own cigarettes','Trying to give up smoking', 'Not interested in stopping smoking'] 

    \item SMOK\_UNKNOWN=['Refusal to give smoking status','Tobacco consumption unknown'] 

    \item SMOKED=SMOKER +EX\_SMOKER 
\end{itemize}

\begin{center}
{\footnotesize
\setlength{\tabcolsep}{2pt}

\begin{longtable}{@{}lllllll@{}}
\caption{Demographic distribution for training and validation sets for male cohort.  }
\label{tab: Traintest_male}\\
\toprule
\multirow{2}{*}{}  & \multicolumn{3}{l}{\textbf{Training}}                                & \multicolumn{3}{l}{\textbf{Test}}                                    \\* \cmidrule(l){2-7} 
                   & \textbf{ Unique Admissions (\%)} & \textbf{LOS 0} & \textbf{LOS 1} & \textbf{ Unique Admissions} & \textbf{LOS 0} & \textbf{LOS 1} \\* \midrule
\endfirsthead
\multicolumn{7}{c}%
{{\bfseries Table \thetable\ continued from previous page}} \\
\toprule
\multirow{2}{*}{}  & \multicolumn{3}{l}{\textbf{Training}}                                & \multicolumn{3}{l}{\textbf{Test}}                                    \\* \cmidrule(l){2-7} 
                   & \textbf{ Unique Admissions} & \textbf{LOS 0} & \textbf{LOS 1} & \textbf{ Unique Admissions (\%)} & \textbf{LOS 0} & \textbf{LOS 1} \\* \midrule
\endhead
Total              & 12550                              & 6275           & 6275           & 16138                              & 9863           & 6275           \\* \midrule
Age                &                                    &                &                &                                    &                &                \\* \midrule
\textless{}20      & 41                                 & 29             & 12             & 38                                 & 25             & 13             \\
20-29              & 564                                & 295            & 269            & 751                                & 470            & 281            \\
30-39              & 1865                               & 1051           & 814            & 2412                               & 1638           & 774            \\
40-49              & 2644                               & 1419           & 1225           & 3493                               & 2241           & 1252           \\
50-59              & 3017                               & 1537           & 1480           & 3816                               & 2363           & 1453           \\
60-69              & 2580                               & 1209           & 1371           & 3306                               & 1914           & 1392           \\
70-79              & 1446                               & 627            & 819            & 1787                               & 985            & 802            \\
80+                & 393                                & 108            & 285            & 535                                & 348            & 227            \\* \midrule
Ethnic groups      &                                    &                &                &                                    &                &                \\* \midrule
Asian              & 239                                & 99             & 140            & 276                                & 148            & 128            \\
Black              & 35                                 & 18             & 17             & 37                                 & 15             & 22             \\
Other              & 31                                 & 20             & 11             & 41                                 & 26             & 15             \\
Unknown            & 2365                               & 1135           & 1230           & 3110                               & 1917           & 1193           \\
White              & 9880                               & 5003           & 4877           & 12674                              & 7757           & 4917           \\* \midrule
WIMD               &                                    &                &                &                                    &                &                \\* \midrule
1 (Most Deprived)  & 3529                               & 17820          & 1747           & 4489                               & 2806           & 1683           \\
2                  & 2656                               & 1341           & 1315           & 3540                               & 2181           & 1359           \\
3                  & 1959                               & 939            & 1020           & 2458                               & 1404           & 1054           \\
4                  & 1640                               & 765            & 875            & 2101                               & 1249           & 852            \\
5 (Least Deprived) & 1232                               & 610            & 622            & 1521                               & 937            & 584            \\
Unknown            & 1534                               & 838            & 696            & 2029                               & 1286           & 743      \\ \bottomrule     
\end{longtable}
}
\end{center}

\begin{center}
{\footnotesize
\setlength{\tabcolsep}{2pt}
\begin{longtable}{@{}lllllll@{}}
\caption{Demographic distribution for training and validation sets for female cohort.  }
\label{tab: Traintest_female}\\
\toprule
\multirow{2}{*}{}  & \multicolumn{3}{l}{Training}                     & \multicolumn{3}{l}{Test}                  \\* \cmidrule(l){2-7} 
                   &  Unique Admissions  & LOS 0        & LOS 1 &  Unique Admissions & LOS 0 & LOS 1 \\* \midrule
\endfirsthead
\multicolumn{7}{c}%
{{\bfseries Table \thetable\ continued from previous page}} \\
\toprule
\multirow{2}{*}{}  & \multicolumn{3}{l}{Training}                     & \multicolumn{3}{l}{Test}                  \\* \cmidrule(l){2-7} 
                   &  Unique Admissions  & LOS 0        & LOS 1 &  Unique Admissions & LOS 0 & LOS 1 \\* \midrule
\endhead
Total              & 11626                     & 5813         & 5813  & 14984                     & 9170  & 5814  \\* \midrule
Age                &                           &              &       &                           &       &       \\* \midrule
\textless{}20      & 31                        & 16           & 15    & 33                        & 22    & 11    \\
20-29              & 571                       & 282          & 289   & 708                       & 445   & 263   \\
30-39              & 1595                      & 915          & 680   & 2101                      & 1393  & 708   \\
40-49              & 2231                      & 1251         & 980   & 3099                      & 2090  & 1009  \\
50-59              & 2614                      & 1432         & 1182  & 3428                      & 2206  & 1222  \\
60-69              & 2435                      & 1216         & 1219  & 3203                      & 1966  & 1237  \\
70-79              & 1460                      & 517          & 943   & 1661                      & 763   & 898   \\
80+                & 689                       & 184          & 505   & 751                       & 285   & 466   \\* \midrule
Ethnic group          &                           &              &       &                           &       &       \\* \midrule
Asian              & 207                       & 109          & 98    & 274                       & 172   & 102   \\
Black              & 19                        & \textless{}5 & *     & 18                        & 6     & 12    \\
Other              & 27                        & *            & *     & 27                        & 18    & 9     \\
Unknown            & 2002                      & 945          & 1057  & 2613                      & 1567  & 1046  \\
White              & 9371                      & 4738         & 4633  & 12052                     & 7407  & 4645  \\* \midrule
WIMD               &                           &              &       &                           &       &       \\* \midrule
1 (Most Deprived)  & 3195                      & 1598         & 1597  & 4090                      & 2501  & 1589  \\
2                  & 2589                      & 1357         & 1232  & 3455                      & 2233  & 1222  \\
3                  & 1833                      & 862          & 971   & 2374                      & 1376  & 998   \\
4                  & 1766                      & 889          & 877   & 2185                      & 1321  & 864   \\
5 (Least Deprived) & 1109                      & 559          & 550   & 1369                      & 801   & 568   \\
Unknown            & 1134                      & 548          & 586   & 1511                      & 938   & 573  \\ \bottomrule
\end{longtable}
Additional cells have been masked to prevent disclosure
}
\end{center}

\begin{center}
    {\footnotesize
\setlength{\tabcolsep}{2pt}
\begin{longtable}{lll}
\caption{Parameter configurations for all classification models.  }
\label{tab:Class_param}\\
\toprule
\textbf{Classification models}                     & \textbf{Parameters}          & \textbf{Value}                        \\ \midrule
\endfirsthead
\multicolumn{3}{c}%
{{\bfseries Table \thetable\ continued from previous page}} \\
\toprule
\textbf{Classification models}                     & \textbf{Parameters}          & \textbf{Value}                        \\ \midrule
\endhead
\multirow{5}{*}{\ac{lr}} & C                   & 1.0                          \\   
                                          & Max\_iter           & 1000                         \\   
                                          & penalty             & l2                           \\   
                                          & Random\_state       & none                         \\   
                                          & Multi\_class        & auto                         \\ \midrule
\multirow{6}{*}{\ac{rf}}                       & Max\_depth=2        & None                         \\   
                                          & Random\_state=0     & None                         \\   
                                          & N\_estimators       & 100                          \\   
                                          & Max\_samples        & None                         \\   
                                          & Max\_features       & sqrt                         \\   
                                          & criterion           & gini                         \\ \midrule
\multirow{5}{*}{\ac{knn}}                      & N\_neighbors        & 5                            \\   
                                          & Weights             & Uniform                      \\   
                                          & Metric=euclidean    & Euclidean  (i.e., minkowski) \\   
                                          & p                   & 2                            \\   
                                          & Leaf\_size          & 30                           \\ \midrule
\multirow{9}{*}{GBoost}                   & Learning\_rate      & 0.1                          \\   
                                          & N\_estimators       & 100                          \\   
                                          & Max\_depth          & 1                            \\   
                                          & Min\_samples\_split & 2                            \\   
                                          & Min\_samples\_leaf  & 1                            \\   
                                          & Random\_state       & 0                            \\   
                                          & subsample           & 1.0                          \\   
                                          & loss                & Log\_loss                    \\   
                                          & criterion           & Friedman\_mse                \\ \midrule
\multirow{6}{*}{\ac{svm}}                      & C                   & 1.0                          \\   
                                          & kernel              & linear                       \\   
                                          & gamma               & scale                        \\   
                                          & degree              & 3                            \\   
                                          & Probability         & True                         \\   
                                          & Max\_iter           & -1                           \\ \midrule
\multirow{5}{*}{HISTGBoost}               & Max\_bins           & 255                          \\   
                                          & Max\_iter           & 100                          \\   
                                          & loss                & Log\_loss                    \\   
                                          & Learning\_rate      & 0.1                          \\   
                                          & Max\_depth          & None                         \\ \midrule
\multirow{4}{*}{XGBoost}                  & Learning rate       & None                         \\   
                                          & Max\_depth          & None                         \\   
                                          & Eval\_metric        & `mlog\_loss’                 \\   
                                          & N\_estimators       & 100                          \\ \bottomrule
\end{longtable}
}
{Remark: Other parameters not included in this table take in default values set by the sklearn \ac{ml} library.}

\begin{center}
    {\footnotesize
\setlength{\tabcolsep}{2pt}
\begin{longtable}{llll}
\caption{Parameter summary for the sequential \ac{nn} with the male cohort.  }
\label{tab: NN_male}\\
\toprule
\multicolumn{1}{l}{Layer (type)} & \multicolumn{1}{l}{Output shape}    & \multicolumn{1}{l}{Param \#} & Tr. Param \# \\ \midrule
\endfirsthead
\endhead
\multicolumn{1}{l}{Linear-1}     & \multicolumn{1}{l}{{[}11578, 75{]}} & \multicolumn{1}{l}{5\,700}    & 5\,700        \\ 
\multicolumn{1}{l}{ReLU-2}       & \multicolumn{1}{l}{{[}11578, 75{]}} & \multicolumn{1}{l}{0}        & 0            \\ 
\multicolumn{1}{l}{Linear-3}     & \multicolumn{1}{l}{{[}11578, 2{]}}  & \multicolumn{1}{l}{152}      & 152          \\ \midrule
\multicolumn{4}{l}{Total params: 5\,852}                                                                                \\ 
\multicolumn{4}{l}{Trainable params: 5\,852}                                                                            \\ 
\multicolumn{4}{l}{Non-trainable params: 0}                                                                            \\ \bottomrule
\end{longtable}
    }
\end{center}

\begin{center}
    {\footnotesize
\setlength{\tabcolsep}{2pt}
\begin{longtable}{llll}
\caption{Parameter summary for the sequential \ac{nn} with the female cohort.  }
\label{tab: NN_female}\\
\toprule
Layer (type) & Output shape    & Param \# & Tr. Param \# \\ \midrule
\endfirsthead
\endhead
Linear-1     & {[}10732, 78{]} & 6\,162   & 6\,162       \\
ReLU-2       & {[}10732, 78{]} & 0        & 0            \\
Linear-3     & {[}10732, 2{]}  & 158      & 158          \\ \midrule
\multicolumn{4}{l}{Total params: 6\,320}                 \\
\multicolumn{4}{l}{Trainable params: 6\,320}             \\
\multicolumn{4}{l}{Non-trainable params: 0}             \\ \bottomrule
\end{longtable}
    }
\end{center}

\end{center}

\label{LastPageOfSupplementary}

\clearpage

\fancyfoot[r]{\small\sffamily\bfseries\thepage/\pageref{LastPageOfMain}}



\bibliographystyle{ieeetr}
\bibliography{references}  






\end{document}